%% file: main.tex
\documentclass{article}

\usepackage{iclr2027_conference,times}

\usepackage[normalem]{ulem}

\usepackage{amsmath}
\usepackage{amsfonts}
\usepackage{amssymb}
\usepackage{colortbl}
\usepackage{stackengine}
\usepackage{float}
\usepackage{wrapfig} 
\usepackage{multirow}
\usepackage{wrapfig}
\usepackage{tabularx}
\usepackage{caption}
\usepackage{graphicx}
\usepackage{tabularx}
\usepackage{booktabs}
\usepackage{placeins}
\usepackage{CJKutf8}
\usepackage[dvipsnames,table]{xcolor}
\newcommand{\inc}[1]{%
    {\scriptsize\textcolor{green!45!black}{$_{+#1}$}}%
}

\newcommand{\dec}[1]{%
    {\scriptsize\textcolor{red!70!black}{$_{-#1}$}}%
}
\definecolor{mygray}{rgb}{0.9, 0.9, 0.9}
\usepackage[utf8]{inputenc}
\usepackage[T1]{fontenc}

\usepackage[colorlinks,citecolor=gray]{hyperref}
\usepackage{url}
\usepackage{nicefrac}
\usepackage{microtype}

\usepackage{pifont}

\usepackage{algorithm}
\usepackage{algpseudocode}
\usepackage{listings}

\newcommand{\zh}[1]{\begin{CJK}{UTF8}{gbsn}#1\end{CJK}}
\newcommand{\myparagraph}[1]{\vspace{-2pt}\paragraph{#1}}

\usepackage[resetlabels]{multibib}
\newcites{Appendix}{Appendix References}
\bibliographystyleAppendix{iclr2027_conference}

\usepackage{fancyhdr}
\lstdefinestyle{pythonappendix}{
    language=Python,
    basicstyle=\ttfamily\scriptsize,
    keywordstyle=\bfseries,
    commentstyle=\color{gray},
    stringstyle=\color{gray},
    showstringspaces=false,
    breaklines=true,
    breakatwhitespace=false,
    columns=fullflexible,
    keepspaces=true,
    frame=single,
    framerule=0.2pt,
    xleftmargin=0.5em,
    xrightmargin=0.5em
}

\title{Do Vision Transformers Need All-to-All \\ Attention? Global Communication Through \\ Elastic Learned Cores}

\author{
\textbf{Alan Z. Song}$^{\dagger 1}$,
\textbf{Yinjie Chen}$^{\dagger 2}$,
\textbf{Mu Nan}$^{2}$,
\\[-0.15em]
\hspace{0.00em}
\textbf{Deva Ramanan}$^{1}$,
\textbf{Michael J. Tarr}$^{1}$,
\textbf{Andrew F. Luo}$^{2*}$
\\[0.15em]
$^{1}$ Carnegie Mellon University,
\qquad
$^{2}$ The University of Hong Kong
\\
$^{\dagger}$ Co-first authors, equal contribution
\qquad
$^{*}$ Corresponding author
\\
\texttt{zixis@andrew.cmu.edu}
\qquad
\texttt{maxwellcaffrey915@gmail.com}
\\
\texttt{deva@cs.cmu.edu}
\qquad
\texttt{michaeltarr@cmu.edu}
\qquad
\texttt{aluo@hku.hk}
}

\iclrfinalcopy

\begin{document}

\maketitle
\begingroup

\input{sections/00_Abstract}
\input{sections/01_Intro}
\input{sections/02_Related}

\input{sections/03_Methods}

\input{sections/04_Results}

\input{sections/05_Conclusion}
\clearpage

\zh{\bibliography{myref}}

\clearpage

\appendix
\input{sections/90_appendix_arxiv}

%

\end{document}

%% file: sections/00_Abstract.tex
\begin{abstract}

Vision Transformers (ViTs) learn rich visual-semantic representations through all-to-all self-attention among patch tokens. However, this design implicitly assumes that direct pairwise patch interactions are necessary for effective representation learning. In this work, we challenge this assumption and show that representations supporting both global recognition and dense prediction can be learned without direct patch-to-patch interaction. We propose \textbf{VECA} (Visual Elastic-Core Attention), a vision transformer with core-periphery structured attention mediated by a small set of learned cores. Patch tokens exchange global information exclusively through these cores, while the full set of dense patches are preserved and iteratively updated across layers. This reduces attention complexity from $O(N^2)$ to $O(N)$, linear in the number of patches $N$ for a fixed core budget $C$. Unlike prior latent-token cross-attention architectures, VECA facilitates sparse global communication without compressing the spatial representation itself. Nested training along the core axis further enables a single model to elastically trade off computation and accuracy at inference time without retraining. Across image classification and dense prediction tasks, VECA remains competitive with full-attention backbones and outperforms the evaluated linear-complexity alternatives on most benchmarks. Moreover, without explicit supervision, these cores develop semantically organized structures that support object-label transfer across video frames. These results show that effective visual representations can be learned without direct all-to-all patch interaction. \href{https://github.com/alansong1322/VECA}{Project repository here.}
\vspace{-3mm}
\end{abstract}

%% file: sections/01_Intro.tex
\section{Introduction}

Recent advances in computer vision have been driven by increases in computational resources and dataset scale. These developments have enabled a shift away from strong inductive biases and towards more flexible, data-driven models~\citep{tolstikhin2021mlp, touvron2022resmlp,liu2021pay,lee2022fnet,rao2023gfnet}. Whereas convolutional neural networks directly encode locality and translation equivariance into their structure~\citep{luo2016understanding,islam2020much}, these design choices constrain the representations that can be learned at scale. As datasets have grown, the value of these inductive biases has diminished. Vision Transformers (ViTs, ~\cite{dosovitskiy2020image}) rely on self-attention with only weak spatial constraints from patchification, enabling global receptive fields at every layer~\citep{raghu2021vision,park2022vision}. In standard ViTs, this global communication is realized through dense token-to-token interaction, where every patch attends to every other patch. While effective, these all-to-all interactions incur \emph{quadratic complexity}, becoming increasingly costly at high resolutions~\citep{li2023rethinking,nauen2025transformer}. Latent-mediated cross-attention offers an efficient alternative, but compressing inputs into a small set of latents can discard the spatially aligned representations maintained by self-attention. This motivates a fundamental question: \emph{Do Vision Transformers require all-to-all patch interactions to learn effective visual representations?}

\input{fig_text/0_teaser_text}

In this work, we introduce Visual Elastic-Core Attention (\textbf{VECA}), which replaces standard self-attention with a block-sparse attention pattern mediated by a small, resolution-invariant set of $C$ learned core tokens. Viewed as an attention connectivity graph, VECA forms a \textit{core-periphery network}~\citep{rombach2014core,zhang2015identification}: the cores form a fully connected clique~\citep{luce1949method} that interacts with all image patches, while patches never directly attend to one another. Information is therefore exchanged through the cores while all $N$ dense patch representations are preserved and iteratively updated across layers. The resulting attention cost is $(2NC+C^2)$, which is linear in $N$ for a fixed core budget $C$. VECA's core-based design also naturally supports nested dropout~\citep{rippel2014learning} along the core axis. This allows the number of active cores to vary at inference time and enables a single model to elastically trade computation for accuracy. 

Following recent work, we explore our proposed architecture by {\em distilling} representations from strong vision foundation models~\citep{wei2022contrastive,ranzinger2024radio,shang2024theia,sariyildiz2024unic}. Specifically, we train VECA using a DINOv3~\citep{simeoni2025dinov3} teacher. Across image classification and dense prediction tasks, VECA remains competitive with full-attention backbones and outperforms the evaluated alternatives on most benchmarks. Notably, at $512\times512$ resolution, VECA-B/16 with only $C=8$ cores retains $95.7\%$ of the PASCAL VOC~\citep{everingham2015pascal} performance of the full-attention ViT-B/16 while using only $1.6\%$ as many attention interactions. We illustrate the architecture in Figure~\ref{fig:teaser}. We also observe \emph{emergent} structure in the cores, which develop semantically aligned organization without explicit supervision and support label transfer from a reference frame to later video frames. Our contributions are therefore threefold: \textbf{(1)} a core-periphery attention architecture that removes direct patch-to-patch interaction while preserving and continually updating dense patch representations; \textbf{(2)} nested core training that enables elastic computation at inference time; and \textbf{(3)} an analysis showing that repeated interaction with the continually updated patch tokens leads cores to specialize to distinct objects and parts, supporting object-label transfer across video frames. Together, our results show that effective visual representations can be learned without all-to-all patch interaction.

%% file: fig_text/0_teaser_text.tex

\begin{figure}[t!]
  \centering
  \vspace{-0.5em}
\includegraphics[width=1.0\linewidth]{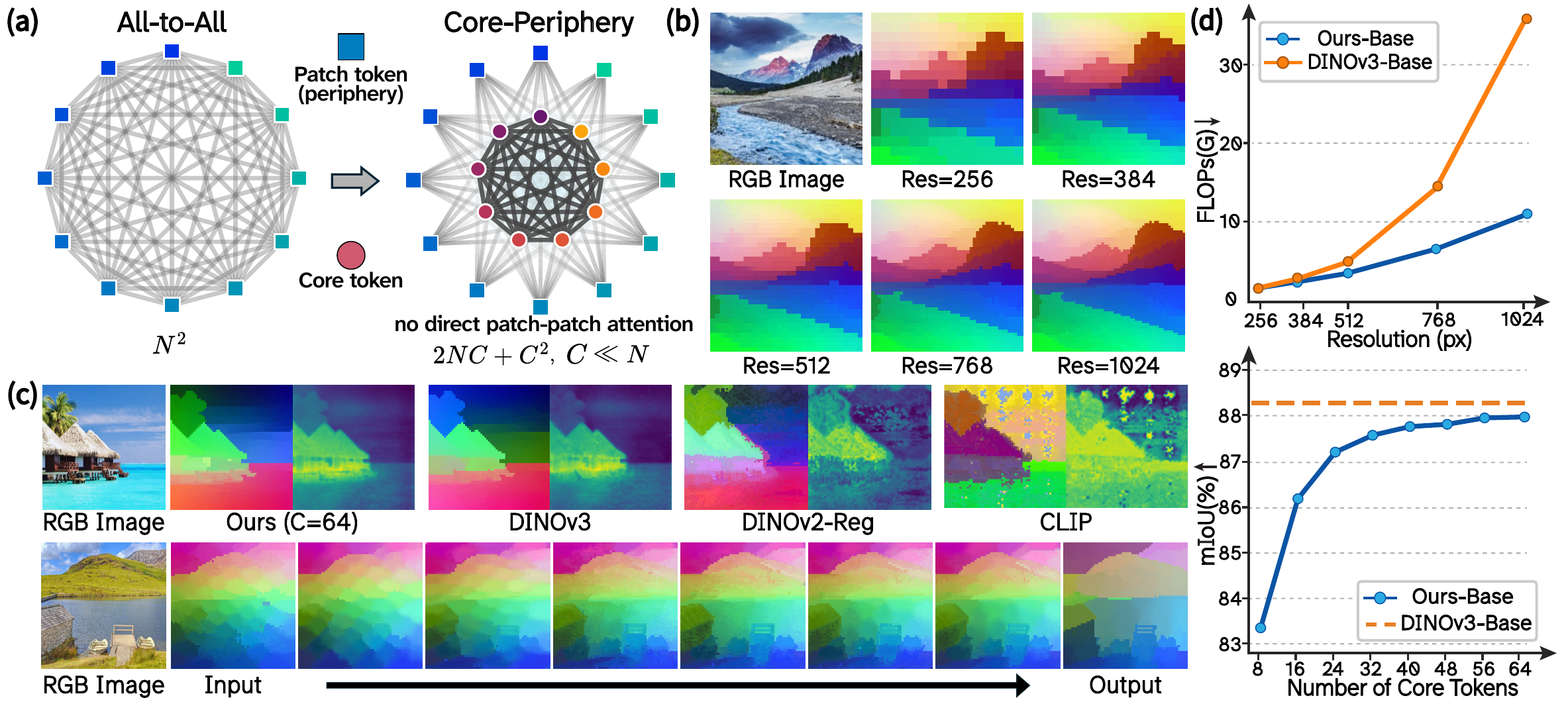}
  \vspace{-3mm}
    \caption{\textbf{VECA overview and learned representations. }
    \textbf{(a)} Standard attention utilizes a fully connected attention matrix with $N^2$ comparisons given N input patches. \textbf{VECA} constructs a core-periphery matrix with $C$ core tokens that form a clique, eliminating dense patch-to-patch interaction and requiring only $2NC+C^2$ comparisons.
    \textbf{(b)} Dense feature visualization with UMAP. Our method produces stable embeddings across resolutions.
    \textbf{(c)} \textbf{Top}: We visualize our features and \texttt{[CLS]}-patch cosine similarity of our model, comparing the visualizations to its teacher (DINOv3) as well as other baseline embeddings. \textbf{Bottom}: We visualize the cores' attention maps to patches (Appendix~\ref{supp_feedforward_clustering}) with UMAP. The maps start visually isotropic (spherical) and become increasingly semantic with depth.
    \textbf{(d)} \textbf{Top}: FLOPs of our attention block versus DINOv3 across resolutions. \textbf{Bottom}: Elastically varying the number of active core tokens, trading off accuracy and speed in PASCAL VOC.
    }
  \vspace{-0.4cm}
  \label{fig:teaser}
\end{figure}

%% file: sections/02_Related.tex
\section{Related work}

\input{fig_text/1_arch_text}

\myparagraph{Efficient and Latent-Mediated Attention.}
To mitigate the quadratic cost of self-attention, prior work approximates softmax~\citep{shen2021efficient,katharopoulos2020transformers,choromanski2020rethinking,zhang2024hedgehog,joshi2025replacing,lu2026zeros}, factorizes attention at low rank~\citep{wang2020linformer,xiong2021nystromformer}, restricts it to hashed, sparse, or sliding-window patterns~\citep{kitaev2020reformer,zaheer2020big,beltagy2020longformer,roy2021efficient,acharya2024star}, or replaces it with unparameterized transforms~\citep{lee2022fnet,tay2021synthesizer}.
Linear RNNs offer an alternative for sequences~\citep{peng2021random,irie2021going,gu2022parameterization,peng2022abc,mao2022fine,yang2023gated,sun2023retentive,gu2023mamba,qin2024hgrn2,zhang2024gated,beck2024xlstm,peng2025rwkv,hu2025improving}, and linear attention has been applied to vision with some performance tradeoffs~\citep{bolya2022hydra,cai2022efficientvit,you2023castling,han2023flatten,meng2025polaformer}.
Latent-mediated designs instead route communication through a small token set: Set Transformer and Agent Attention use layer-local latent tokens~\citep{lee2019set,han2024agent}, Luna packs inputs into a fixed-length sequence via nested attention~\citep{ma2021luna}, Perceiver compresses inputs into a latent array~\citep{jaegle2021perceiver,jaegle2021perceiverio,dolga2024latte,jajal2026adaperceiver}, and bottleneck tokens mediate multimodal fusion~\citep{nagrani2021attention}.
VECA retains exact softmax attention and obtains linear cost from a sparse core-periphery graph. Unlike Perceiver, it updates the full patch stream at every layer; unlike Set Transformer and Agent Attention, its cores persist across depth and share a single attention operator with the patches. VECA is also natively elastic, enabling test-time tradeoffs between compute and accuracy.

\myparagraph{Elastic and Adaptive Computation.}
Elastic inference trades computation for accuracy at test time~\citep{zilberstein1996using}. Matryoshka representations learn truncatable embeddings~\citep{kusupati2022matryoshka}, and nested dropout learns ordered parameter hierarchies~\citep{rippel2014learning,staley2022triangular,shen2024distributional,ho2025ordered,ladjal2019pca}, later extended to width-adaptive networks~\citep{tann2016runtime,kim2018nestednet,yu2018slimmable} and sub-network extraction~\citep{cai2019once,valipour2023sortednet,li2024subnetwork,zhang2024slicing}. Transformer-specific designs vary model width~\citep{kudugunta2024matformer,cai2024flextron}, whereas VECA uses core-based elasticity. Input-dependent computation uses early exiting~\citep{teerapittayanon2016branchynet,wu2018blockdrop,liu2020fastbert} or iterative processing~\citep{graves2016adaptive,dehghani2018universal,xie2025accelerating,jeddi2026loopformer}, while data-level elasticity uses token merging~\citep{bolya2022token}, pruning of spatially redundant tokens~\citep{yin2022vit,rao2021dynamicvit}, scratchpads~\citep{xue2023adaptive}, or iterative expansion~\citep{hojjat2025thinkingvit}. Sparse experts offer another axis~\citep{gu2025elastic,wang2025training}. Closest to our work, MQT learns elastic queries~\citep{hu2024matryoshka}, but as a single layer on a frozen, quadratic-time CLIP model. In contrast, VECA is an elastic \emph{visual backbone} competitive with state-of-the-art vision foundation models on both \emph{classification} and \emph{dense tasks}.

\myparagraph{Differentiable Clustering and Structured Representations.}
Learnable clustering jointly optimizes representations and their groupings, using discrete assignment~\citep{xie2016unsupervised,yang2017towards,duan2019improving} or continuous relaxations~\citep{shah2018deep,bianchi2020spectral,saha2023end,stewart2023differentiable}, with unknown cluster counts inferred via Bayesian methods, density estimation, or pruning~\citep{ronen2022deepdpm,ren2020deep,comaniciu2002mean,kong2018recurrent,leiber2021dip,leiber2024dying}.
For images, superpixels~\citep{achanta2012slic,jampani2018superpixel} and region-aware tokenization~\citep{aasan2024spitting,mei2024spformer,aasan2025differentiable} partition images into content-adaptive regions, token aggregation methods~\citep{bolya2022token,aasan2024spitting,mei2024spformer,aasan2025differentiable,khosla2025ren,pu2025linear} compress dense tokens for efficiency, and slot attention~\citep{locatello2020object} and clustering-based transformers~\citep{xu2022groupvit,ke2022learning,liang2023clusterfomer,zeng2024tcformer,braso2025native} organize features into coherent regions.
VECA's cores can likewise be interpreted as cluster seeds that become increasingly semantic with depth (Section~\ref{behavior}). However, rather than merging tokens or iteratively refining slots, VECA preserves every patch token and refines cores and patches jointly in a single feed-forward pass, allowing it to serve as a vision backbone with elastic, linear-cost inference.

%% file: fig_text/1_arch_text.tex
\begin{figure}[t!]
  \centering
\includegraphics[width=1.0\linewidth]{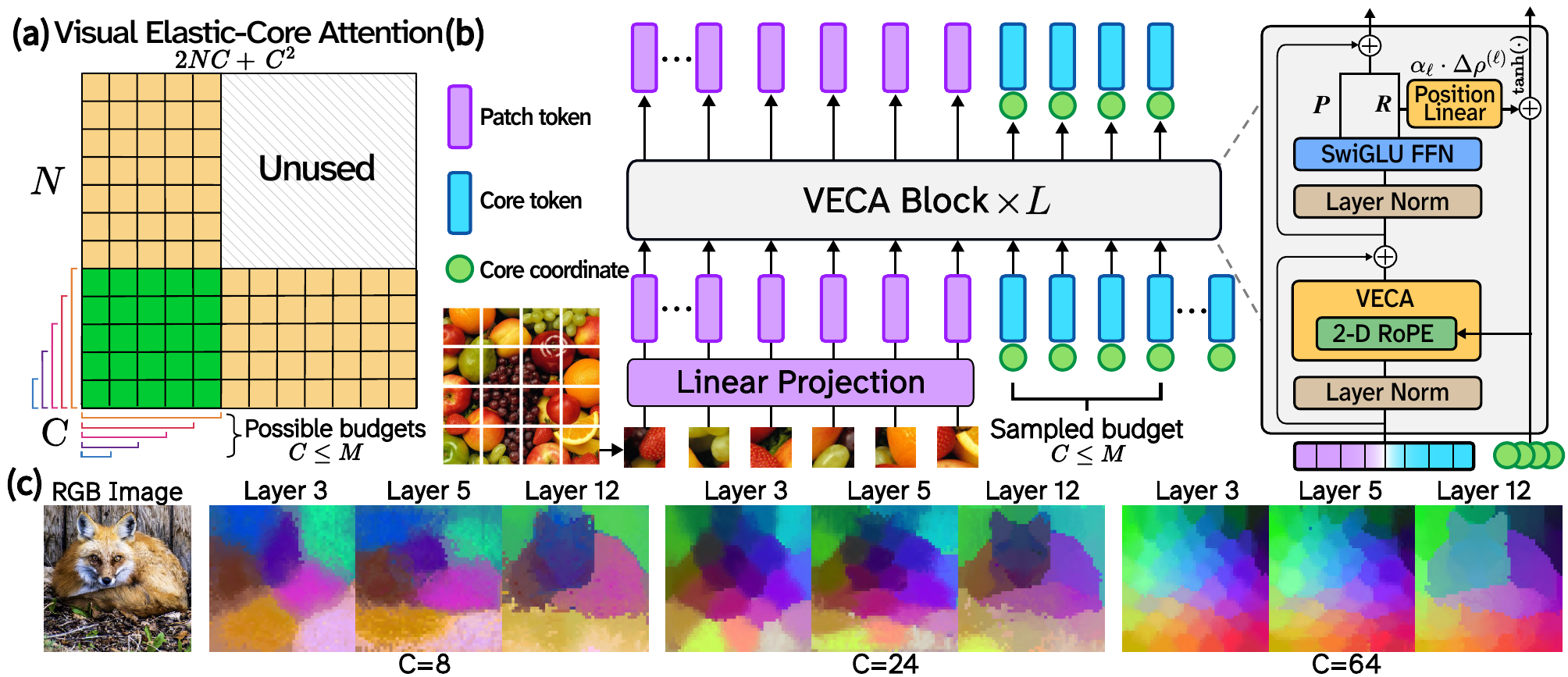}
   \vspace{-4mm}
    \caption{\textbf{Architecture of VECA.}
    \textbf{(a)} Core-periphery attention: cores attend to all tokens, patches attend only to the $C$ active cores, and patch-to-patch attention is unused, yielding $2NC+C^2$ interactions and graph diameter $2$. Budgets $C \le M$ are nested prefixes of a single core bank.
    \textbf{(b)} VECA preserves $N$ dense patch tokens alongside a sampled budget of $C$ core tokens; both are updated at every layer, together with learned core coordinates used in 2-D RoPE.
    \textbf{(c)} Joint UMAP across layers of the cores' attention to patches. Fewer cores summarize broader regions, larger budgets yield finer partitions, and semantic organization emerges with depth.}
  \vspace{-0.4cm}
  \label{fig:1_arch}
\end{figure}

%% file: sections/03_Methods.tex
\input{fig_text/4-1-3-multiresolutions}

\section{Methods}
\label{sec:methods}

In this section, we introduce our \textbf{VECA} (\textbf{V}isual \textbf{E}lastic-\textbf{C}ore \textbf{A}ttention) architecture, illustrated in Figure~\ref{fig:1_arch}. Unlike standard Vision Transformers, which compute full self-attention, VECA organizes attention around a block-sparse core-periphery structure~\citep{rombach2014core,zhang2015identification}. Image patches form the ``periphery,'' and a small set of learned ``cores'' forms a fully connected communication interface. Patch tokens are updated throughout the network but exchange information only through the cores, preserving spatially aligned dense representations that remain coherent as input resolution increases (Figure~\ref{fig:resolution}). Section~\ref{sec:vit_background} reviews the relevant background on Vision Transformers. We then introduce the VECA module and its key properties in Section~\ref{sec:core_attention}, followed by the budget-adaptive training mechanism in Section~\ref{sec:budget_learning}.

\input{table_text/4-1-seg_depth}

\subsection{Background on Vision Transformers}
\label{sec:vit_background}

Here, we go over the fundamental components of Vision Transformers and outline recent advancements. Given an RGB image $\mathcal{I} \in \mathbb{R}^{H\times W \times 3}$, the image is divided into a grid of non-overlapping $N=(H/P)\times(W/P)$ spatial patches, each with width $P$. Each patch is flattened and linearly projected to a latent dimension $D$. Positional embeddings can be added to these projections to retain spatial coordinates, resulting in a 1D sequence $\mathbf{X} \in \mathbb{R}^{N \times D}$.

The sequence $\mathbf{X}$ is then passed through Transformer blocks driven by self-attention. The input is linearly projected into Query ($\mathbf{Q}$), Key ($\mathbf{K}$), and Value ($\mathbf{V}$) matrices. The self-attention mechanism computes contextualized representations by taking a similarity-weighted sum of the values:
\begin{equation}
    \text{Self-Attention}(\mathbf{Q}, \mathbf{K}, \mathbf{V}) =
    \text{softmax}\left(\frac{\mathbf{Q}\mathbf{K}^\top}{\sqrt{d_k}}\right)\mathbf{V},
\end{equation}
where $d_k$ is the dimension of the keys. Typically, an additional \texttt{[CLS]} token and $N_R$ register tokens~\citep{burtsev2020memory,darcet2023vision,chen2025vision} are concatenated to the image tokens, creating a total attention matrix of size $(N+N_R+1)^2$, quadratic in the number of image patches. In practice, models employ Multi-Head Self-Attention, executing this operation in parallel and concatenating the outputs. While the original ViT used learnable positional embeddings that were adapted to higher resolutions via interpolation, recent work has adopted 2-D axial RoPE~\citep{fang2024eva,lu2024unified,lu2024fit,simeoni2025dinov3}. Following DINOv3, we adopt 2-D axial RoPE in our model, where core tokens have content- and layer-dependent learnable positions.

\input{fig_text/multires}

\subsection{Core-Mediated Visual Attention}
\label{sec:core_attention}

The basic building block of VECA is a block-sparse attention operation between active core tokens and patch tokens. Given patch tokens $Z=\{z_1,\ldots,z_N\}$, we additionally maintain an ordered bank of learnable core tokens $R_M=\{r_1,\ldots,r_M\}$, where $M$ denotes the maximum core capacity. For each image, we select an active budget $C \leq M$. The input sequence is formed by concatenating the active core prefix $R_C=R_M[:C]$ with the patch sequence, $X=[R_C;Z]$.
Unlike standard self-attention, patch tokens do not attend directly to all other patch tokens. Instead, core tokens attend to the full sequence, allowing them to aggregate global information, while patch tokens attend only to the active cores, allowing them to receive global context through the core interface. In standard query-key-value notation, the block computes
\begin{align}
    R' &= \mathrm{Attn}(R_C,X,X), \qquad
    Z' = \mathrm{Attn}(Z,R_C,R_C),
\end{align}
which are concatenated to form $X'=[R';Z']$, where we use 2-D axial RoPE.

This gives the attention matrix a core-periphery structure: fully connected core tokens serve as a compact global communication interface, while patch tokens retain spatially aligned dense representations. The resulting attention cost is $(2NC+C^2)$, compared with $N^2$ (ignoring \texttt{[CLS]} and registers) for dense patch self-attention. Therefore, for fixed $C$, VECA attention cost scales linearly in the number of patches $N$ while still allowing information exchange across the full image. This core-periphery structure has graph diameter $2$, compared with diameter $1$ for full self-attention, and therefore requires two blocks for information from one patch to influence another patch through the core interface. As we demonstrate in Section~\ref{experiments}, VECA nevertheless learns competitive representations under this restricted communication structure.

\input{table_text/ablation}
\input{fig_text/4-2-1-sweepcores}

{\bf Learned Core Coordinates.} Each core token is associated with a continuous image- and layer-dependent planar coordinate in $[-1,1]\times[-1,1]$, while patch tokens use fixed image-grid coordinates. Core and patch tokens are modulated with two-dimensional rotary positional encoding. Core coordinates are initialized to cover the image plane and are updated across layers. For layer $\ell$ and core $i$, where $r_i^\ell$ is the feature of core token $i$, $\rho_i^\ell\in\mathbb{R}^2$ is the unconstrained coordinate, $u_i^\ell\in[-1,1]^2$ is the coordinate used for RoPE, $f_\ell$ is a lightweight coordinate head, and $\alpha_\ell$ is a learned layer-wise scalar:
\begin{align}
    \rho_i^{\ell+1} &= \rho_i^\ell+\alpha_\ell f_\ell(r_i^\ell),
    \qquad
    u_i^{\ell+1} = \tanh(\rho_i^{\ell+1}).
\end{align}
This allows each core to maintain both a semantic representation and an evolving spatial position.

\subsection{Budget-Adaptive Representation Learning}
\label{sec:budget_learning}

VECA is trained to support multiple active-core budgets within a single shared model. The ordered core bank allows us to select prefixes of variable size~\citep{rippel2014learning,hu2024matryoshka}. We define a distribution $p_C(\cdot)$ over active core budgets. During training, we sample $C\sim p_C(\cdot)$ and retrieve the active prefix $R_C=R_M[:C]=(r_1,\ldots,r_C)$. Since the sampled cores form a prefix, early cores are active under more budgets and are encouraged to encode the most broadly useful information, while later cores provide additional capacity for higher-fidelity representations.

This nested design encourages a spatially and semantically coarse-to-fine organization of the core representation. With a small core budget, global information must be communicated through a compact set of cores, while the full dense patch representation is still preserved. As the budget increases, additional cores can specialize to more spatially local or semantically specific information. At inference time, a single VECA model can trade accuracy for computation by varying the number of active cores, without retraining. We use the first core token, \(r_1\), as the \texttt{[CLS]} representation.



\input{fig_text/4-1-3-core-token-similarity}

%% file: fig_text/4-1-3-multiresolutions.tex
\begin{figure*}[!t]
    \centering
    \vspace{-1mm}
    \includegraphics[width=0.95\textwidth]{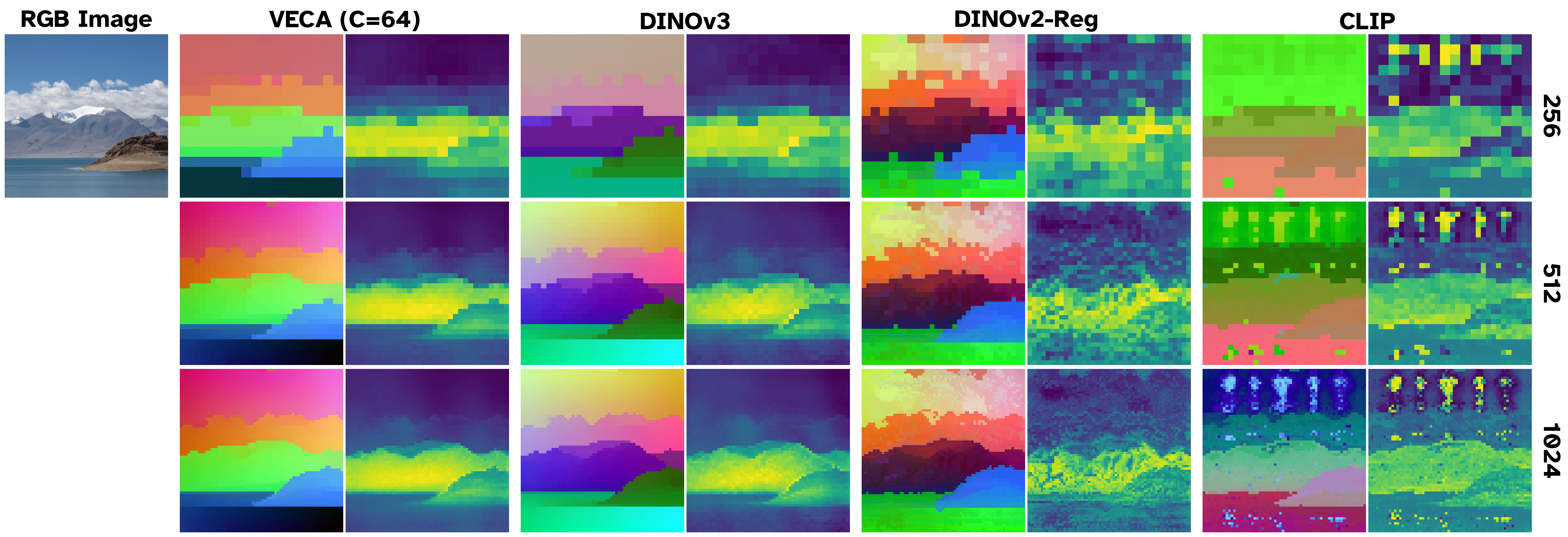}
    \caption{
    \textbf{Learned dense representations across resolutions.}
    UMAP visualizations ({\bf left}) and \texttt{[CLS]}-dense cosine-similarity maps ({\bf right}) show that VECA preserves coherent and semantically organized structure as resolution increases, despite removing direct patch-to-patch attention.
    }
    \vspace{-0.4cm}
    \label{fig:resolution}
\end{figure*}

%% file: table_text/4-1-seg_depth.tex
\begin{table*}[t]
\centering\vspace{-0.3em}
\caption{\textbf{Frozen-backbone evaluation across global and dense visual tasks.}
All models use approximately 86M parameters and $16{\times}16$ patches unless noted; ImageNet-1K top-1 is evaluated at $256$ ($252$ for patch-14 models) and dense tasks at $512$ ($518$). Bold marks the best linear-complexity result.
Controlled models are trained with the same protocol; among them, VECA leads the linear-complexity architectures on six of seven benchmarks and remains close to the full-attention ViT-B/16.}
\label{tab:representation_quality}
\vspace{-2.5mm}

\small
\renewcommand{\arraystretch}{1.08}
\setlength{\tabcolsep}{3pt}

\resizebox{0.95\textwidth}{!}{%
\begin{tabular}{l c c cccc cc}
\toprule
\multirow{2}{*}{Method}
& \multirow{2}{*}{Params (M)}
& \multirow{2}{*}{IN1K $\uparrow$}
& \multicolumn{4}{c}{Segmentation (mIoU $\uparrow$)}
& \multicolumn{2}{c}{Depth (RMSE $\downarrow$)} \\
\cmidrule(lr){4-7}\cmidrule(lr){8-9}
& & & VOC & Context & ADE & Stuff & NYUv2 & KITTI \\
\midrule

\multicolumn{9}{l}{\cellcolor{blue!10}\textbf{\textit{External pretrained ViT references}}} \\
CLIP-B/16~\citep{Radford2021LearningTV_CLIP}
& 86.2 & 79.33 & 72.87 & 45.99 & 34.55 & 36.18 & 0.6176 & 4.3499 \\
DINOv2-reg-B/14~\citep{darcet2023vision}
& 86.6 & 83.44 & 84.23 & 55.02 & 48.92 & 45.42 & 0.3953 & 2.8307 \\
DINOv3-B/16 (Teacher for all models below)
& 85.6 & 83.56 & 87.50 & 57.74 & 51.35 & 48.41 & 0.3684 & 2.6915 \\

\midrule
\multicolumn{9}{l}{\cellcolor{blue!8}\textbf{\textit{Controlled quadratic softmax baseline}}} \\
ViT-B/16~\citep{dosovitskiy2020image}
& 85.7 & 82.30 & 87.60 & 58.01 & 50.94 & 48.49 & 0.3683 & 2.6121 \\

\addlinespace[2pt]
\multicolumn{9}{l}{\cellcolor{blue!8}\textbf{\textit{Latent/inducing-token linear architectures}}} \\
Set Transformer~\citep{lee2019set}
& 85.7 & 80.15 & 85.87 & 56.07 & 48.34 & 46.28 & 0.3958 & 2.8072 \\
Perceiver IO~\citep{jaegle2021perceiverio}
& 86.1 & 79.49 & 79.05 & 51.18 & 42.50 & 41.23 & 0.4418 & 3.5240 \\

\addlinespace[2pt]
\multicolumn{9}{l}{\cellcolor{blue!8}\textbf{\textit{Linear-attention architectures}}} \\
Efficient Attention~\citep{shen2021efficient}
& 85.7 & 81.24 & 86.40 & 56.85 & 49.47 & 47.13 & 0.3836 & 2.7087 \\
Flowformer~\citep{wu2022flowformer}
& 85.7 & 81.11 & 86.69 & 56.60 & 50.01 & 46.94 & 0.3857 & 2.7355 \\
Hydra Attention~\citep{bolya2022hydra}
& 85.7 & 79.12 & 83.59 & 53.46 & 43.60 & 43.20 & 0.4362 & 2.9828 \\
PolaFormer~\citep{meng2025polaformer}
& 85.7 & 81.85 & 86.74 & 57.27 & 50.07 & 47.54 & 0.3792 & \textbf{2.6718} \\

\addlinespace[2pt]
\multicolumn{9}{l}{\cellcolor{blue!8}\textbf{\textit{Landmark-based linear approximation}}} \\
Nystr\"omformer~\citep{xiong2021nystromformer}
& 85.7 & 81.34 & 86.48 & 56.86 & 50.08 & 47.26 & 0.3843 & 2.7949 \\

\midrule
\multicolumn{9}{l}{\textbf{\textit{Ours}}} \\

\rowcolor[gray]{0.93}
\textbf{VECA-B/16 ($C=64$)}
& 85.7
& \textbf{82.18}
& \textbf{87.07}
& \textbf{57.46}
& \textbf{50.69}
& \textbf{47.92}
& \textbf{0.3705}
& 2.7252 \\

\bottomrule

\end{tabular}%
}

\vspace{-0.2cm}
\end{table*}

%% file: fig_text/multires.tex
\begin{figure*}[t]
    \centering
    \includegraphics[
        width=\textwidth
    ]{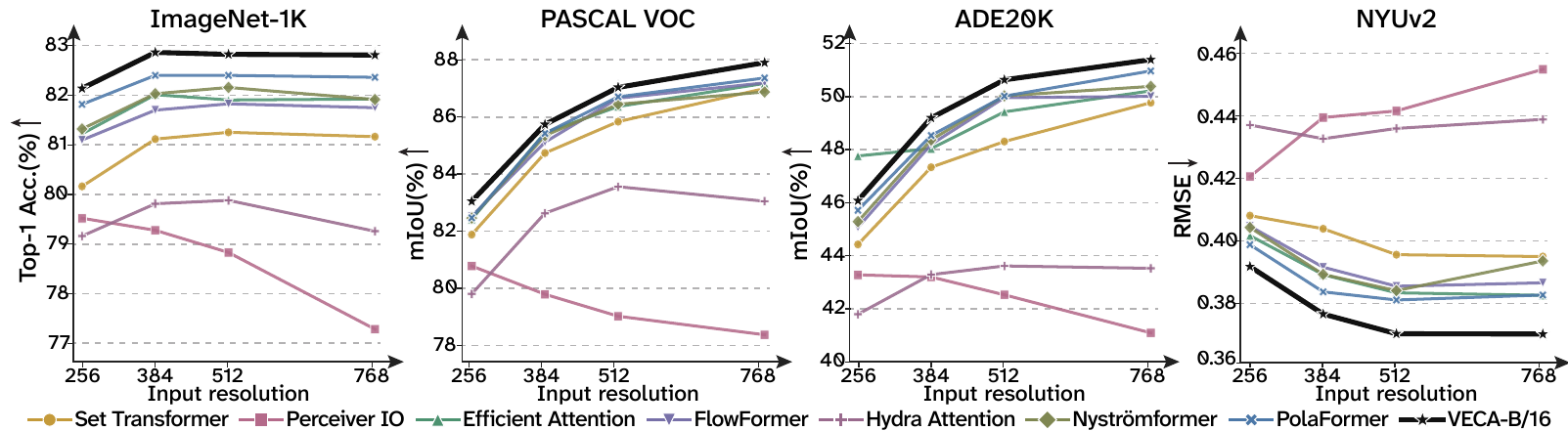}
    \vspace{-5mm}
    \caption{
    \textbf{Representation quality across input resolutions.}
    We compare VECA-B/16 with linear-complexity baselines from $256$ to $768$ on classification, segmentation, and depth. VECA improves with resolution and leads all baselines at $768$, whereas Perceiver IO degrades as resolution increases.
    }
    \vspace{-0.4cm}
    \label{fig:multires}
\end{figure*}

%% file: table_text/ablation.tex
\begin{table*}[t]
\centering
\caption{\textbf{Ablation of VECA's core-periphery design.}
We compare VECA with latent-token architectures that lack dense patch updates and core-patch interaction (Perceiver IO, VECA-CoreOnly) and with a variant using fixed core positions.
FLOPs are for a $256{\times}256$ forward pass; colored values are differences from VECA-B at $C=64$.
Removing both causes the largest drop on every task.} 
\label{tab:ablation}
\vspace{-2.5mm}

\small
\renewcommand{\arraystretch}{1.04}
\setlength{\tabcolsep}{5pt}

\resizebox{0.95\textwidth}{!}{%
\begin{tabular}{lccccc}
\toprule
Variant
& IN1K Acc. $\uparrow$
& ADE mIoU $\uparrow$
& NYUv2 RMSE $\downarrow$
& FLOPs (G)
& Params (M) \\
\midrule

\multicolumn{6}{l}{\cellcolor{blue!8}\textbf{\textit{No patch updates or core-patch interaction}}} \\

Perceiver IO (256 latents)~\citep{jaegle2021perceiverio}
& 79.49\dec{2.69}
& 42.50\dec{8.19}
& 0.4418\inc{0.0713}
& 23.36
& 86.06 \\

VECA-CoreOnly ($C=64$)
& 71.08\dec{11.10}
& 40.49\dec{10.20}
& 0.4584\inc{0.0879}
& 9.39
& 85.73 \\

VECA-CoreOnly ($C=256$)
& 80.30\dec{1.88}
& 49.10\dec{1.59}
& 0.3857\inc{0.0152}
& 27.13
& 85.87 \\

\addlinespace[1pt]
\multicolumn{6}{l}{\cellcolor{blue!8}\textbf{\textit{No core coordinate updates}}} \\

Fixed core coordinates ($C=64$)
& 81.71\dec{0.47}
& 49.80\dec{0.89}
& 0.3771\inc{0.0066}
& 28.01
& 85.71 \\

\midrule
\multicolumn{6}{l}{\textbf{\textit{Full VECA architecture}}} \\

VECA-B ($C=8$)
& 80.45\dec{1.73}
& 44.51\dec{6.18}
& 0.4330\inc{0.0625}
& 22.65
& 85.73 \\

\rowcolor[gray]{0.93}
\textbf{VECA-B ($C=64$)}
& \textbf{82.18}
& \textbf{50.69}
& \textbf{0.3705}
& 28.01
& 85.73 \\

\bottomrule
\end{tabular}%
}
\end{table*}

%% file: fig_text/4-2-1-sweepcores.tex
\begin{figure*}[!t]
\centering
\vspace{-1mm}


\begingroup
\captionsetup{
    type=table,
    width=\textwidth,
    skip=3pt
}

\caption{
\textbf{Ablation of nested elastic training.}
At $C=64$, nested elastic training matches or slightly exceeds fixed-budget training on classification, segmentation, and depth, so elasticity costs no full-budget performance.}
\label{tab:elastic_training}

\begin{minipage}{0.78\textwidth}
\centering
\scriptsize
\renewcommand{\arraystretch}{0.92}
\setlength{\tabcolsep}{4pt}

\begin{tabularx}{\linewidth}{
    @{\hspace{6pt}}
    >{\raggedright\arraybackslash}X
    >{\centering\arraybackslash}p{0.16\linewidth}
    >{\centering\arraybackslash}p{0.16\linewidth}
    >{\centering\arraybackslash}p{0.19\linewidth}
    @{\hspace{6pt}}
}
\toprule

\textbf{Training scheme}
& \textbf{IN1K Acc. $\uparrow$}
& \textbf{ADE mIoU $\uparrow$}
& \textbf{NYUv2 RMSE $\downarrow$} \\

\midrule

Fixed-budget training
& 82.14\dec{0.04}
& 50.66\dec{0.03}
& 0.3713\inc{0.0008} \\

\rowcolor[gray]{0.93}
\textbf{VECA Nested elastic training}
& \textbf{82.18}
& \textbf{50.69}
& \textbf{0.3705} \\

\bottomrule
\end{tabularx}

\end{minipage}
\endgroup

\vspace{2mm}


\includegraphics[
    width=\textwidth
]{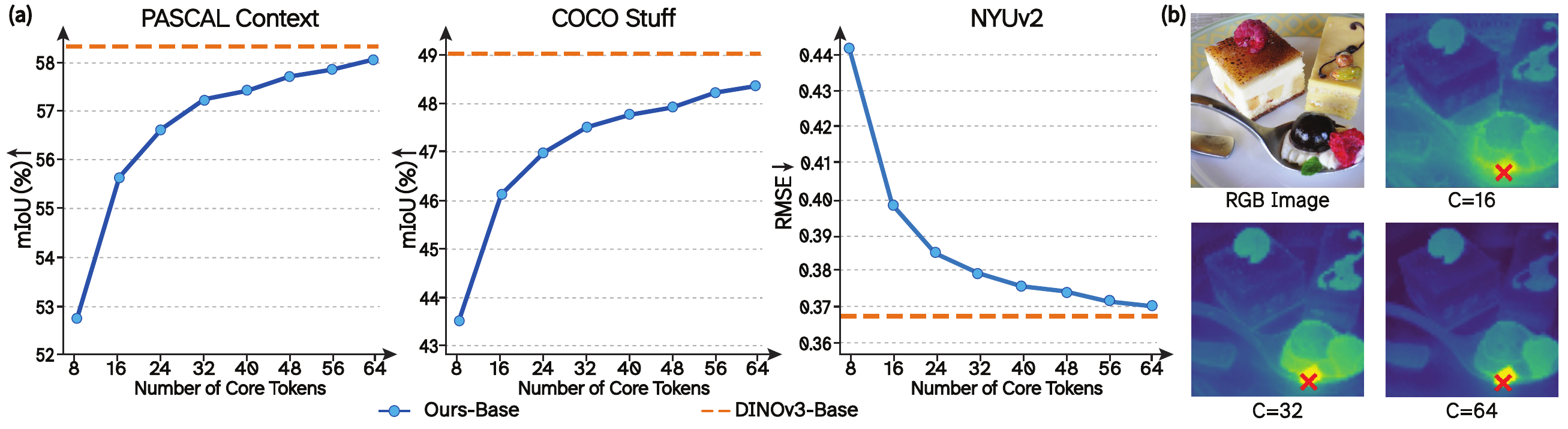}

\captionsetup{
    type=figure,
    width=\textwidth,
    skip=3pt
}

\caption{
\textbf{Elastic computation through the core-token budget.}
\textbf{(a)} Frozen-backbone dense prediction at $768$ px
as the active core budget varies from $C=8$ to $C=64$.
Performance improves smoothly with additional cores across
multiple datasets.
\textbf{(b)} Patch-feature similarity across core budgets.
Larger budgets produce finer and more spatially localized
semantic organization. 
}
\label{fig:sweep}

\vspace{-5mm}
\end{figure*}

%% file: fig_text/4-1-3-core-token-similarity.tex
\begin{figure*}[!t]
    \centering
    \vspace{-2mm}
    \includegraphics[
        width=0.70\textwidth,
        trim={0cm 0cm 0cm 0.0cm},
        clip
    ]{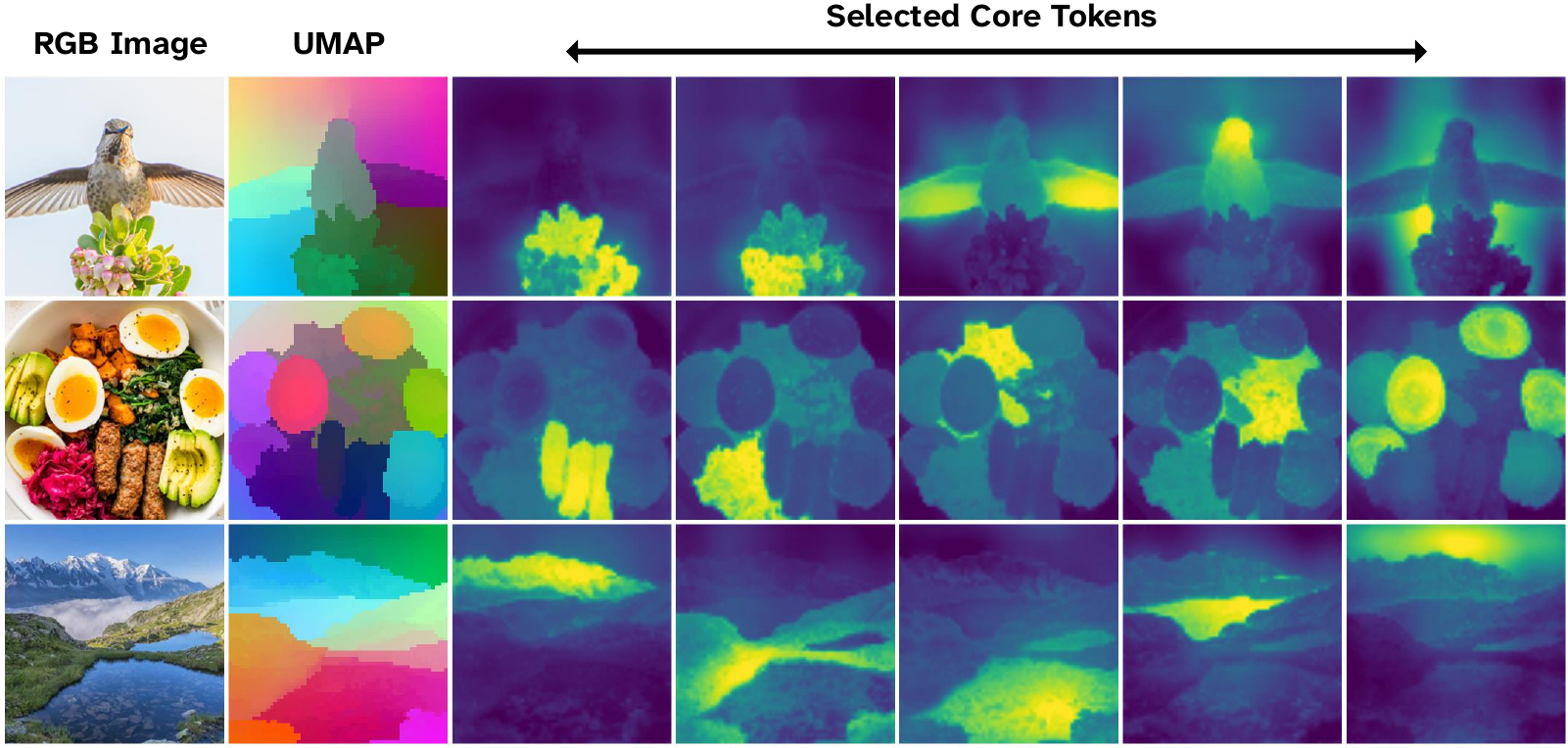}
    \vspace{-1.5mm}
    \caption{
    \textbf{Semantic roles of core tokens.}
    We visualize cosine similarity between selected core tokens and spatial patch features.
    Brighter colors indicate higher similarity.
    Cores localize to different objects and parts, suggesting that the cores develop distinct yet complementary semantic roles.
    }
    \vspace{-4mm}
    \label{fig:core_token}
\end{figure*}

%% file: sections/04_Results.tex
\section{Experiments}
\label{experiments}


In this section, we demonstrate that VECA learns effective global and dense visual representations without direct patch-to-patch interaction. Section~\ref{sec:training_setup} describes the training and evaluation setup; all downstream evaluations freeze the backbone and train only a linear prediction head. Section~\ref{sec:representation_quality} evaluates VECA across image classification and dense prediction tasks and input resolutions. Section~\ref{sec:elasticity} ablates the architectural design choices underlying its performance and examines the compute-accuracy trade-off enabled by varying the active core budget within a single model. Section~\ref{behavior} studies the emergent semantic organization of the core tokens and whether they can be used as compressed reference representations to transfer object labels across video frames.


\subsection{Experimental Setup}
\label{sec:training_setup}

Inspired by prior work~\citep{heinrich2025radiov2}, we train VECA and all controlled baselines by distilling from a frozen DINOv3 teacher on unlabeled Objects365 images~\citep{shao2019objects365}, so differences isolate the attention structure rather than the training recipe. For an input image $\mathcal{I}$ and active core budget $C$, VECA produces a global feature $x_{\mathrm{cls}}^{(C)}$ from the final representation of the first core token and dense features $x_{\mathrm{dense}}^{(C)}$ from the final patch tokens. The teacher provides corresponding targets $x_{\mathrm{cls}}$ and $x_{\mathrm{dense}}$; the remaining cores receive no teacher target. We optimize
\begin{equation}
    \mathcal{L}(\mathcal{I},C)
    =
    \mathcal{L}_{\mathrm{cls}}
    \!\left(x_{\mathrm{cls}}^{(C)}, x_{\mathrm{cls}}\right)
    +
    \lambda
    \mathcal{L}_{\mathrm{dense}}
    \!\left(x_{\mathrm{dense}}^{(C)}, x_{\mathrm{dense}}\right),
\end{equation}
where $\mathcal{L}_{\mathrm{cls}}$ aligns global representations and $\mathcal{L}_{\mathrm{dense}}$ aligns spatial patch representations.

Training proceeds in two stages. In Stage 1, we distill VECA at $256\times256$ resolution while sampling active core budgets. In Stage 2, we continue training with the same objective while adapting the model to resolutions from $\{256, 384, 512, 768\}$. Additional architecture, optimization, and evaluation details are provided in the appendix.

\input{table_text/emergent_behavior}

\subsection{Representation Quality and Resolution Scaling}
\label{sec:representation_quality}

\myparagraph{Representation Quality.}
The results in Table~\ref{tab:representation_quality} show that global and dense representations can be learned without direct patch-to-patch interaction. VECA-B/16 achieves the best performance among the evaluated linear-complexity architectures on six of seven benchmarks, spanning image classification, semantic segmentation, and depth estimation. VECA reaches $82.18\%$ top-1 accuracy on ImageNet-1K~\citep{deng2009imagenet} and $50.69$ mIoU on ADE20K~\citep{ade20k}, improving over the strongest linear-complexity baseline by $0.33$ percentage points and $0.61$ mIoU, respectively. It also remains close to the full-attention DINOv3-B/16 reference. At $512\times512$, VECA with $C=64$ uses only $12.9\%$ of the attention interactions of dense patch self-attention, while remaining within $0.25$ mIoU of the ViT-B/16 reference on ADE20K. These results demonstrate that representations supporting both image-level recognition and dense spatial prediction can be learned without direct all-to-all patch interaction.

\myparagraph{Scaling with Resolution.}
We next study the representation quality under increasing input resolutions from $256\times256$ to $768\times768$. In Figure~\ref{fig:multires}, VECA improves $5.37$ mIoU on ADE20K and decreases $0.0216$ RMSE on NYUv2~\citep{silberman2012nyu}, while yielding a slight performance gain on ImageNet-1K. At $768\times768$, VECA outperforms other evaluated linear-complexity architectures on all four benchmarks. By contrast, Perceiver IO degrades as resolution increases, dropping from $79.49\%$ to $77.19\%$ top-1 accuracy on ImageNet-1K and from $80.81$ to $78.40$ mIoU on PASCAL VOC. These results highlight the robustness of VECA, which adapts to higher resolutions through persistent patch updates and core-to-patch interaction, whereas the fixed latent bottleneck of Perceiver IO increasingly constrains representation quality.


\subsection{Design Ablations and Elastic Inference}
\label{sec:elasticity}
\myparagraph{Architectural Analysis.}
The architectural choices underlying VECA's scaling behavior are examined in Table~\ref{tab:ablation}. VECA-CoreOnly retains dense patch updates only in the first and last layers, while its intermediate layers process core tokens alone. At $C=64$, this variant falls to $71.08\%$ accuracy on ImageNet-1K, $40.49$ mIoU on ADE20K, and $0.4584$ RMSE on NYUv2. Increasing the core budget to $C=256$ substantially improves performance, but it remains below full VECA. Notably, VECA-B with only $C=8$ cores achieves slightly higher ImageNet-1K accuracy than VECA-CoreOnly with $C=256$ ($80.45\%$ versus $80.30\%$), highlighting the importance of persistent patch updates and core-patch interaction. Dense prediction remains more sensitive to the core budget, indicating that additional cores help preserve fine spatial detail. Overall, VECA supports strong global representations even at small budgets, while larger budgets improve dense spatial reasoning. Fixed core coordinates also reduce performance, supporting VECA's learned core-coordinate design.

\textbf{Elastic Computation.} VECA naturally supports nested elastic training through its core structure, with essentially no degradation in full-budget performance. Table~\ref{tab:elastic_training} shows that nested training closely matches fixed-budget training at $C=64$ across classification, segmentation, and depth estimation. At inference time, varying the active core budget $C$ directly adjusts the attention cost, which scales as $(2NC+C^2)$. As shown in Figure~\ref{fig:sweep}, performance improves smoothly with increasing core budget at $768\times768$, with the largest gains occurring at smaller budgets and diminishing improvements thereafter. Increasing $C$ from 8 to 64 improves PASCAL Context by $5.27$ mIoU and reduces NYUv2 by $0.0715$ RMSE. At $C=64$, VECA approaches DINOv3-B across the evaluated dense tasks, while larger budgets produce finer and more spatially localized representations. These results show that a single VECA model can trade computation for accuracy at inference time without retraining.


\input{table_text/video}

\subsection{Emergent Core Structure and Cross-Frame Semantic Transfer}
\label{behavior}

\myparagraph{Emergent Semantic Organization.}
Although core tokens are introduced only as a communication interface, their repeated interaction with persistent dense patch representations leads them to develop structured and complementary visual roles without explicit supervision. Figure~\ref{fig:core_token} shows that different cores respond to distinct objects, parts, and spatial regions. Figure~\ref{fig:feedforward_clustering} further shows that this organization emerges progressively across depth and becomes finer as the active core budget increases. On PartImageNet~\citep{he2022partimagenet}, many-to-one IoU increases from $21.62$ to $57.51$ for $C=8$, and from $22.08$ to $72.98$ for $C=64$. At the larger budget, M2O increases substantially more than O2O, indicating that additional cores specialize to finer semantic regions. Both metrics decrease slightly in the final layer, suggesting a shift toward the final training objective.

\myparagraph{Cross-Frame Label Transferability.}
Because cores specialize to distinct objects and parts, we next test whether cores from one frame can propagate object labels to later frames. On DAVIS-2017~\citep{ponttuset2017davis}, each of the 64 cores of the mask-annotated first frame inherits an object label from its most similar first-frame patches; each later-frame patch then retrieves its label from these labeled cores alone (details in Appendix~\ref{supp_video}). Accurate propagation thus requires each core to represent its object consistently enough to be matched in later frames. We compare against two post hoc references of equal budget, computed from the same first-frame patch features after the forward pass and labeled identically: 64 uniformly sampled patches and 64 $k$-means centroids. As shown in Table~\ref{tab:video-memory}, VECA cores reach 49.6 $\mathcal{J}\&\mathcal{F}$, retaining 81.0\% of the dense baseline with $25\times$ fewer tokens and outperforming both references, including $k$-means (44.7). In contrast, VECA-CoreOnly reaches only 1.3 despite using 256 cores: without persistent patch updates and core-patch interaction across depth, its cores fail to retain the object structure needed for cross-frame transfer. Figure~\ref{fig:video-memory-qual} shows that VECA cores recover label maps close to the dense reference.

%% file: table_text/emergent_behavior.tex
\begin{figure*}[!t]
\centering
\vspace{-2.5mm}


\begin{minipage}[t]{0.64\textwidth}
    \vspace{0pt}
    \centering

    \includegraphics[
        width=\linewidth
    ]{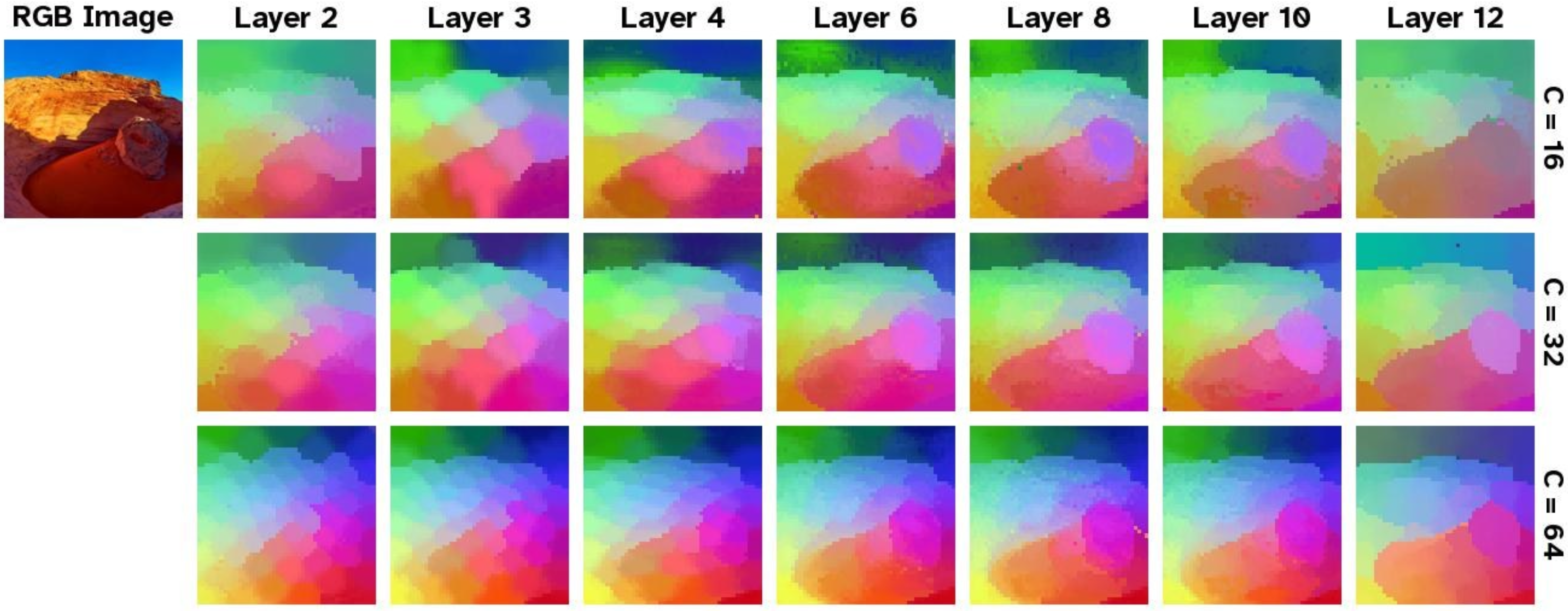}

    \captionsetup{
        type=figure,
        font=small,
        width=\linewidth,
        skip=2pt
    }
    \caption{
        \textbf{Emergence of semantic core organization across depth.}
        Joint UMAP embeddings of the cores' attention to patches reveal increasingly coherent and semantic organization across depth, while larger core budgets yield finer semantic partitions.
    }
    \label{fig:feedforward_clustering}
\end{minipage}
\hspace{0.012\textwidth}
\begin{minipage}[t]{0.33\textwidth}
    \vspace{0pt}
    \centering

    \includegraphics[
        width=\linewidth
    ]{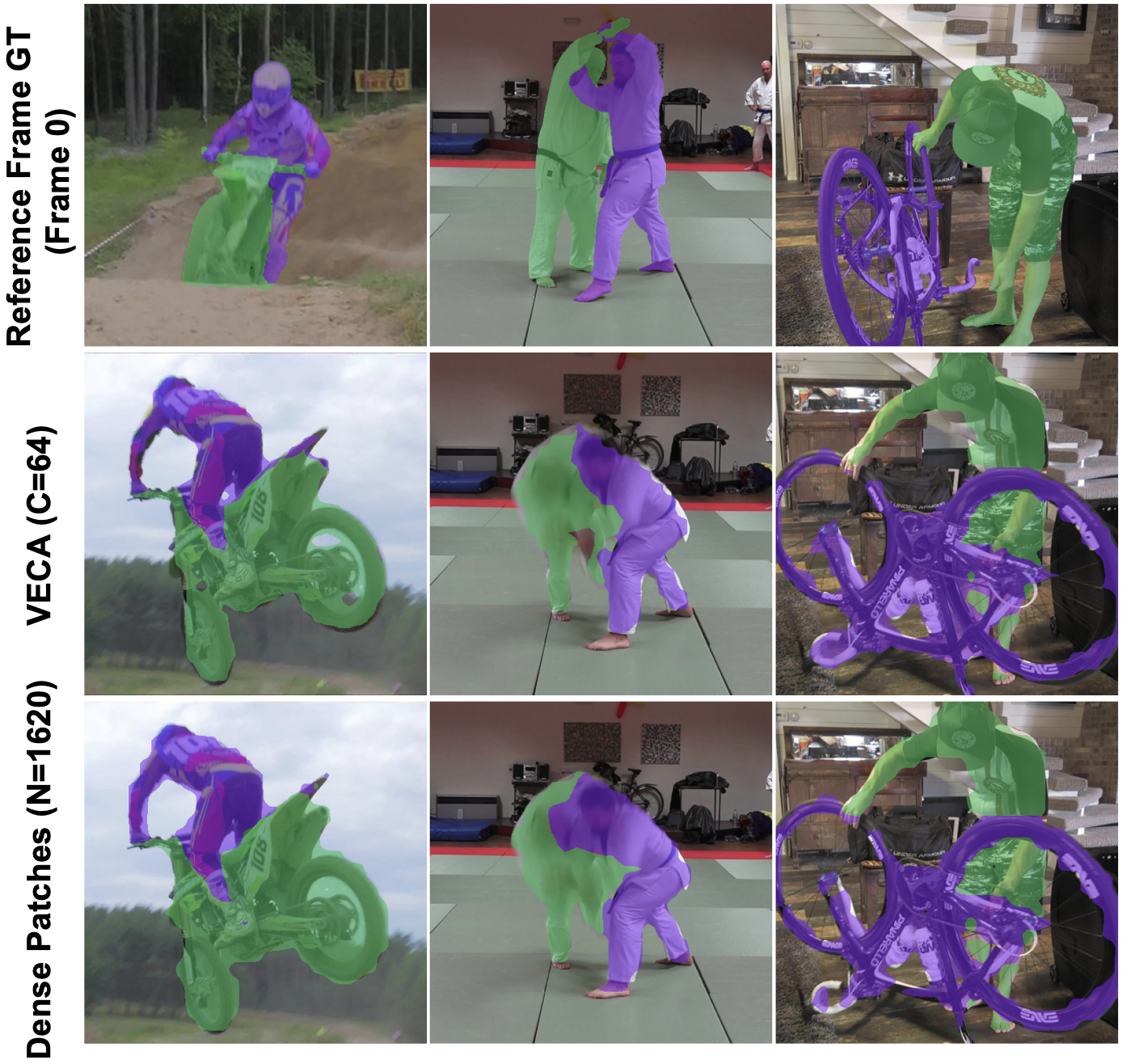}

    \captionsetup{
        type=figure,
        font=small,
        width=\linewidth,
        skip=2pt
    }
    \caption{
        \textbf{Cross-frame semantic transferability using VECA cores.}
    }
    \label{fig:video-memory-qual}
\end{minipage}

\vspace{1mm}


\captionsetup{
    type=table,
    font=small,
    width=\textwidth,
    skip=3pt
}

\caption{
    \textbf{Quantitative evaluation of semantic core organization.}
    On the PartImageNet test set, many-to-one IoU (M2O) allows several cores per part, while one-to-one IoU (O2O) enforces a one-to-one matching. Colored values denote changes from Layer 1.
    Both metrics rise substantially with depth, showing unsupervised semantic organization; at $C=64$, M2O grows far more than O2O, as additional cores partition parts more finely.
}
\label{tab:partimagenet_core_organization}


\small
\renewcommand{\arraystretch}{1.08}
\setlength{\tabcolsep}{5.2pt}

\resizebox{0.95\textwidth}{!}{%
\begin{tabular}{clccccccc}
\toprule
Budget
& Metric
& L1
& L2
& L4
& L6
& L8
& L10
& L12 \\
\midrule

\multirow{2}{*}{$C=8$}
& Many-to-one (M2O) $\uparrow$
& 21.62
& 23.88\inc{2.26}
& 31.53\inc{9.91}
& 45.99\inc{24.37}
& 55.01\inc{33.39}
& 57.51\inc{35.89}
& 52.22\inc{30.60} \\

& One-to-one (O2O) $\uparrow$
& 20.14
& 22.11\inc{1.97}
& 25.83\inc{5.69}
& 34.07\inc{13.93}
& 40.86\inc{20.72}
& 43.03\inc{22.89}
& 38.81\inc{18.67} \\

\addlinespace[2pt]
\midrule
\addlinespace[1pt]

\multirow{2}{*}{$C=64$}
& Many-to-one (M2O) $\uparrow$
& 22.08
& 27.37\inc{5.29}
& 41.02\inc{18.94}
& 64.07\inc{41.99}
& 68.17\inc{46.09}
& 72.98\inc{50.90}
& 72.47\inc{50.39} \\

& One-to-one (O2O) $\uparrow$
& 17.94
& 21.34\inc{3.40}
& 23.73\inc{5.79}
& 25.11\inc{7.17}
& 26.11\inc{8.17}
& 26.70\inc{8.76}
& 22.53\inc{4.59} \\

\bottomrule
\end{tabular}%
}

\vspace{-3mm}
\end{figure*}

%% file: table_text/video.tex
\begin{table}[t]
\centering
\caption{\textbf{Label propagation from compact reference representations.}
First-frame representations propagate object labels to later frames on the DAVIS-2017 validation split, using the first frame's ground-truth annotation to label reference features. Scores are in \%; Retained is $\mathcal{J}\&\mathcal{F}$ relative to dense patches.
With $25\times$ fewer tokens, VECA cores retain 81\% of dense performance and outperform post hoc references, whereas VECA-CoreOnly cores fail despite $4\times$ more tokens.}
\label{tab:video-memory}
\vspace{-2.5mm}

\scriptsize
\renewcommand{\arraystretch}{1.15}
\setlength{\tabcolsep}{5pt}

\begin{tabular}{l l c c c c c c}
\toprule
& Reference representation
& Tokens
& Compression
& $\mathcal{J}$ Mean $\uparrow$
& $\mathcal{F}$ Mean $\uparrow$
& $\mathcal{J}\&\mathcal{F}$ Mean $\uparrow$
& Retained \\
\midrule

\textit{Full}
& Dense patches
& 1,620 & $1{\times}$
& 59.3 & 63.2 & 61.2 & 100\% \\
\midrule

\multirow{2}{*}{\textit{Post hoc}}
& Uniform sampling
& 64 & $25.3{\times}$
& 35.4 & 35.7 & 35.5 & 58.0\% \\
& K-means centroids
& 64 & $25.3{\times}$
& 44.1 & 45.2 & 44.7 & 73.0\% \\
\midrule

& VECA-CoreOnly ($C{=}256$)
& 256 & $6.3{\times}$
& 1.3 & 1.3 & 1.3 & 2.1\% \\
\rowcolor[gray]{0.93}
\cellcolor{white}\multirow{-2}{*}{\textit{Learned cores}}
& \textbf{VECA cores ($C{=}64$)}
& 64 & $25.3{\times}$
& \textbf{49.1} & \textbf{50.0} & \textbf{49.6} & \textbf{81.0\%} \\

\bottomrule
\end{tabular}

\vspace{-0.4cm}
\end{table}

%% file: sections/05_Conclusion.tex
\section{Conclusion}
\label{sec:conclusion}

We introduce \textbf{VECA}, a core-periphery vision architecture that learns effective global and dense representations without direct patch-to-patch interaction. VECA preserves dense patch representations while enabling global communication through a small, elastic set of learned cores, reducing attention complexity to $(2NC+C^2)$. VECA remains competitive across classification and dense prediction while supporting inference-time compute-accuracy trade-offs through the active core budget. Its continuously updated patch stream, together with core-patch interaction, allows the cores to develop semantic organization that transfers across video frames. More broadly, VECA opens opportunities for video, multi-view, and 3D vision in which cores could mediate communication across frames or views without all-to-all interaction.

%% file: sections/90_appendix_arxiv.tex
\appendix
\section{Technical Appendices and Supplementary Material}
\renewcommand\thefigure{S.\arabic{figure}}
\setcounter{figure}{0}
\renewcommand\thetable{S.\arabic{table}}
\setcounter{table}{0}
\renewcommand\thealgorithm{S.\arabic{algorithm}}
\setcounter{algorithm}{0}

\textbf{\Large Sections}
\begin{enumerate}
    \item Architecture details of VECA (Section~\ref{supp_arch_details})
    \item Training and optimization details (Section~\ref{supp_training_details})
    \item Core-Periphery attention pseudocode (Section~\ref{supp_attention_pseudocode})
    \item Baseline model details (Section~\ref{supp_baseline_details})
    \item Implementation details for downstream tasks (Section~\ref{supp_downstream_details})
    \item Additional multi-resolution evaluation across tasks (Section~\ref{supp_multires_eval})
    \item Additional core-to-patch similarity visualizations (Section~\ref{supp_core_to_patch_vis})
    \item Additional classification results (Section~\ref{supp_more_cls_results})
    \item Additional emergent behavior visualizations (Section~\ref{supp_feedforward_clustering})
    \item Efficiency analysis and computational cost (Section~\ref{supp_efficiency})
    \item Additional elastic computation studies (Section~\ref{supp_ablation})
    \item Cross-frame semantic transferability (Section~\ref{supp_video})
\end{enumerate}
\clearpage


\subsection{Architecture details of VECA}
\label{supp_arch_details}
\input{supp_sections/1_arch_details}
\clearpage

\subsection{Training and optimization details}
\label{supp_training_details}
\input{supp_sections/2_training_details}
\clearpage

\subsection{Core-mediated attention pseudocode}
\label{supp_attention_pseudocode}
\input{supp_sections/3_attention_pseudocode}
\clearpage

\subsection{Baseline model details}
\label{supp_baseline_details}
\input{supp_sections/4_baseline_details}
\clearpage

\subsection{Implementation details for downstream tasks}
\label{supp_downstream_details}
\input{supp_sections/5_downstream_details}
\clearpage

\subsection{Additional multi-resolution evaluation across tasks}
\label{supp_multires_eval}
\input{supp_sections/6_multires_eval}
\clearpage

\subsection{Additional core-to-patch similarity visualizations}
\label{supp_core_to_patch_vis}
\input{supp_sections/10_core_to_patch_vis}
\clearpage

\subsection{Additional classification results}
\label{supp_more_cls_results}
\input{supp_sections/7_more_cls_results}
\clearpage

\subsection{Additional emergent behavior visualizations}
\label{supp_feedforward_clustering}
\input{supp_sections/11_feedforward_clustering}
\clearpage

\subsection{Efficiency analysis and computational cost}
\label{supp_efficiency}
\input{supp_sections/13_efficiency}
\clearpage


\subsection{Additional elastic computation studies}
\label{supp_ablation}
\input{supp_sections/14_ablation}
\clearpage

\subsection{Cross-frame semantic transferability}
\label{supp_video}
\input{supp_sections/15_video}
\clearpage

%% file: supp_sections/1_arch_details.tex

This section specifies the concrete VECA model configurations used in our experiments.
The main paper defines the core-mediated attention mechanism and its motivation; here we report implementation-level architectural settings.

\myparagraph{Model configurations.}
We instantiate VECA at four model sizes: Small, Small+, Base, and Large.
All variants use a \(16\times16\) patch embedding, a maximum core capacity \(M=64\), and nested active-core budgets \(\{8,16,24,32,40,48,56,64\}\). Effectively, we implement a granularity of $8$ cores for nested training. The model scale is controlled by the number of layers, hidden dimension, number of attention heads, and feed-forward expansion ratio.

\begin{table}[h]
\centering
\caption{\textbf{VECA architecture configurations.}
All variants use patch size \(16\times16\), maximum core capacity \(M=64\), and active-core budgets \(\{8,16,24,32,40,48,56,64\}\).}
\label{tab:supp_arch_config}
\begin{tabular}{lcccc}
\toprule
\textbf{Model} & \textbf{Layers} & \textbf{Hidden dim.} & \textbf{Heads} & \textbf{MLP ratio} \\
\midrule
VECA-Small  & 12 & 384  & 6  & 2.67 \\
VECA-Small+ & 12 & 384  & 6  & 4.00 \\
VECA-Base   & 12 & 768  & 12 & 2.67 \\
VECA-Large  & 24 & 1024 & 16 & 2.67 \\
\bottomrule
\end{tabular}
\end{table}

\myparagraph{Parameter counts.}
VECA closely matches the parameter count of the corresponding DINOv3 backbone at each model scale.
The small increase comes from the learnable core tokens, core coordinates, and coordinate-update layers.

\begin{table}[h]
\centering
\caption{\textbf{Parameter counts.}
We compare VECA with the corresponding DINOv3 backbone at each model size.}
\label{tab:supp_param_count}
\begin{tabular}{lrr}
\toprule
\textbf{Model} & \multicolumn{1}{c}{\textbf{VECA}} & \multicolumn{1}{c}{\textbf{DINOv3}} \\
\midrule
Small  & 21.63 M  & 21.59 M \\
Small+ & 28.72 M  & 28.68 M \\
Base   & 85.73 M  & 85.64 M \\
Large  & 303.20 M & 303.08 M \\
\bottomrule
\end{tabular}
\end{table}

\myparagraph{Shared architectural settings.}
Table~\ref{tab:supp_shared_arch_config} summarizes the settings shared across model sizes.
Attention is implemented using PyTorch scaled dot-product attention.

\begin{table}[h]
\centering
\caption{\textbf{Shared architectural settings across VECA variants.}}
\label{tab:supp_shared_arch_config}
\begin{tabular}{ll}
\toprule
\textbf{Component} & \textbf{Setting} \\
\midrule
Patch size & \(16 \times 16\) \\
Input channels & 3 \\
Patch embedding & Strided convolution \\
Patch embedding normalization & None \\
Dropout & 0.0 \\
Maximum core capacity & \(M=64\) \\
Core-token chunk size & 8 \\
Active-core budgets & \(\{8,16,24,32,40,48,56,64\}\) \\
Global representation & First core token \(r_1\) \\
Dense representation & All patch tokens \(Z\) \\
FFN type & SwiGLU \\
Attention implementation & PyTorch scaled dot-product attention (SDPA) \\
\bottomrule
\end{tabular}
\end{table}

\myparagraph{Token and parameter layout.}
The learnable core-token bank is denoted by
\[
R_M=\{r_1,\ldots,r_M\},
\]
where \(M\) is the maximum core capacity.
The bank is stored as chunked parameters, with 8 core tokens per chunk and 8 chunks in total.
For an active budget \(C\), the model activates the first \(C/8\) chunks, yielding the active prefix
\[
R_C=R_M[:C]=(r_1,\ldots,r_C).
\]
Given patch tokens \(Z=\{z_1,\ldots,z_N\}\), the input sequence to each block is ordered as
\[
X=[R_C;Z].
\]
This fixed ordering is used by the attention implementation in Section~\ref{supp_attention_pseudocode}.

\myparagraph{Transformer block.}
Each block uses pre-normalization, core-mediated attention, and a SwiGLU feed-forward network:
\[
X_{\ell}' =
X_{\ell}
+
\mathrm{VECA\_Attn}
\!\left(
\mathrm{LN}(X_{\ell})
\right),
\qquad
X_{\ell+1}
=
X_{\ell}'
+
\mathrm{FFN}
\!\left(
\mathrm{LN}(X_{\ell}')
\right).
\]
The exact implementation is given in Section~\ref{supp_attention_pseudocode}.

\myparagraph{Feed-forward network.}
The feed-forward network uses a gated SwiGLU form:
\[
[u,v]=W_1z,
\qquad
\mathrm{FFN}(z)=W_2\left(\mathrm{SiLU}(u)\odot v\right).
\]
This design is used for all model sizes; VECA-Small+ differs from VECA-Small only by using a larger MLP expansion ratio.

\myparagraph{Core coordinates and positional encoding.}
Each patch token is assigned a fixed two-dimensional coordinate from the patch grid.
Each core token is assigned a learnable two-dimensional coordinate initialized by farthest-point sampling in the normalized image plane.
For core \(i\) at layer \(\ell\), we denote the unconstrained coordinate by \(\rho_i^\ell\in\mathbb{R}^2\) and the bounded coordinate used for RoPE by \(u_i^\ell\in[-1,1]^2\):
\[
u_i^\ell=\tanh(\rho_i^\ell).
\]
Both patch and core tokens use two-dimensional rotary positional encoding.
For layers \(\ell>0\), the core coordinate state is updated using a learned residual:
\[
\Delta \rho_i^\ell
=
W_{\ell}^{\mathrm{pos}} r_i^\ell,
\qquad
\rho_i^\ell
=
\rho_i^{\ell-1}
+
\alpha_{\ell}\Delta \rho_i^\ell,
\qquad
u_i^\ell
=
\tanh(\rho_i^\ell),
\]
where \(r_i^\ell\) is the feature of core token \(i\) at layer \(\ell\), \(W_{\ell}^{\mathrm{pos}}\) is a lightweight coordinate-update head, and \(\alpha_{\ell}\) is a learned scalar initialized to a small value.

\myparagraph{Output representations.}
After the final transformer block, LayerNorm is applied to the full token sequence.
The first final core token is used as the global image representation,
\[
y(x)=r_1^L,
\]
and the final patch tokens are used as the dense representation,
\[
Z^L(x)=\{z_1^L,\ldots,z_N^L\}.
\]
Recent models have adopted randomly augmented embeddings~\citep{kim2023region,ranzinger2024radio}, 2D axial RoPE~\citep{fang2024eva,lu2024unified,lu2024fit,simeoni2025dinov3}, or more sophisticated (sometimes non-commutative) 2D RoPE variants~\citep{heo2024rotary,ostmeier2024liere,schenck2025learning,van2025circular,yu2025comrope}. For this work, we follow DINOv3 with the adoption of 2D axial RoPE, and leave the exploration of other positional encodings to future work.

%% file: supp_sections/2_training_details.tex

This section provides the optimization and training settings used for VECA.
We focus on implementation details that are not specified in the main paper.

\myparagraph{Training stages.}
VECA is trained in two stages.
First, we train the model at \(256\times256\) resolution.
Second, we continue from the single-resolution checkpoint and perform multi-resolution finetuning for \(50{,}000\) steps over resolutions \(\{256,384,512,768\}\).
Both stages use the same frozen-teacher objective and the same nested active-core training scheme.

\begin{table}[h]
\centering
\caption{\textbf{Training stages.}}
\label{tab:supp_training_stages}
\begin{tabular}{lcc}
\toprule
\textbf{Stage} & \textbf{Resolution} & \textbf{Duration} \\
\midrule
Single-resolution training & \(256\times256\) & 135 epochs \\
Multi-resolution finetuning & \(\{256,384,512,768\}\) & 50,000 steps \\
\bottomrule
\end{tabular}
\end{table}

\myparagraph{Training objective.}
For an image \(x\) and sampled active-core budget \(C\), the student produces a global representation
\(y^{(C)}(x)\in\mathbb{R}^{D}\) and dense patch representation
\(Z^{(C)}(x)=\{z_i^{(C)}(x)\}_{i=1}^{N}\), where \(z_i^{(C)}(x)\in\mathbb{R}^{D}\).
The frozen teacher provides corresponding targets \(y^\star(x)\) and
\(Z^\star(x)=\{z_i^\star(x)\}_{i=1}^{N}\).
We optimize
\[
\mathcal{L}(x,C)
=
\mathcal{L}_{\mathrm{global}}
\!\left(
y^{(C)}(x),y^\star(x)
\right)
+
\lambda_{\mathrm{dense}}
\mathcal{L}_{\mathrm{dense}}
\!\left(
Z^{(C)}(x),Z^\star(x)
\right).
\]
The global loss is cosine distance:
\[
\mathcal{L}_{\mathrm{global}}
=
1-
\frac{
\left\langle y^{(C)}(x),y^\star(x)\right\rangle
}{
\left\|y^{(C)}(x)\right\|_2
\left\|y^\star(x)\right\|_2
}.
\]
The dense loss combines patch-wise cosine distance and mean-squared error:
\[
\mathcal{L}_{\mathrm{dense}}
=
\frac{1}{N}
\sum_{i=1}^{N}
\left(
1-
\frac{
\left\langle z_i^{(C)}(x),z_i^\star(x)\right\rangle
}{
\left\|z_i^{(C)}(x)\right\|_2
\left\|z_i^\star(x)\right\|_2
}
\right)
+
\beta_{\mathrm{mse}}
\frac{1}{ND}
\sum_{i=1}^{N}
\left\|
z_i^{(C)}(x)-z_i^\star(x)
\right\|_2^2 .
\]
We set \(\lambda_{\mathrm{dense}}=1.0\), \(\beta_{\mathrm{mse}}=1.0\), and use normalization epsilon \(10^{-6}\).

\myparagraph{Training data.}
VECA is trained on Objects365 images.
For each training image, we apply both the global representation loss and the dense patch-level loss using targets from the frozen teacher.
No labeled class supervision is used during backbone training.

\myparagraph{Teacher models.}
Each VECA model is trained with the corresponding DINOv3 teacher at the same scale.
The teacher is frozen throughout training.
When the teacher includes register tokens, we remove them before constructing dense patch targets.

\begin{table}[h]
\centering
\caption{\textbf{Teacher models used for VECA training.}
All teachers are DINOv3 models from the \texttt{facebook/*-pretrain-lvd1689m} family.}
\label{tab:supp_teacher_models}
\begin{tabular}{llc}
\toprule
\textbf{Student} & \textbf{Teacher} & \textbf{Teacher micro-batch} \\
\midrule
VECA-Small  & ViT-S/16  & 128 \\
VECA-Small+ & ViT-S+/16 & 128 \\
VECA-Base   & ViT-B/16  & 128 \\
VECA-Large  & ViT-L/16  & 64  \\
\bottomrule
\end{tabular}
\end{table}

\myparagraph{Batch size.}
For VECA-Small, VECA-Small+, and VECA-Base, we use a per-rank batch size of 128 during single-resolution training.
For VECA-Large, we use a per-rank micro-batch size of 64 and accumulate gradients for 2 steps.

\begin{table}[h]
\centering
\caption{\textbf{Per-rank batch settings for single-resolution training.}}
\label{tab:supp_batch_settings}
\begin{tabular}{lccc}
\toprule
\textbf{Model} & \textbf{Micro-batch} & \textbf{Grad. accum.} & \textbf{Effective batch} \\
\midrule
VECA-Small  & 128 & 1 & 128 \\
VECA-Small+ & 128 & 1 & 128 \\
VECA-Base   & 128 & 1 & 128 \\
VECA-Large  & 64  & 2 & 128 \\
\bottomrule
\end{tabular}
\end{table}

\myparagraph{Input preprocessing and augmentation.}
For single-resolution training, images are resized to a short side of 288 pixels and cropped to \(256\times256\).
For multi-resolution finetuning, the crop resolution is sampled from \(\{256,384,512,768\}\), and the short side is resized to \(1.125\) times the crop resolution.
Training uses random crop, random horizontal flip, and \texttt{TrivialAugmentWide}.
Validation uses center crop.
All images are normalized with
\[
\mu=(0.485,0.456,0.406),
\qquad
\sigma=(0.229,0.224,0.225).
\]

\myparagraph{Active-core budget sampling.}
During both stages, one active-core budget is sampled at each optimizer step from
\[
C \in \{8,16,24,32,40,48,56,64\}
\]
with sampling weights
\[
(1,1,2,2,3,3,4,4).
\]
The sampled budget is broadcast to all distributed ranks before the forward pass.
Validation uses the full active-core budget unless otherwise specified.

\myparagraph{Optimization.}
We use a two-optimizer setup.
Selected two-dimensional linear weights are optimized with NorMuon~\citep{jordan2024muon,li2025normuon}, while the remaining trainable parameters are optimized with AdamW~\citep{kingma2014adam,loshchilov2017decoupled}. We utilize the Dion library~\citep{ahn2025dion} for all optimizers. The AdamW group includes core tokens, core coordinates, coordinate-update layers, positional scaling parameters, biases, normalization parameters, patch embedding parameters, and selected output projection parameters. All other parameters are optimized with NorMuon.

\begin{table}[h]
\centering
\caption{\textbf{Optimizer and schedule settings.}}
\label{tab:supp_optimizer_settings}
\begin{tabular}{lcc}
\toprule
\textbf{Setting} & \textbf{Single-resolution} & \textbf{Multi-resolution finetuning} \\
\midrule
AdamW learning rate & \(4\times10^{-4}\) & \(8\times10^{-5}\) \\
NorMuon learning rate & \(4\times10^{-4}\) & \(8\times10^{-5}\) \\
Minimum learning rate & \(5\times10^{-5}\) & \(3\times10^{-5}\) \\
Weight decay & 0.015 & 0.01 \\
AdamW epsilon & \(10^{-8}\) & \(10^{-8}\) \\
Schedule & Warmup + cosine & Warmup + cosine \\
Mixed precision & bfloat16 & bfloat16 \\
Training hardware & 6 NVIDIA L40S GPUs & 6 NVIDIA L40S GPUs \\
\bottomrule
\end{tabular}
\end{table}

\myparagraph{NorMuon settings.}
For NorMuon, we use momentum \(\mu=0.95\), second-moment coefficient \(\beta_2=0.95\), Nesterov updates, RMS-normalized AdamW compatible learning-rate adjustment~\citep{liu2025muon}, and cautious weight decay~\citep{chen2025cautious}.
The orthogonalization routine uses five Polar Express~\citep{amsel2025polar} iterations.
We do not use parameter flattening.

\begin{table}[h]
\centering
\caption{\textbf{NorMuon settings.}}
\label{tab:supp_normuon_settings}
\begin{tabular}{ll}
\toprule
\textbf{Setting} & \textbf{Value} \\
\midrule
Momentum \(\mu\) & 0.95 \\
\(\beta_2\) & 0.95 \\
Nesterov & True \\
Learning-rate adjustment & RMS norm \\
Cautious weight decay & True \\
Newton--Schulz / Polar Express iterations & 5 \\
Triton kernels & False \\
Parameter flattening & False \\
\bottomrule
\end{tabular}
\end{table}

\myparagraph{Distributed training and checkpointing.}
All models are trained using PyTorch DistributedDataParallel on 6 NVIDIA L40S GPUs.
The active-core budget is sampled on the main process and broadcast to all ranks before each optimizer step.
We use bfloat16 automatic mixed precision for both student and teacher forward passes.
Validation metrics are averaged across distributed ranks.
During multi-resolution finetuning, validation is performed over the resolution and budget grid, and the best checkpoint is selected according to the averaged validation loss.

%% file: supp_sections/3_attention_pseudocode.tex

This section gives implementation-level code snippets for the VECA attention block.
The model configurations are provided in Section~\ref{supp_arch_details}.
We use \(M\) for the maximum core capacity, \(C\leq M\) for the active number of core tokens, \(N\) for the number of patch tokens, and \(D\) for the hidden dimension.
The ordered core-token bank is
\[
R_M=(r_1,\ldots,r_M).
\]
For an active budget \(C\), the model uses the active prefix
\[
R_C=R_M[:C]=(r_1,\ldots,r_C)\in\mathbb{R}^{C\times D}.
\]
Given patch tokens \(Z\in\mathbb{R}^{N\times D}\), the input sequence is ordered as
\[
X=[R_C;Z]\in\mathbb{R}^{(C+N)\times D}.
\]
The listings below are generalized from the training implementation and omit model-size-specific constants.
In the Python code, the variable \texttt{active\_c} corresponds to the mathematical budget \(C\).

\myparagraph{Core-mediated attention.}
The core-mediated attention layer uses separate query, key, and value projections.
After applying two-dimensional RoPE to queries and keys, active core tokens attend to the full sequence, while patch tokens attend only to the active core-token prefix.

\begin{lstlisting}[style=pythonappendix]
class CoreMediatedAttention(nn.Module):
    def __init__(self, dim: int, num_heads: int, dropout: float = 0.0):
        super().__init__()
        assert dim % num_heads == 0
        self.num_heads = num_heads
        self.head_dim = dim // num_heads

        self.q_proj = nn.Linear(dim, dim, bias=True)
        self.k_proj = nn.Linear(dim, dim, bias=True)
        self.v_proj = nn.Linear(dim, dim, bias=True)
        self.out_proj = nn.Linear(dim, dim, bias=True)

        self.drop_p = float(dropout)
        self.out_drop = nn.Identity() if dropout == 0.0 else nn.Dropout(dropout)

    def forward(self, x: torch.Tensor, rope: Rotary2D,
                coords: torch.Tensor, active_c: int) -> torch.Tensor:
        B, T, D = x.shape
        H, Hd = self.num_heads, self.head_dim
        assert 1 <= active_c < T

        q = self.q_proj(x)
        k = self.k_proj(x)
        v = self.v_proj(x)

        q = q.view(B, T, H, Hd).transpose(1, 2)
        k = k.view(B, T, H, Hd).transpose(1, 2)
        v = v.view(B, T, H, Hd).transpose(1, 2)

        cos, sin = rope.cos_sin(coords)
        q = rope.apply(q, cos.to(q.dtype), sin.to(q.dtype))
        k = rope.apply(k, cos.to(k.dtype), sin.to(k.dtype))

        p = self.drop_p if self.training else 0.0

        # Active core tokens R_C attend to all active cores and all patches.
        y_core = F.scaled_dot_product_attention(
            q[:, :, :active_c, :], k, v,
            dropout_p=p,
            is_causal=False,
        )

        # Patch tokens Z attend only to the active core-token prefix R_C.
        y_patch = F.scaled_dot_product_attention(
            q[:, :, active_c:, :],
            k[:, :, :active_c, :],
            v[:, :, :active_c, :],
            dropout_p=p,
            is_causal=False,
        )

        y = torch.cat([y_core, y_patch], dim=2)
        y = y.transpose(1, 2).contiguous().view(B, T, D)
        y = self.out_proj(y)
        return self.out_drop(y)
\end{lstlisting}

For visualization, the corresponding attention pattern can be written with output rows ordered as \((Z,R)\) and source columns ordered as \((R,Z)\):
\[
A =
\begin{bmatrix}
A_{Z\leftarrow R} & 0 \\
A_{R\leftarrow R} & A_{R\leftarrow Z}
\end{bmatrix}.
\]
The zero block denotes the removed patch-to-patch interaction \(A_{Z\leftarrow Z}\).
This display ordering is only for illustrating the sparsity pattern; the implementation keeps the token sequence ordered as \([R_C;Z]\).

\myparagraph{VECA transformer block.}
Each block applies pre-normalized core-mediated attention followed by a pre-normalized SwiGLU feed-forward network.

\begin{lstlisting}[style=pythonappendix]
class VECABlock(nn.Module):
    def __init__(self, dim: int, num_heads: int,
                 mlp_ratio: float, dropout: float = 0.0):
        super().__init__()
        self.norm_attn = nn.LayerNorm(dim)
        self.attn = CoreMediatedAttention(dim, num_heads, dropout)

        hidden = int(dim * mlp_ratio)
        self.norm_ffn = nn.LayerNorm(dim)
        self.fc1 = nn.Linear(dim, 2 * hidden, bias=True)
        self.fc2 = nn.Linear(hidden, dim, bias=True)
        self.drop = nn.Identity() if dropout == 0.0 else nn.Dropout(dropout)

    def forward(self, x: torch.Tensor, rope: Rotary2D,
                coords: torch.Tensor, active_c: int) -> torch.Tensor:
        x = x + self.attn(
            self.norm_attn(x),
            rope=rope,
            coords=coords,
            active_c=active_c,
        )

        y = self.norm_ffn(x)
        u, v = self.fc1(y).chunk(2, dim=-1)
        y = F.silu(u) * v
        y = self.drop(y)
        y = self.fc2(y)
        y = self.drop(y)

        return x + y
\end{lstlisting}

\myparagraph{Active core selection and coordinate update.}
VECA stores the core-token bank \(R_M\) and the corresponding core-coordinate bank in chunks.
For a budget \(C\), the forward pass activates the first \(C\) core tokens and updates only their coordinates layer by layer.

\begin{lstlisting}[style=pythonappendix]
class VECAEncoder(nn.Module):
    def _build_active_cores(self, batch_size: int, active_c: int):
        n_chunks = active_c // self.core_chunk_size

        cores = torch.cat(
            [self.core_token_chunks[j] for j in range(n_chunks)],
            dim=1,
        ).expand(batch_size, -1, -1)

        core_raw = torch.cat(
            [self.core_coord_chunks[j] for j in range(n_chunks)],
            dim=1,
        ).expand(batch_size, -1, -1)

        return cores, core_raw

    @torch.no_grad()
    def _patch_grid(self, hp: int, wp: int,
                    device: torch.device) -> torch.Tensor:
        y = (torch.arange(hp, device=device).float() + 0.5) / hp * 2.0 - 1.0
        x = (torch.arange(wp, device=device).float() + 0.5) / wp * 2.0 - 1.0
        gy, gx = torch.meshgrid(y, x, indexing="ij")
        return torch.stack([gx, gy], dim=-1).reshape(-1, 2)
\end{lstlisting}

\myparagraph{VECA forward pass.}
The full encoder first computes patch tokens \(Z\), concatenates the active core-token prefix \(R_C\) with the patch sequence, and then applies VECA blocks.
The first final core token is returned as the global representation, and the final patch tokens are returned as dense features.

\begin{lstlisting}[style=pythonappendix]
class VECAEncoder(nn.Module):
    def forward(self, images: torch.Tensor,
                active_c: Optional[int] = None):
        B = images.shape[0]
        active_c = self.num_core_tokens if active_c is None else int(active_c)

        assert 1 <= active_c <= self.num_core_tokens
        assert active_c % self.core_chunk_size == 0

        patches, (hp, wp) = self.patch_embed(images)
        cores, core_raw = self._build_active_cores(B, active_c)

        patch_coords = self._patch_grid(hp, wp, patches.device)
        patch_coords = patch_coords.unsqueeze(0).expand(B, -1, -1)
        patch_coords = self.rope.augment_coords_patch(patch_coords)

        core_coords = torch.tanh(core_raw)
        x = torch.cat([cores, patches], dim=1)

        for layer_idx, block in enumerate(self.blocks):
            if layer_idx > 0:
                active_cores = x[:, :active_c, :]
                delta = self.pos_linears[layer_idx - 1](active_cores)
                alpha = self.pos_alpha[layer_idx - 1].to(delta.dtype)
                core_raw = core_raw + alpha * delta
                core_coords = torch.tanh(core_raw)

            coords = torch.cat([core_coords, patch_coords], dim=1)
            x = block(x, self.rope, coords, active_c)

        x = self.final_norm(x)

        cores = x[:, :active_c, :]
        patches = x[:, active_c:, :]

        global_feature = cores[:, 0, :]
        dense_features = patches
        return global_feature, dense_features
\end{lstlisting}

%% file: supp_sections/4_baseline_details.tex

This section summarizes the usage and implementation of baseline backbones and controlled linear-complexity alternatives. 

We perform rigorous controlled experiments in this work. All controlled linear-complexity alternatives utilize approximately the same number of parameters as our model, is trained on the same data-diet, utilizes the same strong DINOv3 teacher, and undergoes multi-resolution training to improve their accuracy at high resolutions. We also perform best-effort hyperparmeter tuning to improve their accuracy. We show conclusively via these carefully controlled experiments that our method yields results stronger than alternative techniques.

\subsubsection{Vision Backbones}

\myparagraph{Baseline backbones.}
We compare VECA with a set of widely used vision backbones spanning contrastive
vision-language pretraining, self-supervised representation learning, and
recent vision foundation models. All baseline models are evaluated using their
publicly released checkpoints.

\begin{table}[h]
\centering
\small
\setlength{\tabcolsep}{4pt}
\caption{\textbf{Baseline models used in our evaluation.}
The asterisk indicates a resolution-specific checkpoint variant.}
\label{tab:supp_baseline_models}
\begin{tabular}{lll}
\toprule
\textbf{Family} & \textbf{Backbone} & \textbf{Checkpoint} \\
\midrule
CLIP & ViT-B/16 & \texttt{openai/clip-vit-base-patch16} \\
OpenCLIP & ViT-B/16 & \texttt{ViT-B-16/laion2b\_s34b\_b88k} \\
DFN-CLIP & ViT-B/16 & \texttt{apple/DFN2B-CLIP-ViT-B-16} \\
DINOv2 & ViT-B/14 & \texttt{facebook/dinov2-base} \\
DINOv2-Reg & ViT-B/14 & \texttt{facebook/dinov2-with-registers-base} \\
DINOv3 & ViT-B/16 & \texttt{facebook/dinov3-vitb16} \\
SigLIP2$^\ast$ & ViT-B/16 & \texttt{google/siglip2-base-patch16} \\
AM-RADIOv2.5 (C-RADIOv3) & ViT-B/16 & \texttt{nvidia/C-RADIOv3-B} \\
\bottomrule
\end{tabular}
\end{table}

\myparagraph{Model usage.}
For each baseline, we use the official or checkpoint-compatible loading
interface associated with the released model. All backbone parameters are kept
frozen during downstream evaluation. For OpenCLIP and DFN-CLIP, we use the
visual backbone from the released checkpoint. For AM-RADIOv2.5 (C-RADIOv3)~\citep{heinrich2025radiov2},
we use the model-provided external preprocessor together with the released
backbone.

\subsubsection{Controlled Architecture Baselines}

To isolate the effect of the token-interaction architecture, we additionally
construct a quadratic ViT reference and several linear-complexity baselines.
All controlled models use a $16{\times}16$ patch embedding, a
768-dimensional representation, and an approximately 86M-parameter budget.
Their parameters are learned through the same distillation and
multiresolution training protocol used for VECA. Feed-forward dimensions are
adjusted where necessary to match the parameter budget.

Every controlled backbone produces both a global image representation and a
dense patch representation. We therefore apply the same frozen-backbone
evaluation protocol to all architectures.

\myparagraph{Quadratic softmax reference.}
ViT-B/16~\citep{dosovitskiy2020image} is a standard 12-layer transformer with
12-head softmax self-attention, interpolated learned positional embeddings,
and SwiGLU feed-forward layers. Unlike the linear-complexity models below, it
forms a complete token-to-token attention matrix and serves as the
parameter-matched quadratic reference.

\myparagraph{Set Transformer.}
Set Transformer~\citep{lee2019set} uses six Induced Set Attention Blocks (ISABs), each with 64 block-specific learned inducing tokens. Inducing tokens first summarize all patch tokens, after which patches read back from this summary, preserving dense outputs without direct all-to-all patch attention. To further strengthen this baseline, we add fixed 2D Fourier coordinates through a learned projection; this is a vision adaptation because the original method treats its input as an unordered set. A single learned pooling seed produces the global representation.

\myparagraph{Perceiver IO.}
Perceiver IO~\citep{jaegle2021perceiverio} maps patches into 256 learned latents, processes them with 12 self-attention blocks, and decodes them through cross-attention. Fixed 2D Fourier coordinates with learned projections encode both input patches and dense output queries, consistent with the original Perceiver design; the latent tokens themselves have no spatial positions. One learned query produces the global representation, while spatial queries produce dense patch features.

\myparagraph{Efficient Attention.}
Efficient Attention~\citep{shen2021efficient} replaces softmax self-attention
with a dual-softmax factorization. Queries are normalized across feature
channels and keys are normalized across tokens, allowing the multiplication
order to be rearranged so that an $N{\times}N$ attention matrix is never
constructed. We use this operator in all 12 transformer blocks while
preserving the standard CLS and patch-token sequence.

\myparagraph{Flowformer.}
Flowformer~\citep{wu2022flowformer} uses noncausal Flow Attention in all 12
blocks. It applies sigmoid feature maps to queries and keys, then uses
conservation-based normalization, source competition, and sink allocation to
control how information is distributed among tokens. The resulting operation
has linear complexity in sequence length.

\myparagraph{Hydra Attention.}
Hydra Attention~\citep{bolya2022hydra} normalizes queries and keys along the
feature dimension and summarizes the sequence through a globally aggregated
feature-wise context. Each token interacts with this shared context rather
than with every other token individually, yielding linear complexity in the
number of tokens. We replace attention in all 12 transformer blocks with this
operator.

\myparagraph{PolaFormer.}
PolaFormer~\citep{meng2025polaformer} separates positive and negative
query-key interactions so that both similar and opposite-polarity relations
can contribute to the output. It additionally uses learned power
transformations and output gating. A $5{\times}5$ depthwise-convolution branch
provides local spatial information that is otherwise difficult to preserve in
factorized linear attention.

\myparagraph{Nystr\"omformer.}
Nystr\"omformer~\citep{xiong2021nystromformer} approximates softmax attention
using 64 landmarks formed by averaging contiguous token groups. Interactions
between the full token sequence and these landmarks approximate the complete
attention matrix. We use six Newton-Schulz iterations to approximate the
landmark-matrix pseudoinverse and include the method's depthwise residual
convolution.

%% file: supp_sections/5_downstream_details.tex

This section provides implementation details for downstream evaluation.
We separate the classification protocol from dense prediction protocols.

\subsubsection{Image Classification}
\label{supp_cls_details}

\myparagraph{Linear probing protocol.}
For image classification benchmarks, we freeze the visual backbone and train only a linear classification head on top of the extracted global feature.
For VECA, the global feature is the first final core token.
For each baseline, we use the corresponding global visual representation defined by its released implementation.
The backbone is kept in evaluation mode throughout linear probing.

\myparagraph{Optimization.}
The linear head is trained with SGD using momentum \(0.9\), learning rate \(0.05\), and weight decay \(0\).
We train for \(12{,}500\) steps with batch size 128 and a cosine learning-rate schedule.
Validation is performed every \(1{,}250\) steps, and we keep the checkpoint with the best validation top-1 accuracy.

\myparagraph{Resolution and preprocessing.}
For classification benchmarks, all backbones with patch size 16 are evaluated with a \(256\times256\) crop, and those with patch size 14 (DINOv2, DINOv2-Reg) with a \(252\times252\) crop.
Before cropping, we resize the short side to \(1.125\) times the crop size.
We use the checkpoint-compatible normalization for each backbone, and interpolate absolute positional embeddings to the input resolution where needed.

\myparagraph{Evaluation metric.}
We report top-1 accuracy for classification benchmarks.

\subsubsection{Dense Prediction}
\label{supp_dense_details}



\myparagraph{Linear probing protocol.}
For semantic segmentation and depth estimation benchmarks, we train a linear prediction head while keeping the backbone frozen. The entire evaluation pipeline is implemented based on the MMSegmentation framework~\citep{mmseg2020}. For both VECA and all baselines, we extract dense patch features and append the \textbf{CLS} token to each patch representation to enrich the feature representations~\citep{bhat2021adabins}. For SigLIP 2~\citep{tschannen2025siglip}, which does not include a \textbf{CLS} token, we directly use patch features for depth evaluation. All backbones remain in evaluation mode throughout linear probing.

\myparagraph{Optimization.} 
The linear head is trained using the AdamW optimizer with betas \((0.9, 0.999)\), a learning rate of \(0.01\), and a weight decay of \(0.0001\), together with a polynomial learning rate scheduler. The number of training steps depends on the benchmark. We use \(20{,}000\) steps for PASCAL VOC 2012 (VOC), \(40{,}000\) for PASCAL Context-60 (Context)~\citep{context} and  ADE20K-150 (ADE), \(80{,}000\) for COCO-Stuff (Stuff)~\citep{caesar2018coco} and COCO-Object (Object)~\citep{cocoobjects}, and \(38{,}400\) for NYUv2 and KITTI~\citep{kitti_geiger2013vision}. The batch size is set to 8. For evaluation, we retain the checkpoint with the best validation mIoU or RMSE, depending on the task.

\myparagraph{Resolution and preprocessing.} 
For dense prediction benchmarks, each backbone is evaluated using a fixed crop resolution of \(512 \times 512\) for models with patch size 16 and \(518 \times 518\) for models with patch size 14. Before cropping, we resize the shorter side to 512 or 518 and apply random cropping during training. For CLIP, OpenCLIP~\citep{OpenCLIP_ilharco_gabriel_2021_5143773}, and DFN-CLIP~\citep{fang2023data_dfn} models with absolute positional embeddings, we interpolate the positional embeddings to match the input resolution.

\myparagraph{Evaluation metrics.}
We report mIoU for semantic segmentation benchmarks and RMSE for depth estimation benchmarks.




%% file: supp_sections/6_multires_eval.tex

This section provides additional multi-resolution evaluation against state-of-the-art vision backbones.
We consider image classification, dense prediction, and qualitative comparisons.

\subsubsection{Image Classification}
\label{supp_multires_cls}

\myparagraph{Protocol.}
We evaluate VECA, DINOv3, and DINOv2 across model sizes and input resolutions.
VECA and DINOv3 use resolutions \(\{256,384,512,768\}\), while DINOv2 uses the nearest patch-aligned resolutions \(\{252,392,518,770\}\) for its patch size of 14.
All backbones are frozen, and a separate linear head is trained for each model and resolution following Section~\ref{supp_cls_details}.

\begin{table}[h]
\centering
\caption{\textbf{Multi-resolution classification.}
Each cell reports ImageNet-1K top-1 accuracy.
Headers show the VECA/DINOv3 resolution, with the corresponding DINOv2 resolution in parentheses.}
\label{tab:supp_multires_cls}
\small
\setlength{\tabcolsep}{6pt}
\renewcommand{\arraystretch}{1.08}
\begin{tabular}{lcccc}
\toprule
\textbf{Model} &
\textbf{256 (252)} &
\textbf{384 (392)} &
\textbf{512 (518)} &
\textbf{768 (770)} \\
\midrule
DINOv3-S  & 77.31 & 78.43 & 78.81 & 78.55 \\
DINOv3-S+ & 79.95 & 80.77 & 81.23 & 81.19 \\
DINOv3-B  & 83.56 & 84.14 & 84.15 & 84.08 \\
DINOv3-L  & 86.40 & 86.67 & 86.67 & 86.74 \\
\midrule
DINOv2-S  & 78.77 & 79.76 & 79.94 & 79.41 \\
DINOv2-B  & 82.45 & 83.02 & 83.34 & 82.81 \\
DINOv2-L  & 84.58 & 84.79 & 84.74 & 84.41 \\
\midrule
VECA-S    & 74.63 & 75.39 & 75.74 & 75.42 \\
VECA-S+   & 77.03 & 77.58 & 77.82 & 77.50 \\
VECA-B    & 82.18 & 82.92 & 82.89 & 82.87 \\
VECA-L    & 84.51 & 84.67 & 84.73 & 84.53 \\
\bottomrule
\end{tabular}
\end{table}

\myparagraph{Summary.}
Table~\ref{tab:supp_multires_cls} shows that VECA remains competitive with DINOv2 and DINOv3 across resolutions.
At Base scale, VECA-B retains \(98.4\%{-}98.6\%\) of DINOv3-B ImageNet-1K accuracy.
At Large scale, VECA-L retains \(97.5\%{-}97.8\%\) on ImageNet-1K.

\subsubsection{Dense Prediction}
\label{supp_multires_dense}

\myparagraph{Protocol.}
We compare VECA and DINOv3 on multi-resolution semantic segmentation and monocular depth estimation, as DINOv3 provides the strongest dense prediction baseline.
Both are evaluated at resolutions \(\{256,384,512,768\}\).
All backbones are frozen, and a separate linear head is trained for each model and resolution following Section~\ref{supp_dense_details}.

\begin{table*}[h]
\centering
\caption{\textbf{Multi-resolution semantic segmentation.}
Each cell reports mIoU on \((\mathrm{Context}, \mathrm{ADE}, \mathrm{Stuff}, \mathrm{Object})\).}
\label{tab:supp_multires_segmentation}
\scriptsize
\setlength{\tabcolsep}{3pt}
\renewcommand{\arraystretch}{1.08}
\resizebox{0.95\textwidth}{!}{
\begin{tabular}{lcccc}
\toprule
\textbf{Model} &
\textbf{256} &
\textbf{384} &
\textbf{512} &
\textbf{768} \\
\midrule
DINOv3-S  & 51.78 / 41.28 / 41.74 / 54.60 & 54.06 / 44.55 / 43.82 / 57.83 & 55.01 / 46.01 / 44.67 / 59.49 & 55.10 / 46.62 / 45.04 / 59.83 \\
DINOv3-S+ & 51.81 / 41.67 / 42.13 / 55.54 & 54.87 / 46.44 / 44.95 / 59.51 & 55.95 / 48.57 / 45.86 / 61.12 & 56.38 / 49.53 / 46.27 / 61.69 \\
DINOv3-B  & 55.02 / 46.61 / 45.87 / 60.67 & 57.07 / 49.53 / 47.97 / 63.67 & 57.74 / 51.35 / 48.41 / 64.65 & 58.40 / 52.37 / 49.04 / 65.78 \\
DINOv3-L  & 56.50 / 49.75 / 47.36 / 62.55 & 58.33 / 52.66 / 48.93 / 65.13 & 59.13 / 54.56 / 49.71 / 66.53 & 59.70 / 55.52 / 50.12 / 67.28 \\
\midrule
VECA-S    & 50.98 / 40.61 / 40.49 / 52.91 & 52.85 / 43.19 / 42.86 / 55.92 & 53.46 / 44.82 / 43.45 / 57.46 & 53.87 / 45.28 / 43.77 / 57.96 \\
VECA-S+   & 51.19 / 41.07 / 41.32 / 54.16 & 54.02 / 44.78 / 43.53 / 57.50 & 54.68 / 46.55 / 44.33 / 58.76 & 55.02 / 47.03 / 44.78 / 59.08 \\
VECA-B    & 54.91 / 46.09 / 45.43 / 59.71 & 56.79 / 49.24 / 47.27 / 62.69 & 57.46 / 50.69 / 47.92 / 62.58 & 58.05 / 51.46 / 48.37 / 64.14 \\
VECA-L    & 56.66 / 50.05 / 47.36 / 62.28 & 58.61 / 52.99 / 48.92 / 64.94 & 59.64 / 54.17 / 49.53 / 66.19 & 60.23 / 55.54 / 50.06 / 66.85 \\
\bottomrule
\end{tabular}
}
\end{table*}

\begin{table*}[h]
\centering
\caption{\textbf{Multi-resolution depth estimation.}
Each cell reports RMSE on \((\mathrm{NYUv2}, \mathrm{KITTI})\).}
\label{tab:supp_multires_depth}
\scriptsize
\setlength{\tabcolsep}{6pt}
\renewcommand{\arraystretch}{1.08}
\begin{tabular}{lcccc}
\toprule
\textbf{Model} &
\textbf{256} &
\textbf{384} &
\textbf{512} &
\textbf{768} \\
\midrule
DINOv3-S  & 0.4565 / 3.2402 & 0.4252 / 3.1130 & 0.4169 / 2.9083 & 0.4206 / 2.8943 \\
DINOv3-S+ & 0.4651 / 3.3499 & 0.4225 / 2.9987 & 0.4143 / 2.8127 & 0.4140 / 2.8125 \\
DINOv3-B  & 0.3897 / 2.9636 & 0.3727 / 2.8962 & 0.3684 / 2.6915 & 0.3665 / 2.6840 \\
DINOv3-L  & 0.3663 / 2.9207 & 0.3460 / 2.7836 & 0.3446 / 2.5346 & 0.3445 / 2.4821 \\
\midrule
VECA-S    & 0.4528 / 3.2776 & 0.4360 / 3.2106 & 0.4347 / 3.0130 & 0.4298 / 3.0289 \\
VECA-S+   & 0.4597 / 3.3009 & 0.4316 / 3.1318 & 0.4252 / 2.9878 & 0.4231 / 3.0176 \\
VECA-B    & 0.3920 / 2.9765 & 0.3768 / 2.8976 & 0.3705 / 2.7252 & 0.3704 / 2.7468 \\
VECA-L    & 0.3546 / 2.8792 & 0.3409 / 2.8019 & 0.3379 / 2.5904 & 0.3382 / 2.5634 \\
\bottomrule
\end{tabular}
\end{table*}

\myparagraph{Summary.}
Tables~\ref{tab:supp_multires_segmentation} and~\ref{tab:supp_multires_depth} show that VECA remains competitive with DINOv3 across resolutions and model scales.
Both models generally benefit from higher resolution, while VECA scales linearly with the number of input tokens compared with quadratic self-attention in DINOv3.
At Base scale, VECA-B remains close to DINOv3-B across the evaluated tasks.
At Large scale, VECA-L matches or exceeds DINOv3-L on several benchmarks, including Context, ADE, and NYUv2.
These results highlight VECA's favorable accuracy-computation trade-off at high resolutions.

\clearpage

\subsubsection{Additional Qualitative Comparison}

\begin{figure*}[htbp]
    \centering
    \includegraphics[width=0.95\textwidth]{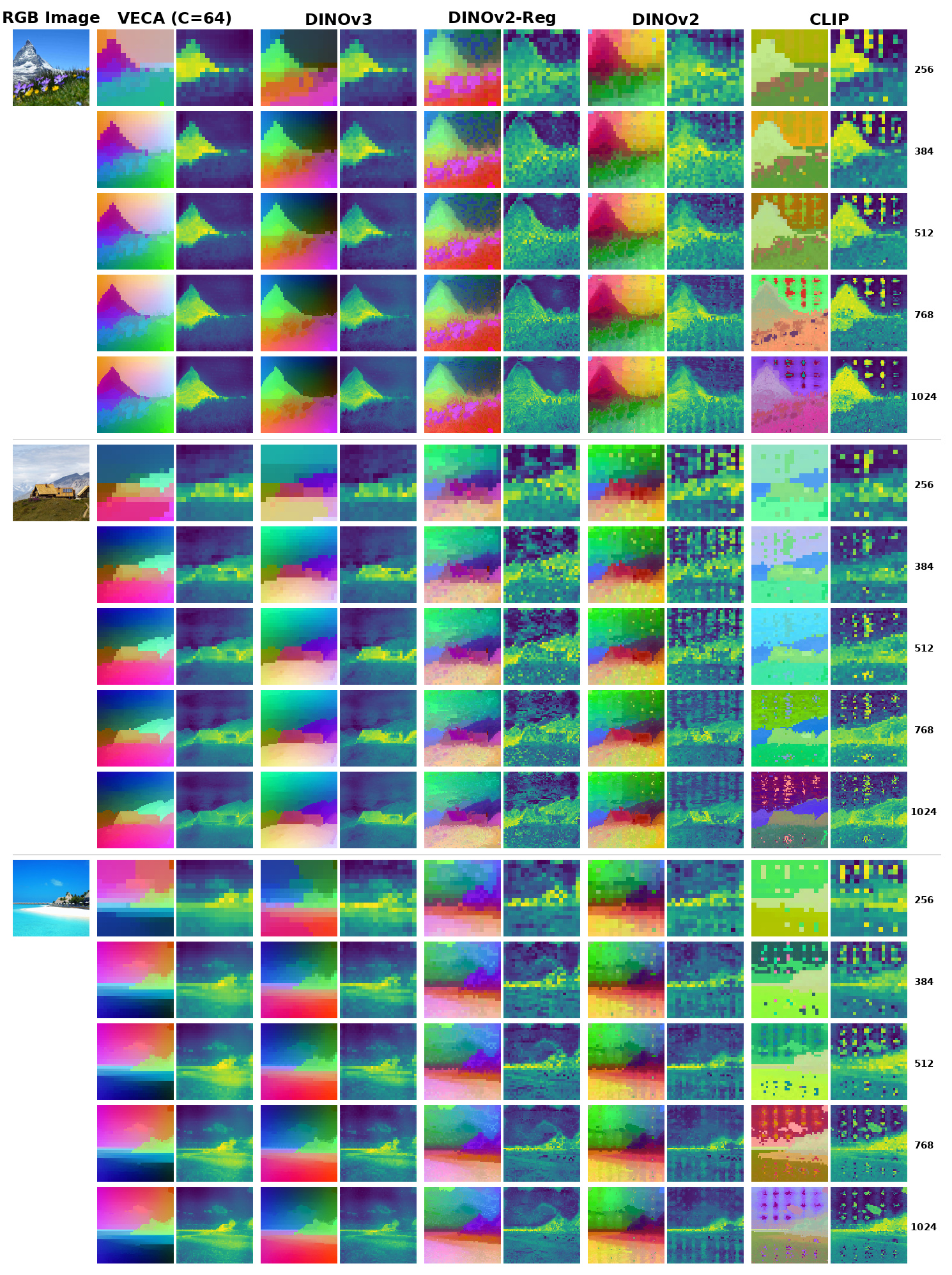}
    \caption{
    \textbf{Additional qualitative multi-resolution comparison.}
    We compare dense patch representations from VECA and baseline ViTs across resolutions.
    VECA preserves coherent object boundaries and smooth semantic regions, while several baselines show more fragmented similarity patterns at higher resolutions.
    }
    \label{fig:multires_qual_supp}
\end{figure*}

\begin{figure*}[htbp]
    \centering
    \includegraphics[width=1\textwidth]{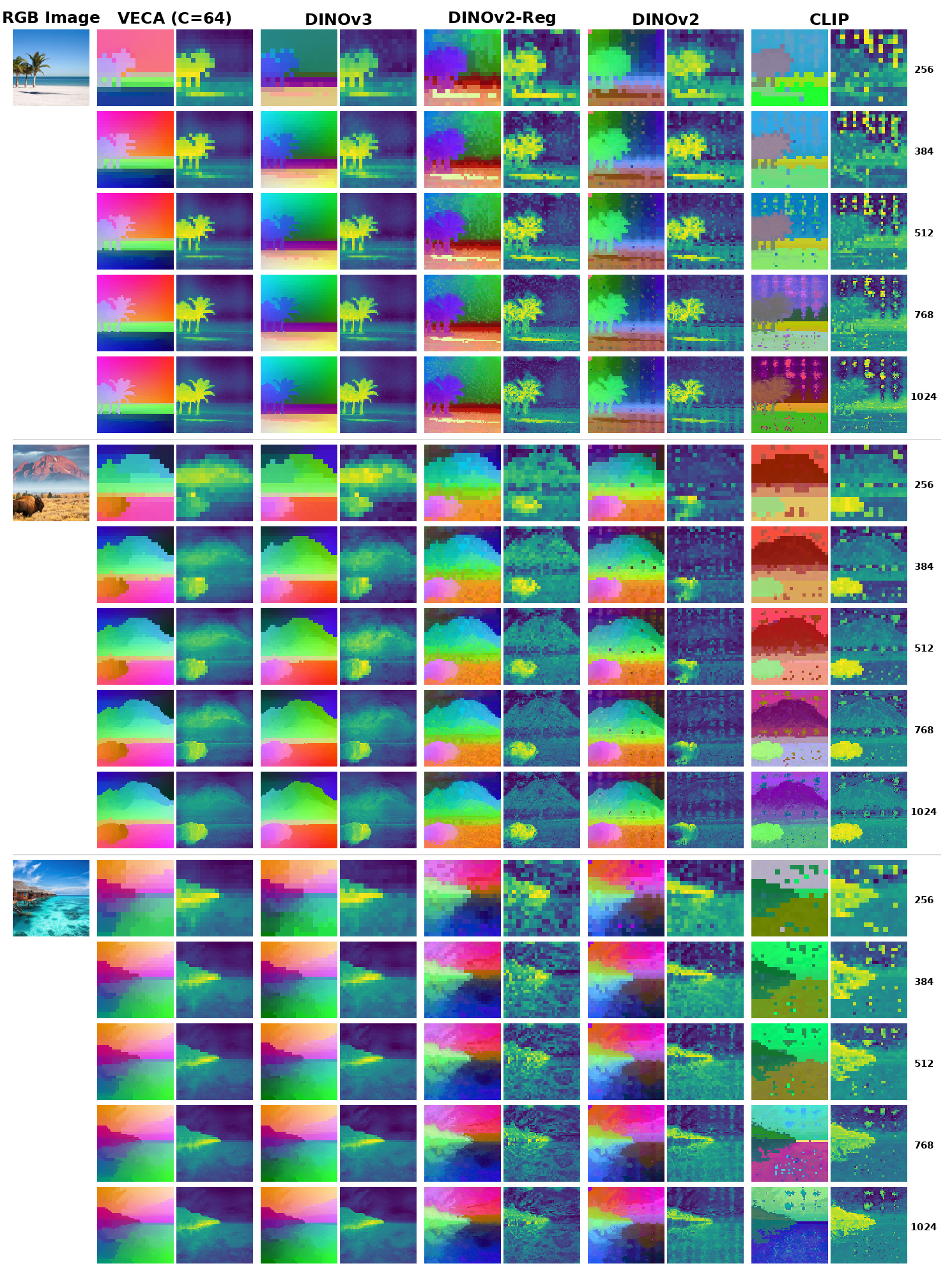}
    \caption{
    \textbf{Additional qualitative multi-resolution comparison.}
    We compare dense patch representations from VECA and baseline ViTs across resolutions.
    VECA preserves coherent object boundaries and smooth semantic regions, while several baselines show more fragmented similarity patterns at higher resolutions.
    }
    \label{fig:multires_qual_supp1}
\end{figure*}

\label{supp_multires_vis}

%% file: supp_sections/10_core_to_patch_vis.tex
\begin{figure*}[htbp]
    \centering
    \includegraphics[width=0.72\textwidth]{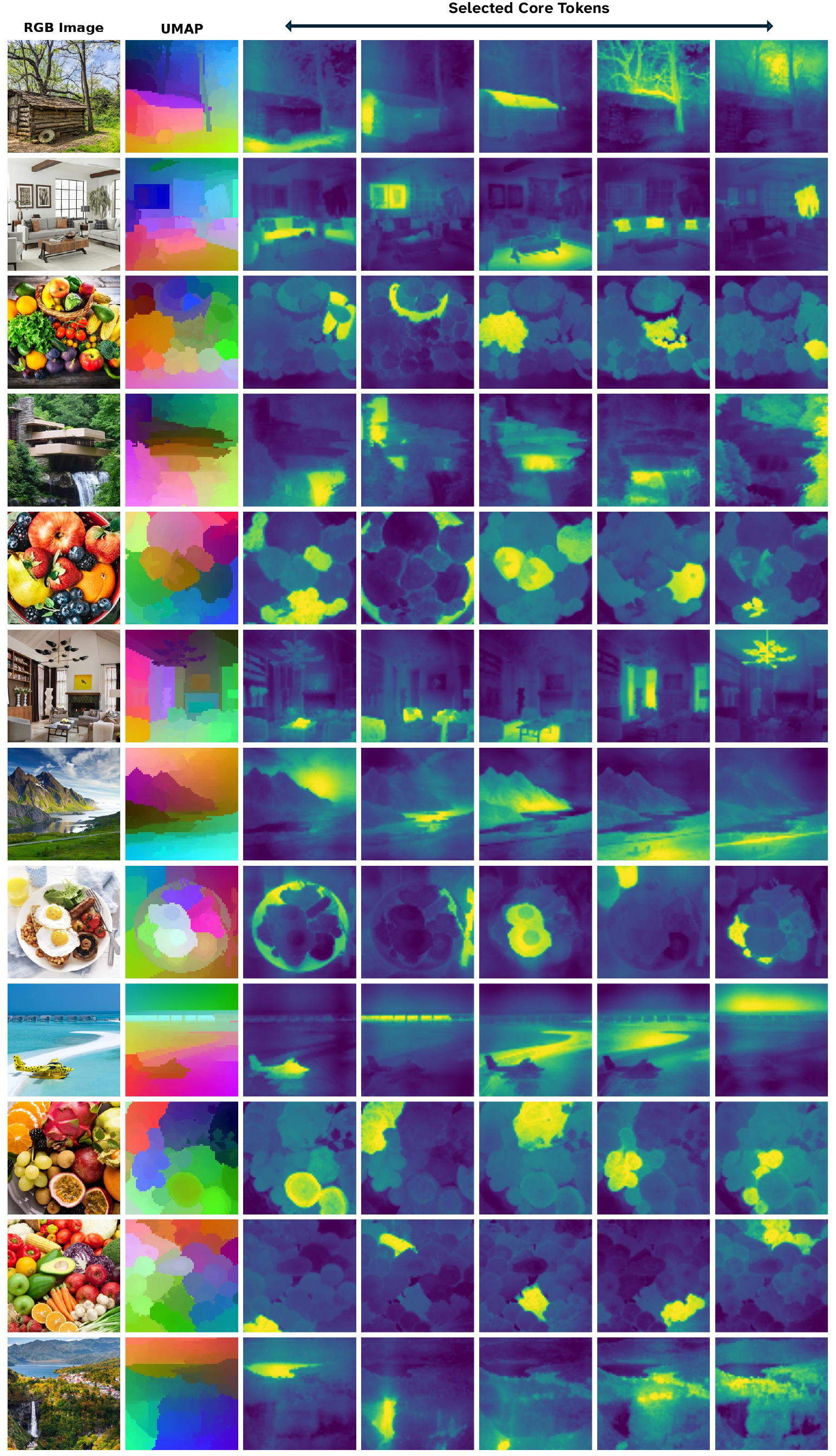}
    \caption{
    \textbf{Additional core-to-patch similarity visualizations.}
    For each RGB image, we show a UMAP projection of dense patch features followed by similarity maps for selected core tokens from our model. 
    Warmer colors indicate image regions with higher cosine similarity to a given core token. 
    The selected cores attend to complementary objects, parts, and spatial regions, illustrating that different core tokens specialize in distinct visual structures while preserving coherent scene-level organization.
    }
    \label{fig:q2p_similarity_supp}
\end{figure*}

%% file: supp_sections/7_more_cls_results.tex

We follow the frozen-backbone linear-probing protocol in
Section~\ref{supp_cls_details}. ImageNetV2~\citep{recht2019imagenetv2} and ImageNet-ReaL~\citep{beyer2020weINreal} reuse the
ImageNet-1K linear head without adaptation, while each remaining dataset
uses an independently trained linear head. The evaluation used $256$ for patch-16 models and $252$ for patch-14 models.

\begin{table*}[h]
\centering
\caption{\textbf{Additional frozen-backbone classification results.}
We report top-1 accuracy. Bold indicates the best result among the
linear-complexity controlled baselines and VECA variants.}
\label{tab:supp_more_cls_results}
\vspace{-2mm}

\scriptsize
\setlength{\tabcolsep}{4.0pt}
\renewcommand{\arraystretch}{1.06}

\resizebox{0.92\textwidth}{!}{%
\begin{tabular}{lcccccccc}
\toprule
\textbf{Model}
& \textbf{IN1K}
& \textbf{V2}
& \textbf{ReaL}
& \textbf{Places}
& \textbf{Food}
& \textbf{Aircraft}
& \textbf{Pets}
& \textbf{Flowers} \\
\midrule

\multicolumn{9}{l}{\textit{External pretrained references}} \\

DINOv3-B/16
& 83.56 & 74.92 & 88.65 & 55.39 & 94.47 & 81.97 & 96.54 & 99.69 \\

DINOv2-Reg-B/14
& 83.44 & 74.75 & 88.18 & 54.81 & 92.87 & 70.78 & 95.97 & 99.67 \\

DINOv2-B/14
& 82.45 & 72.98 & 86.84 & 53.37 & 92.02 & 74.02 & 95.61 & 99.59 \\

SigLIP2-B/16-256
& 81.95 & 73.04 & 87.69 & 56.18 & 94.70 & 62.80 & 92.89 & 97.98 \\

DFN2B-CLIP-B/16
& 81.06 & 71.35 & 86.56 & 55.79 & 93.32 & 66.49 & 93.59 & 98.70 \\

AM-RADIOv2.5-B/16
& 80.53 & 70.60 & 85.78 & 54.13 & 91.64 & 52.18 & 93.54 & 98.52 \\

CLIP-B/16
& 79.33 & 69.31 & 84.93 & 55.27 & 92.42 & 56.65 & 92.72 & 96.36 \\

OpenCLIP-B/16
& 79.18 & 69.05 & 84.72 & 55.35 & 91.33 & 62.89 & 91.91 & 96.50 \\

\midrule
\multicolumn{9}{l}{\textit{Controlled linear-complexity baselines}} \\

Set Transformer
& 80.15 & 70.53 & 86.74 & 55.50 & 90.42 & 67.90 & 95.58 & 98.42 \\

Perceiver IO
& 79.49 & 69.29 & 86.23 & 55.30 & 89.22 & 66.91 & 94.55 & 97.56 \\

Efficient Attention
& 81.24 & 71.80 & 87.49 & 55.66 & 91.54 & 71.50 & 95.56 & 99.04 \\

FlowFormer
& 81.11 & 71.16 & 87.36 & 55.53 & 91.40 & 71.47 & 95.34 & 98.88 \\

Hydra Attention
& 79.12 & 68.42 & 85.96 & 55.42 & 89.15 & 67.33 & 95.07 & 98.00 \\

PolaFormer
& 81.85 & 72.10 & 87.77 & 55.73 & 92.40
& \textbf{74.59} & \textbf{95.94} & 99.32 \\

Nystr\"omformer
& 81.34 & 71.96 & 87.54 & 55.73 & 91.82 & 73.33 & 95.50 & 99.15 \\

\midrule
\multicolumn{9}{l}{\textit{Ours}} \\

\rowcolor[gray]{0.93}
\textbf{VECA-B/16 ($C=64$)}
& \textbf{82.18}
& \textbf{72.25}
& \textbf{87.87}
& \textbf{55.82}
& \textbf{92.44}
& 74.47
& 95.67
& \textbf{99.35} \\

VECA-B/16 ($C=8$)
& 80.45 & 69.18 & 86.45 & 55.30 & 90.04 & 70.72 & 95.12 & 98.98 \\

\bottomrule
\end{tabular}%
}
\end{table*}

\myparagraph{Summary.}
VECA-B/16 outperforms the controlled linear-complexity baselines on six of
eight benchmarks, while PolaFormer is slightly stronger on Aircraft and
Pets~\citep{oxford}. At $C=8$, VECA retains competitive transfer performance, showing
that its inference budget can be reduced substantially while preserving
useful global representations.

%% file: supp_sections/11_feedforward_clustering.tex
In this section, we provide additional visualizations of how VECA organizes patch representations through its learned core-token interface.

\subsubsection{Analysis Method}
\label{supp_emergent_behavior_method}

To visualize how the cores organize the image, we look at how each core attends to the patches. At layer \(\ell\), we take core \(c\) as the query and the patches as keys. Instead of raw attention, we use value-weighted attention~\citep{kobayashi2020attention}: the norm of the attention-weighted value vector after the output projection, summed over heads and normalized over all keys. This gives a weight \(s_{cp}^{\ell}\) for each core \(c\) and patch \(p\).

Each patch is then described by a \(C\)-dimensional feature, one entry per active core, which gives a map of shape \(H/P \times W/P \times C\). We embed these features into three dimensions with UMAP, jointly across the selected layers, and display the result as an \(H/P \times W/P \times 3\) RGB image. We omit the first layer, where attention mostly reflects initialization.

\newpage
\subsubsection{Additional Visualizations}
\label{supp_emergent_behavior_visualizations}

\begin{figure*}[htbp]
    \centering
    \includegraphics[
        width=1.0\textwidth,
        trim={0cm 4.5cm 0cm 4cm},
        clip
    ]{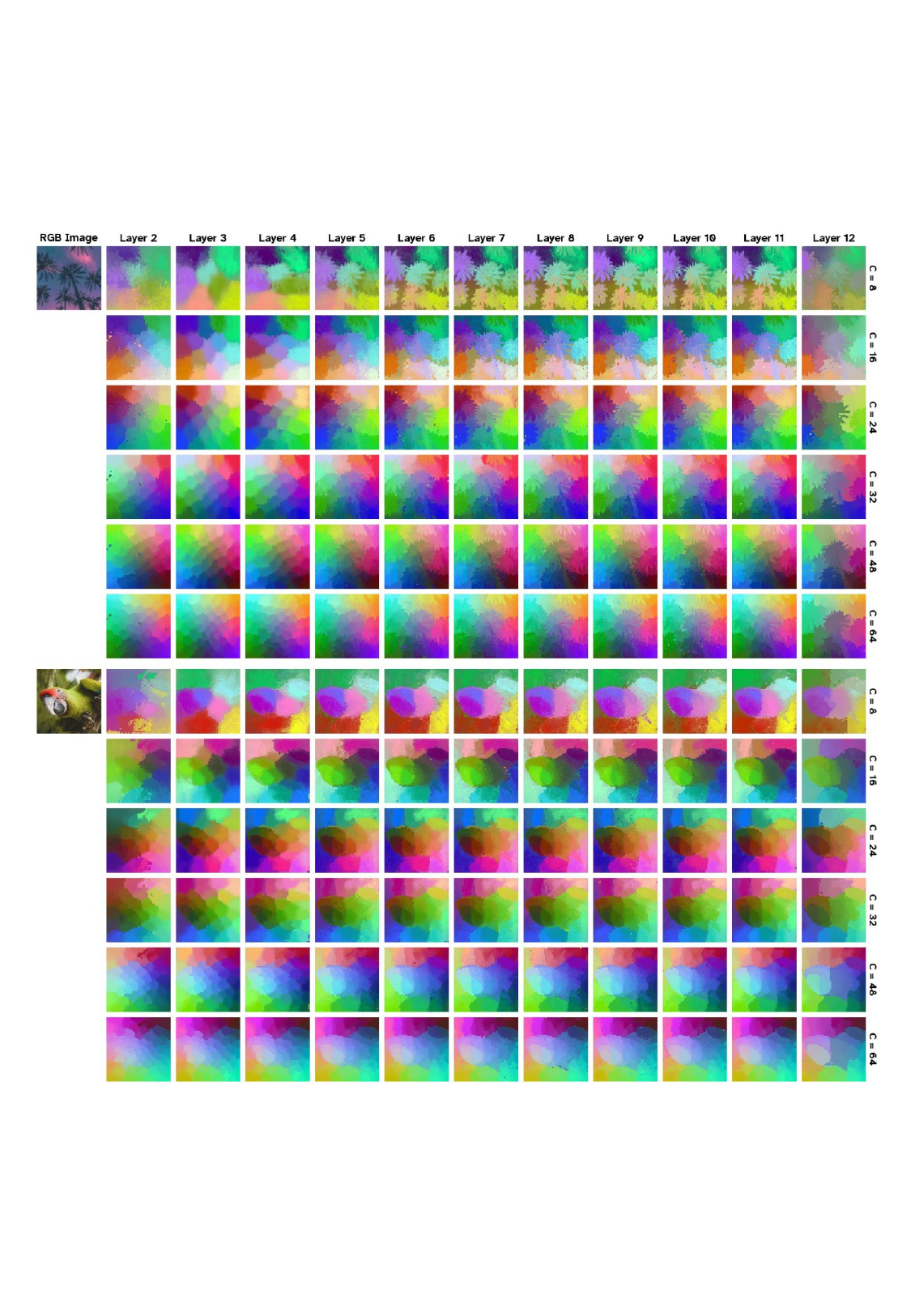}
    \caption{
    \textbf{Additional layer-wise emergent behavior visualizations.}
    We visualize UMAP embeddings of patch contribution profiles across active core queries over layers and core budgets. Please zoom in for details.
    }
    \label{fig:supp_additional_emergent_behavior_1}
\end{figure*}

\begin{figure*}[htbp]
    \centering
    \includegraphics[
        width=1.0\textwidth,
        trim={0cm 4.5cm 0cm 4cm},
        clip
    ]{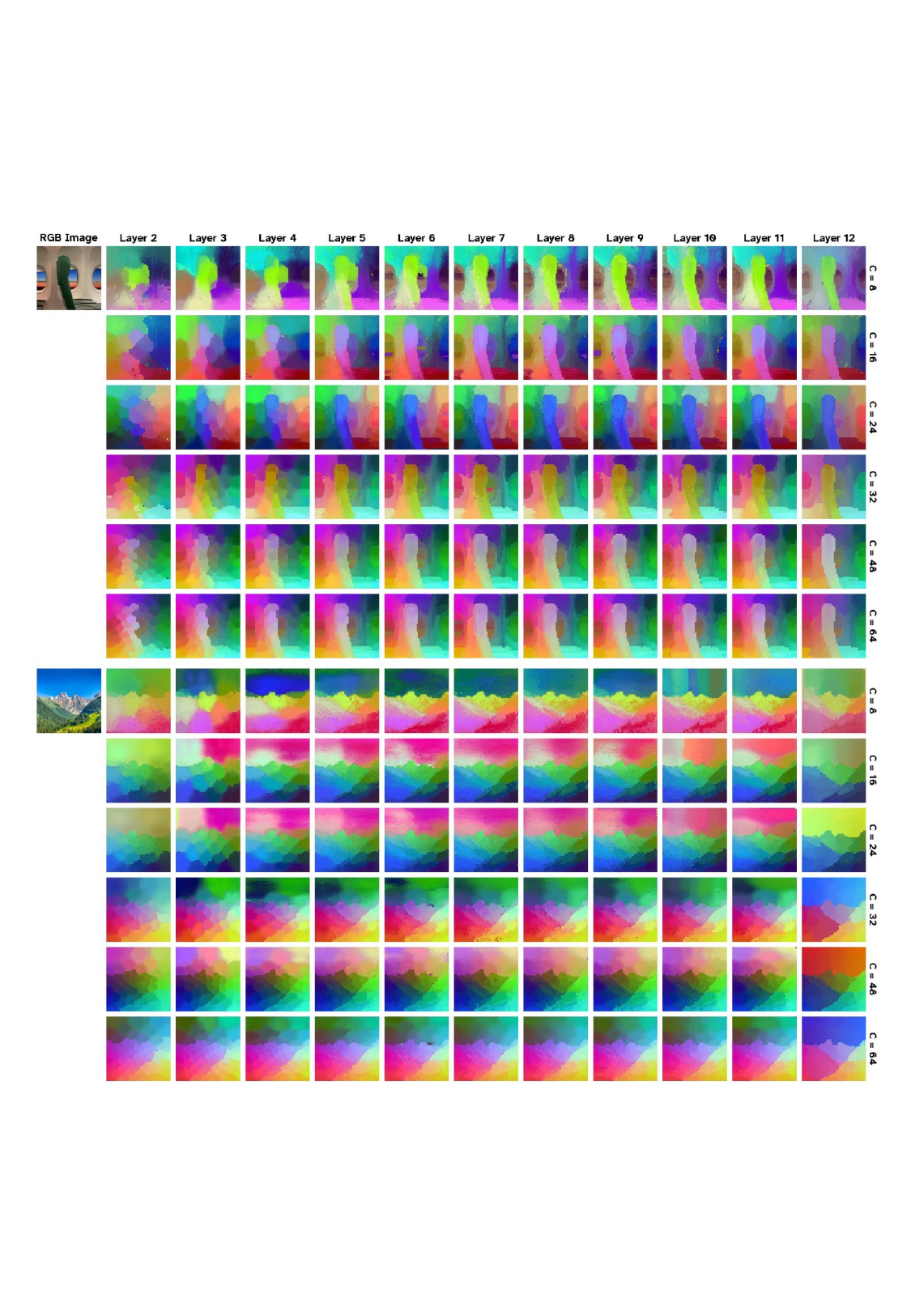}
    \caption{
    \textbf{Additional layer-wise emergent behavior visualizations.}
    We visualize UMAP embeddings of patch contribution profiles across active core queries over layers and core budgets. Please zoom in for details.
    }
    \label{fig:supp_additional_emergent_behavior_2}
\end{figure*}


%% file: supp_sections/13_efficiency.tex

This section provides efficiency analysis for VECA.
The main computational advantage of VECA comes from replacing dense patch-to-patch self-attention with core-mediated attention.
For an active core budget \(C\) and \(N\) patch tokens, the attention interaction cost scales as
\[
N^2
\quad\longrightarrow\quad
2NC + C^2.
\]
Thus, when \(C \ll N\), the benefit becomes more pronounced as the input resolution increases.

\myparagraph{Benchmark design.}
We compare VECA with the corresponding DINOv3 backbone at the Small, Base, and Large scales.
To isolate the cost of the attention mechanism, we benchmark the same projection-attention-projection path for both models: query, key, and value projections, scaled dot-product attention, and the output projection.
For VECA, patch tokens attend only to the active core tokens, while core tokens attend to the full sequence.
For DINOv3, all tokens use standard full self-attention.
Unless otherwise stated, VECA uses the full active core budget \(C=64\).
We measure FLOPs over a resolution sweep from \(256\) to \(1024\), reported per image for a single attention layer and counting one multiply-add as one FLOP, as in Table~\ref{tab:ablation}; GPU latency is measured on an NVIDIA L40S 48GB GPU with bfloat16 precision.

\begin{figure}[h]
    \centering
    \includegraphics[width=\linewidth]{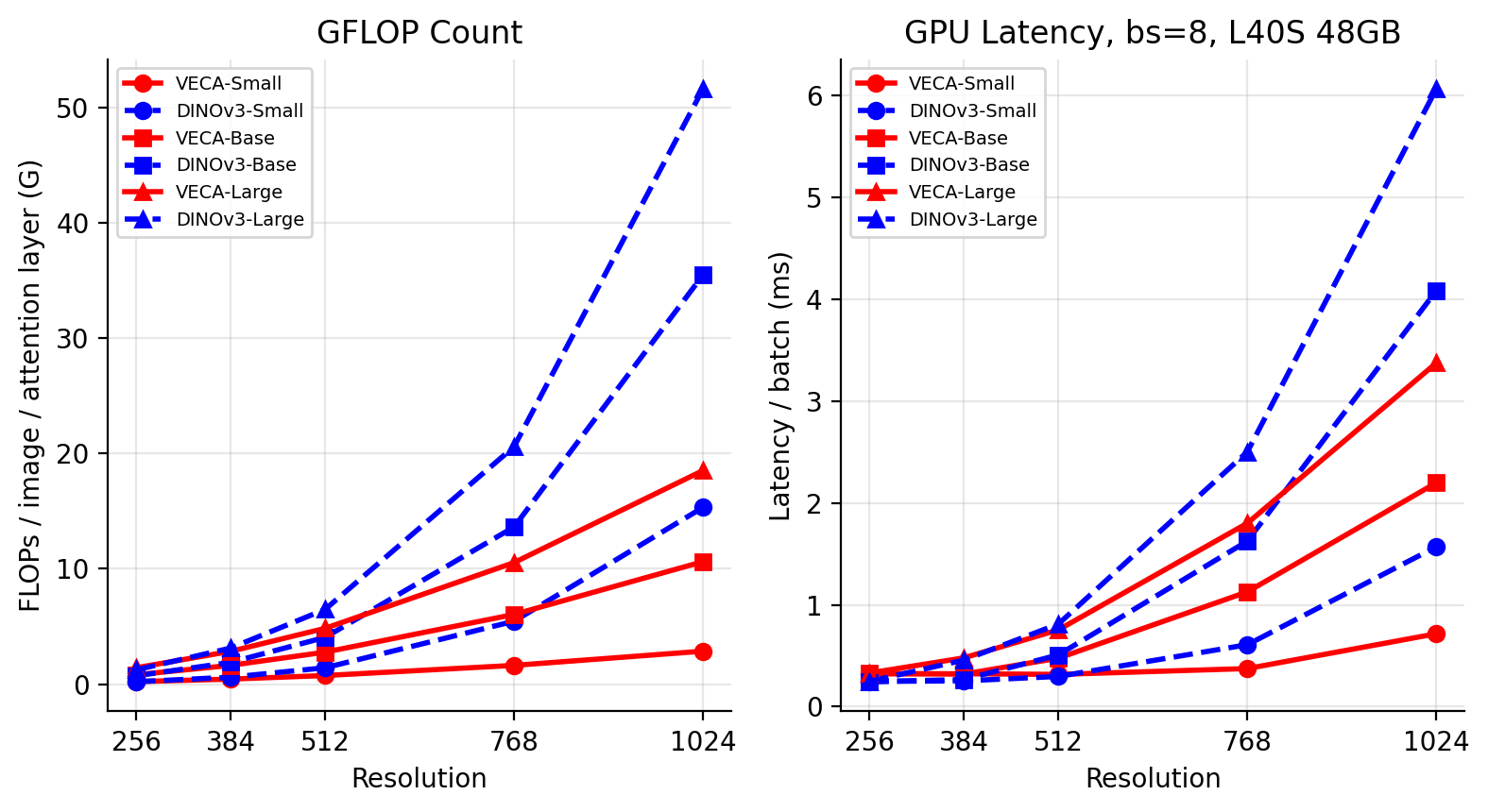}
    \caption{
    \textbf{Resolution sweep for attention cost.}
    We compare VECA and DINOv3 across input resolutions for Small, Base, and Large models.
    Left: FLOPs per image for a single attention layer.
    Right: GPU latency at batch size 8 on an NVIDIA L40S 48GB GPU.
    VECA becomes increasingly efficient as resolution grows because the removed patch-to-patch attention term dominates at high token counts.
    }
    \label{fig:supp_efficiency_res}
\end{figure}

\myparagraph{Resolution scaling.}
Figure~\ref{fig:supp_efficiency_res} shows that the efficiency advantage grows with resolution.
At \(256\), the \(C\) added core tokens make VECA slightly more expensive, since the sequence is short and projection/kernel overheads dominate; VECA requires fewer FLOPs from \(384\) onward and has lower latency from \(512\) (Base, Large) or \(768\) (Small).
At the largest evaluated resolution, VECA substantially reduces both FLOPs and GPU latency across all model sizes.
For VECA-Small, FLOPs decrease from \(15.34\)G to \(2.86\)G, a \(5.36\times\) reduction, while latency decreases from \(1.57\)ms to \(0.72\)ms.
For VECA-Base, FLOPs decrease from \(35.51\)G to \(10.63\)G, and latency decreases from \(4.08\)ms to \(2.20\)ms.
For VECA-Large, FLOPs decrease from \(51.64\)G to \(18.53\)G, and latency decreases from \(6.07\)ms to \(3.37\)ms.

\begin{table}[h]
\centering
\footnotesize
\setlength{\tabcolsep}{5pt}
\caption{\textbf{Attention cost at the largest evaluated resolution ($1024\times1024$).}
FLOPs are reported per image for a single attention layer. GPU latency is reported per batch with batch size 8 on an NVIDIA L40S 48GB GPU.}
\label{tab:supp_efficiency_highres}
\begin{tabular}{lcccccc}
\toprule
& \multicolumn{3}{c}{\textbf{FLOPs (G)}} & \multicolumn{3}{c}{\textbf{Latency (ms)}} \\
\cmidrule(lr){2-4} \cmidrule(lr){5-7}
\textbf{Model} & VECA & DINOv3 & Red. & VECA & DINOv3 & Speedup \\
\midrule
Small & 2.86  & 15.34 & \(5.36\times\) & 0.72 & 1.57 & \(2.19\times\) \\
Base  & 10.63 & 35.51 & \(3.34\times\) & 2.20 & 4.08 & \(1.86\times\) \\
Large & 18.53 & 51.64 & \(2.79\times\) & 3.37 & 6.07 & \(1.80\times\) \\
\bottomrule
\end{tabular}
\end{table}

\myparagraph{Batch-size scaling.}
We further evaluate GPU latency of the attention residual block over batch sizes from 1 to 128 at \(1024\times1024\) resolution.
This tests whether the advantage persists under larger batched workloads.

\begin{figure}[h]
    \centering
    \includegraphics[width=\linewidth]{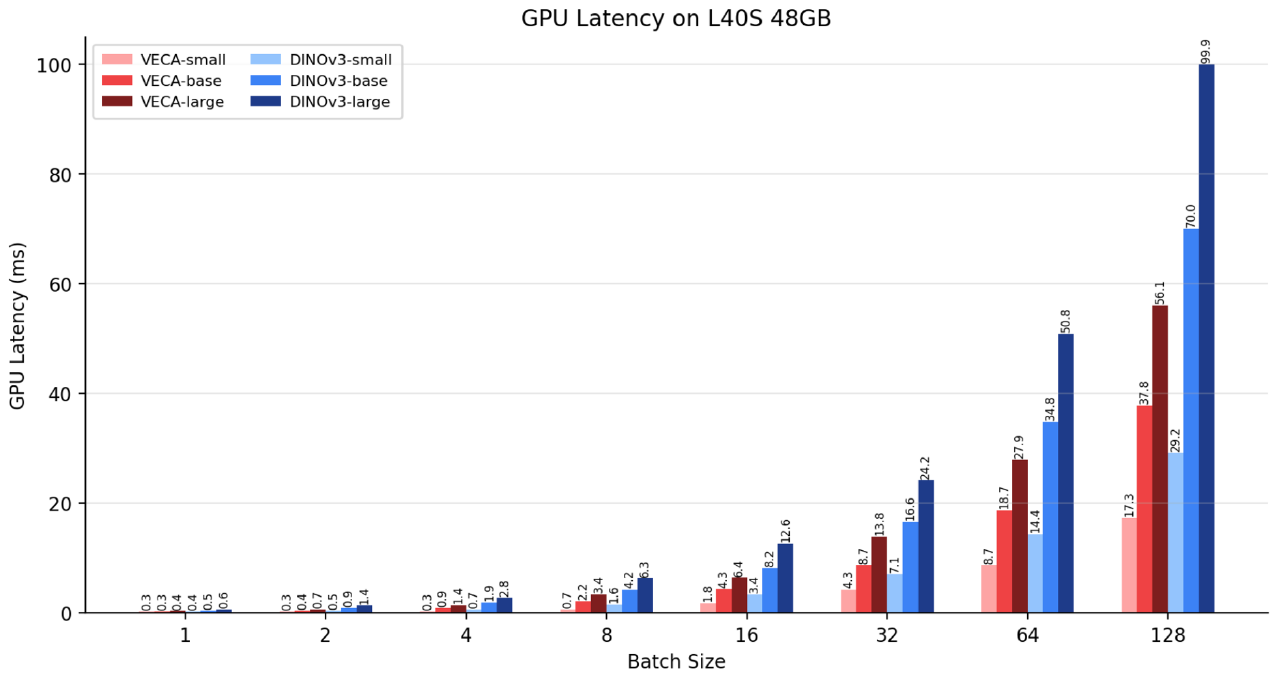}
    \caption{
    \textbf{Batch-size scaling at high resolution.}
    GPU latency is measured on an NVIDIA L40S 48GB GPU.
    The VECA attention residual block remains consistently faster than the corresponding DINOv3 residual block across model sizes and batch sizes.
    }
    \label{fig:supp_efficiency_batch}
\end{figure}

As shown in Figure~\ref{fig:supp_efficiency_batch}, the VECA attention block maintains lower latency across the batch-size sweep.
At the largest batch size, VECA-Small reduces latency from \(29.18\)ms to \(17.33\)ms, VECA-Base reduces latency from \(70.04\)ms to \(37.79\)ms, and VECA-Large reduces latency from \(99.92\)ms to \(56.07\)ms.
These correspond to \(1.68\times\), \(1.85\times\), and \(1.78\times\) speedups, respectively.

\myparagraph{Summary.}
Overall, the efficiency results support the intended scaling behavior of VECA.
The gains are most visible at high resolution, where standard ViT self-attention becomes dominated by patch-to-patch interactions.
For a fixed active core budget \(C\), VECA reduces this cost while retaining bidirectional communication between the compact core set and the dense patch sequence, yielding lower FLOPs and GPU latency once the patch count is large enough for the patch-to-patch term to dominate.

%% file: supp_sections/14_ablation.tex

We present additional analysis of VECA's elastic computation behavior when the number of active core tokens is varied for classification and dense prediction.

\subsubsection{Classification}
\label{supp_ablation_cls}

\myparagraph{Protocol.}
We vary the active core-token budget
$C\in\{8,16,24,32,40,48,56,64\}$, freeze the backbone, and train a
linear classifier on ImageNet-1K at $256{\times}256$ resolution following
Section~\ref{supp_cls_details}.

\begin{table*}[h]
\centering
\caption{\textbf{Elastic computation for ImageNet-1K classification.}
We report frozen-backbone linear-probe top-1 accuracy.
$C$ denotes the number of active core tokens at inference time.}
\label{tab:supp_ablation_cls_core_budget}
\vspace{-1.5mm}

\scriptsize
\setlength{\tabcolsep}{8pt}
\renewcommand{\arraystretch}{1.08}

\begin{tabular}{ccccc}
\toprule
\textbf{$C$}
& \textbf{VECA-S}
& \textbf{VECA-S+}
& \textbf{VECA-B}
& \textbf{VECA-L} \\
\midrule
64 & 74.63 & 77.03 & 82.18 & 84.51 \\
56 & 74.22 & 76.97 & 82.18 & 84.27 \\
48 & 74.04 & 76.97 & 82.19 & 84.35 \\
40 & 73.89 & 76.97 & 82.11 & 84.23 \\
32 & 73.73 & 77.03 & 82.02 & 84.23 \\
24 & 73.52 & 76.65 & 81.87 & 84.15 \\
16 & 72.87 & 76.17 & 81.54 & 83.90 \\
8  & 71.30 & 74.39 & 80.45 & 83.06 \\
\bottomrule
\end{tabular}
\vspace{-0.4cm}
\end{table*}

\myparagraph{Results.}
Classification remains stable across a broad range of core-token budgets.
For VECA-B, reducing the budget from $C=64$ to $C=32$ changes accuracy
by only $0.16$ points, while $C=8$ retains $80.45\%$ accuracy.
The small non-monotonic differences at larger budgets are within normal
linear-probe variation.

\subsubsection{Dense Prediction}
\label{supp_ablation_dense}

\myparagraph{Protocol.}
We use the frozen-backbone protocol in Section~\ref{supp_downstream_details} and vary the same active core-token budgets.
We evaluate semantic segmentation on PASCAL VOC and ADE20K and depth estimation on NYUv2 at \(512\times512\).

\begin{table*}[htbp]
\centering
\caption{\textbf{Elastic computation for dense prediction.}
Each cell reports PASCAL VOC mIoU, ADE20K mIoU, and NYUv2 RMSE, respectively.
\(C\) denotes the number of active core tokens at evaluation time.}
\label{tab:supp_ablation_dense_budget}
\scriptsize
\setlength{\tabcolsep}{4pt}
\renewcommand{\arraystretch}{1.08}
\begin{tabular}{ccccc}
\toprule
\textbf{\(C\)} &
\textbf{VECA-S} &
\textbf{VECA-S+} &
\textbf{VECA-B} &
\textbf{VECA-L} \\
\midrule
64 & 81.56 / 44.82 / 0.4347 & 84.12 / 46.55 / 0.4252 & 87.07 / 50.69 / 0.3705 & 88.62 / 54.17 / 0.3379 \\
56 & 81.45 / 44.80 / 0.4352 & 83.92 / 46.24 / 0.4259 & 86.94 / 50.40 / 0.3751 & 88.62 / 54.17 / 0.3398 \\
48 & 81.51 / 44.53 / 0.4361 & 83.94 / 46.08 / 0.4270 & 86.91 / 50.63 / 0.3748 & 88.59 / 54.09 / 0.3407 \\
40 & 81.40 / 43.87 / 0.4382 & 83.84 / 45.71 / 0.4299 & 86.85 / 50.53 / 0.3754 & 88.51 / 53.81 / 0.3412 \\
32 & 81.00 / 43.98 / 0.4430 & 83.84 / 45.56 / 0.4327 & 86.72 / 49.49 / 0.3789 & 88.37 / 53.90 / 0.3432 \\
24 & 81.06 / 43.23 / 0.4476 & 83.43 / 44.60 / 0.4375 & 86.34 / 49.44 / 0.3829 & 87.78 / 53.65 / 0.3474 \\
16 & 80.40 / 41.20 / 0.4562 & 82.60 / 43.03 / 0.4453 & 85.37 / 47.52 / 0.3965 & 87.67 / 52.20 / 0.3563 \\
8  & 77.56 / 37.48 / 0.4975 & 79.18 / 38.96 / 0.4898 & 83.84 / 44.51 / 0.4330 & 86.50 / 48.10 / 0.3956 \\
\bottomrule
\end{tabular}
\end{table*}

\myparagraph{Results.}
Table~\ref{tab:supp_ablation_dense_budget} shows that dense prediction is more sensitive to the active token budget.
Performance degrades gradually as \(C\) decreases, with larger drops at small budgets, particularly on ADE20K and NYUv2.
For example, reducing VECA-B from \(C=64\) to \(C=32\) changes ADE20K mIoU from \(50.69\) to \(49.49\) and NYUv2 RMSE from \(0.3705\) to \(0.3789\).

Together, the two settings highlight VECA's elastic computation behavior.
A small token budget preserves much of the global information needed for classification, while additional core tokens provide finer spatial information that is more important for dense prediction.

%% file: supp_sections/15_video.tex
This section provides implementation details to test whether the semantic structure captured by VECA cores can
be reused across video frames. Given the annotated first frame of a
DAVIS-2017 sequence, we first associate its representation with object labels.
We then use this labeled representation to predict the object labels of patches
in later frames.

\myparagraph{Assigning object labels to the first-frame representation.}
Let the final-layer patch features of the annotated first frame be
\[
P_1 \in \mathbb{R}^{N\times D},
\]
and let its VECA cores be
\[
R_1 \in \mathbb{R}^{C\times D}.
\]
The ground-truth mask is converted to patch-level soft labels
\[
Y_1 \in [0,1]^{N\times L},
\]
where \(L\) contains the background and object identities in the sequence.
For example, with two objects, each patch has a three-dimensional soft label
over background and the two objects.

To assign labels to the cores, we treat the cores as queries and the
first-frame patches as keys:
\[
Q = R_1,\qquad K = P_1,\qquad V = Y_1.
\]
After $\ell_2$-normalizing the feature vectors, we compute cosine-similarity
weights
\[
A(Q,K)
=
\operatorname{softmax}_{k}
\left(
\frac{\bar Q \bar K^\top}{\tau}
\right),
\]
where \(\operatorname{softmax}_{k}\) keeps, for each row, only the \(k\) largest
entries and sets the rest to \(-\infty\) before normalization. For core labeling,
we use \(k=64\), so each core attends to its 64 most similar first-frame
patches; this neighbor count is independent of the core budget \(C\).
The corresponding soft label of each core is
\[
Y_{\mathrm{core}}
=
A(R_1,P_1)Y_1
\in
[0,1]^{C\times L}.
\]
Thus, each core receives a soft object label by aggregating the labels of the
patches most similar to it. This is a post-hoc feature-similarity readout and
does not use VECA's internal attention weights.

\myparagraph{Propagating labels to later frames.}
For a later frame \(t\), we run VECA normally and take its final-layer patch
features
\[
P_t \in \mathbb{R}^{N_t\times D}.
\]
The current patches are used as queries, while the stored first-frame cores
and their assigned labels form the reference:
\[
Q = P_t,\qquad
K = R_1,\qquad
V = Y_{\mathrm{core}}.
\]
We again compute
\[
A(P_t,R_1)
=
\operatorname{softmax}
\left(
\frac{\bar P_t \bar R_1^\top}{\tau}
\right),
\]
over all cores, and predict the labels of the current-frame patches as
\[
\hat Y_t
=
A(P_t,R_1)Y_{\mathrm{core}}.
\]
Each current patch therefore retrieves object labels from the first-frame
cores according to feature similarity. The predicted patch labels are
upsampled to the original image resolution to obtain the segmentation mask.
The cores produced by frame \(t\) are not used in this retrieval, and no
correspondence between core indices across frames is assumed.

\myparagraph{Reference representations.}
We use the same label-transfer rule for all representations and change only
the first-frame representation used as keys and values. All compact
references are labeled with the same top-64 readout used for VECA cores.

\emph{Dense patches} retain all first-frame patches:
\[
K=P_1,\qquad V=Y_1.
\]

\emph{Uniform sampling} retains 64 spatially distributed first-frame
patches as keys.

\emph{K-means centroids} cluster the $\ell_2$-normalized first-frame patch
features \(P_1\) into 64 clusters with K-means++ initialization, and use the
centroids as keys.

\emph{VECA-CoreOnly} uses the same procedure as VECA cores, but with
\(C=256\) cores from the CoreOnly ablation, in which dense patches bypass the
intermediate blocks and only the cores are updated in those layers.

\emph{VECA cores} use \(C=64\) final cores from the standard VECA model,
with their labels assigned from the first-frame patches as described above.

\myparagraph{Implementation details.}
We use \(\tau=0.07\) throughout. For the dense reference, each current patch
retrieves from its 20 most similar first-frame patches. For reduced-token
representations, all retained tokens are used when propagating labels to later
frames. We report the standard DAVIS-2017 \(\mathcal{J}\), \(\mathcal{F}\), and
\(\mathcal{J}\&\mathcal{F}\) metrics on the validation split.

%% file: myref.bib
@article{ho2025ordered,
  title={Ordered embeddings and intrinsic dimensionalities with information-ordered bottlenecks},
  author={Ho, Matthew and Zhao, Xiaosheng and Wandelt, Benjamin D},
  journal={Machine Learning: Science and Technology},
  volume={6},
  number={3},
  pages={035010},
  year={2025},
  publisher={IOP Publishing}
}

@article{ahn2025dion,
  title={Dion: Distributed orthonormalized updates},
  author={Ahn, Kwangjun and Xu, Byron and Abreu, Natalie and Fan, Ying and Magakyan, Gagik and Sharma, Pratyusha and Zhan, Zheng and Langford, John},
  journal={arXiv preprint arXiv:2504.05295},
  year={2025}
}

@misc{jordan2024muon,
  author       = {Keller Jordan and Yuchen Jin and Vlado Boza and Jiacheng You and
                  Franz Cesista and Laker Newhouse and Jeremy Bernstein},
  title        = {Muon: An optimizer for hidden layers in neural networks},
  year         = {2024},
  url          = {https://kellerjordan.github.io/posts/muon/}
}

@article{li2025normuon,
  title={NorMuon: Making Muon more efficient and scalable},
  author={Li, Zichong and Liu, Liming and Liang, Chen and Chen, Weizhu and Zhao, Tuo},
  journal={arXiv preprint arXiv:2510.05491},
  year={2025}
}

@article{nagrani2021attention,
  title={Attention bottlenecks for multimodal fusion},
  author={Nagrani, Arsha and Yang, Shan and Arnab, Anurag and Jansen, Aren and Schmid, Cordelia and Sun, Chen},
  journal={Advances in neural information processing systems},
  volume={34},
  pages={14200--14213},
  year={2021}
}

@article{staley2022triangular,
  title={Triangular dropout: variable network width without retraining},
  author={Staley, Edward W and Markowitz, Jared},
  journal={arXiv preprint arXiv:2205.01235},
  year={2022}
}

@article{rao2021dynamicvit,
  title={Dynamicvit: Efficient vision transformers with dynamic token sparsification},
  author={Rao, Yongming and Zhao, Wenliang and Liu, Benlin and Lu, Jiwen and Zhou, Jie and Hsieh, Cho-Jui},
  journal={Advances in neural information processing systems},
  volume={34},
  pages={13937--13949},
  year={2021}
}

@article{choromanski2020rethinking,
  title={Rethinking attention with performers},
  author={Choromanski, Krzysztof and Likhosherstov, Valerii and Dohan, David and Song, Xingyou and Gane, Andreea and Sarlos, Tamas and Hawkins, Peter and Davis, Jared and Mohiuddin, Afroz and Kaiser, Lukasz and others},
  journal={arXiv preprint arXiv:2009.14794},
  year={2020}
}

@inproceedings{he2022partimagenet,
  title     = {PartImageNet: A Large, High-Quality Dataset of Parts},
  author    = {He, Ju and Yang, Shuo and Yang, Shaokang and Kortylewski, Adam and Yuan, Xiaoding and Chen, Jie-Neng and Liu, Shuai and Yang, Cheng and Yu, Qihang and Yuille, Alan},
  booktitle = {European Conference on Computer Vision (ECCV)},
  pages     = {128--145},
  year      = {2022},
  publisher = {Springer}
}

@article{ponttuset2017davis,
  author  = {Jordi Pont-Tuset and Federico Perazzi and Sergi Caelles and Pablo Arbel{\'a}ez and Alexander Sorkine-Hornung and Luc Van Gool},
  title   = {The 2017 DAVIS Challenge on Video Object Segmentation},
  journal = {arXiv preprint arXiv:1704.00675},
  year    = {2017}
}

@inproceedings{shen2021efficient,
  author    = {Shen, Zhuoran and Zhang, Mingyuan and Zhao, Haiyu and
               Yi, Shuai and Li, Hongsheng},
  title     = {Efficient Attention: Attention With Linear Complexities},
  booktitle = {Proceedings of the IEEE/CVF Winter Conference on
               Applications of Computer Vision (WACV)},
  pages     = {3531--3539},
  month     = jan,
  year      = {2021}
}

@article{hu2024matryoshka,
  title={Matryoshka query transformer for large vision-language models},
  author={Hu, Wenbo and Dou, Zi-Yi and Li, Liunian H and Kamath, Amita and Peng, Nanyun and Chang, Kai-Wei},
  journal={Advances in Neural Information Processing Systems},
  volume={37},
  pages={50168--50188},
  year={2024}
}

@inproceedings{yin2022vit,
  title={A-vit: Adaptive tokens for efficient vision transformer},
  author={Yin, Hongxu and Vahdat, Arash and Alvarez, Jose M and Mallya, Arun and Kautz, Jan and Molchanov, Pavlo},
  booktitle={Proceedings of the IEEE/CVF conference on computer vision and pattern recognition},
  pages={10809--10818},
  year={2022}
}

@article{zilberstein1996using,
  title={Using anytime algorithms in intelligent systems},
  author={Zilberstein, Shlomo},
  journal={AI magazine},
  volume={17},
  number={3},
  pages={73--83},
  year={1996}
}

@article{wang2025training,
  title={Training Matryoshka Mixture-of-Experts for Elastic Inference-Time Expert Utilization},
  author={Wang, Yaoxiang and Hu, Qingguo and Ding, Yucheng and Wang, Ruizhe and Gong, Yeyun and Jiao, Jian and Shen, Yelong and Cheng, Peng and Su, Jinsong},
  journal={arXiv preprint arXiv:2509.26520},
  year={2025}
}

@article{gu2025elastic,
  title={Elastic MoE: Unlocking the Inference-Time Scalability of Mixture-of-Experts},
  author={Gu, Naibin and Zhang, Zhenyu and Feng, Yuchen and Chen, Yilong and Fu, Peng and Lin, Zheng and Wang, Shuohuan and Sun, Yu and Wu, Hua and Wang, Weiping and others},
  journal={arXiv preprint arXiv:2509.21892},
  year={2025}
}

@article{hojjat2025thinkingvit,
  title={Thinkingvit: Matryoshka thinking vision transformer for elastic inference},
  author={Hojjat, Ali and Haberer, Janek and Pirk, Soren and Landsiedel, Olaf},
  journal={arXiv preprint arXiv:2507.10800},
  year={2025}
}

@inproceedings{xue2023adaptive,
  title={Adaptive computation with elastic input sequence},
  author={Xue, Fuzhao and Likhosherstov, Valerii and Arnab, Anurag and Houlsby, Neil and Dehghani, Mostafa and You, Yang},
  booktitle={International Conference on Machine Learning},
  pages={38971--38988},
  year={2023},
  organization={PMLR}
}

@article{bolya2022token,
  title={Token merging: Your vit but faster},
  author={Bolya, Daniel and Fu, Cheng-Yang and Dai, Xiaoliang and Zhang, Peizhao and Feichtenhofer, Christoph and Hoffman, Judy},
  journal={arXiv preprint arXiv:2210.09461},
  year={2022}
}

@article{jeddi2026loopformer,
  title={Loopformer: Elastic-depth looped transformers for latent reasoning via shortcut modulation},
  author={Jeddi, Ahmadreza and Ciccone, Marco and Taati, Babak},
  journal={arXiv preprint arXiv:2602.11451},
  year={2026}
}

@article{dehghani2018universal,
  title={Universal transformers},
  author={Dehghani, Mostafa and Gouws, Stephan and Vinyals, Oriol and Uszkoreit, Jakob and Kaiser, {\L}ukasz},
  journal={arXiv preprint arXiv:1807.03819},
  year={2018}
}

@article{xie2025accelerating,
  title={Accelerating Optimization via Differentiable Stopping Time},
  author={Xie, Zhonglin and Fong, Yiman and Yuan, Haoran and Wen, Zaiwen},
  journal={arXiv preprint arXiv:2505.22509},
  year={2025}
}

@article{graves2016adaptive,
  title={Adaptive computation time for recurrent neural networks},
  author={Graves, Alex},
  journal={arXiv preprint arXiv:1603.08983},
  year={2016}
}

@inproceedings{liu2020fastbert,
  title={Fastbert: a self-distilling bert with adaptive inference time},
  author={Liu, Weijie and Zhou, Peng and Wang, Zhiruo and Zhao, Zhe and Deng, Haotang and Ju, Qi},
  booktitle={Proceedings of the 58th annual meeting of the association for computational linguistics},
  pages={6035--6044},
  year={2020}
}

@inproceedings{wu2018blockdrop,
  title={Blockdrop: Dynamic inference paths in residual networks},
  author={Wu, Zuxuan and Nagarajan, Tushar and Kumar, Abhishek and Rennie, Steven and Davis, Larry S and Grauman, Kristen and Feris, Rogerio},
  booktitle={Proceedings of the IEEE conference on computer vision and pattern recognition},
  pages={8817--8826},
  year={2018}
}

@inproceedings{teerapittayanon2016branchynet,
  title={Branchynet: Fast inference via early exiting from deep neural networks},
  author={Teerapittayanon, Surat and McDanel, Bradley and Kung, Hsiang-Tsung},
  booktitle={2016 23rd International Conference on Pattern Recognition (ICPR)},
  pages={2464--2469},
  year={2016},
  organization={IEEE}
}

@article{kudugunta2024matformer,
  title={{MatFormer}: {N}ested transformer for elastic inference},
  author={Devvrit, Fnu and Kudugunta, Sneha and Kusupati, Aditya and Dettmers, Tim and Chen, Kaifeng and Dhillon, Inderjit and Tsvetkov, Yulia and Hajishirzi, Hannaneh and Kakade, Sham and Farhadi, Ali and Jain, Prateek},
  journal={Advances in Neural Information Processing Systems},
  volume={37},
  pages={140535--140564},
  year={2024}
}

@article{cai2024flextron,
  title={Flextron: {M}any-in-one flexible large language model},
  author={Cai, Ruisi and Muralidharan, Saurav and Heinrich, Greg and Yin, Hongxu and Wang, Zhangyang and Kautz, Jan and Molchanov, Pavlo},
  journal={arXiv preprint arXiv:2406.10260},
  year={2024}
}

@inproceedings{li2024subnetwork,
  title={Subnetwork-to-go: {E}lastic neural network with dynamic training and customizable inference},
  author={Li, Kai and Luo, Yi},
  booktitle={ICASSP 2024-2024 IEEE International Conference on Acoustics, Speech and Signal Processing (ICASSP)},
  pages={6775--6779},
  year={2024},
  organization={IEEE}
}

@article{cai2019once,
  title={Once-for-all: Train one network and specialize it for efficient deployment},
  author={Cai, Han and Gan, Chuang and Wang, Tianzhe and Zhang, Zhekai and Han, Song},
  journal={arXiv preprint arXiv:1908.09791},
  year={2019}
}

@article{valipour2023sortednet,
  title={Sortednet: A scalable and generalized framework for training modular deep neural networks},
  author={Valipour, Mojtaba and Rezagholizadeh, Mehdi and Rajabzadeh, Hossein and Kavehzadeh, Parsa and Tahaei, Marzieh and Chen, Boxing and Ghodsi, Ali},
  journal={arXiv preprint arXiv:2309.00255},
  year={2023}
}

@article{yu2018slimmable,
  title={Slimmable neural networks},
  author={Yu, Jiahui and Yang, Linjie and Xu, Ning and Yang, Jianchao and Huang, Thomas},
  journal={arXiv preprint arXiv:1812.08928},
  year={2018}
}

@inproceedings{kim2018nestednet,
  title={Nestednet: Learning nested sparse structures in deep neural networks},
  author={Kim, Eunwoo and Ahn, Chanho and Oh, Songhwai},
  booktitle={Proceedings of the IEEE Conference on Computer Vision and Pattern Recognition},
  pages={8669--8678},
  year={2018}
}

@inproceedings{tann2016runtime,
  title={Runtime configurable deep neural networks for energy-accuracy trade-off},
  author={Tann, Hokchhay and Hashemi, Soheil and Bahar, R Iris and Reda, Sherief},
  booktitle={Proceedings of the eleventh IEEE/acm/ifip International Conference on Hardware/Software Codesign and System Synthesis},
  pages={1--10},
  year={2016}
}

@article{ladjal2019pca,
  title={A PCA-like autoencoder},
  author={Ladjal, Sa{\"\i}d and Newson, Alasdair and Pham, Chi-Hieu},
  journal={arXiv preprint arXiv:1904.01277},
  year={2019}
}

@article{shen2024distributional,
  title={Distributional principal autoencoders},
  author={Shen, Xinwei and Meinshausen, Nicolai},
  journal={arXiv preprint arXiv:2404.13649},
  year={2024}
}

@inproceedings{rippel2014learning,
  title={Learning ordered representations with nested dropout},
  author={Rippel, Oren and Gelbart, Michael and Adams, Ryan},
  booktitle={International Conference on Machine Learning},
  pages={1746--1754},
  year={2014},
  organization={PMLR}
}

@inproceedings{lu2024unified,
  title={{Unified-IO}~2: {S}caling autoregressive multimodal models with vision language audio and action},
  author={Lu, Jiasen and Clark, Christopher and Lee, Sangho and Zhang, Zichen and Khosla, Savya and Marten, Ryan and Hoiem, Derek and Kembhavi, Aniruddha},
  booktitle={Proceedings of the IEEE/CVF Conference on Computer Vision and Pattern Recognition},
  pages={26439--26455},
  year={2024}
}

@article{lu2024fit,
  title={{FiT}: {F}lexible vision transformer for diffusion model},
  author={Lu, Zeyu and Wang, Zidong and Huang, Di and Wu, Chengyue and Liu, Xihui and Ouyang, Wanli and Bai, Lei},
  journal={arXiv preprint arXiv:2402.12376},
  year={2024}
}

@article{simeoni2025dinov3,
  title={{DINOv3}},
  author={Sim{\'e}oni, Oriane and Vo, Huy V and Seitzer, Maximilian and Baldassarre, Federico and Oquab, Maxime and Jose, Cijo and Khalidov, Vasil and Szafraniec, Marc and Yi, Seungeun and Ramamonjisoa, Micha{\"e}l and others},
  journal={arXiv preprint arXiv:2508.10104},
  year={2025}
}

@article{fang2024eva,
  title={{EVA-02}: {A} visual representation for neon genesis},
  author={Fang, Yuxin and Sun, Quan and Wang, Xinggang and Huang, Tiejun and Wang, Xinlong and Cao, Yue},
  journal={Image and Vision Computing},
  volume={149},
  pages={105171},
  year={2024},
  publisher={Elsevier}
}

@article{chen2025vision,
  title={Vision transformers with self-distilled registers},
  author={Chen, Yinjie and Yan, Zipeng and Zhou, Chong and Dai, Bo and Luo, Andrew F},
  journal={arXiv preprint arXiv:2505.21501},
  year={2025}
}

@article{darcet2023vision,
  title={Vision transformers need registers},
  author={Darcet, Timoth{\'e}e and Oquab, Maxime and Mairal, Julien and Bojanowski, Piotr},
  journal={arXiv preprint arXiv:2309.16588},
  year={2023}
}

@article{burtsev2020memory,
  title={Memory transformer},
  author={Burtsev, Mikhail S and Kuratov, Yuri and Peganov, Anton and Sapunov, Grigory V},
  journal={arXiv preprint arXiv:2006.11527},
  year={2020}
}

@inproceedings{sariyildiz2024unic,
  title={{UNIC}: {U}niversal classification models via multi-teacher distillation},
  author={Sar{\i}y{\i}ld{\i}z, Mert B{\"u}lent and Weinzaepfel, Philippe and Lucas, Thomas and Larlus, Diane and Kalantidis, Yannis},
  booktitle={European Conference on Computer Vision},
  pages={353--371},
  year={2024},
  organization={Springer}
}

@article{shang2024theia,
  title={Theia: {D}istilling diverse vision foundation models for robot learning},
  author={Shang, Jinghuan and Schmeckpeper, Karl and May, Brandon B and Minniti, Maria Vittoria and Kelestemur, Tarik and Watkins, David and Herlant, Laura},
  journal={arXiv preprint arXiv:2407.20179},
  year={2024}
}

@article{wei2022contrastive,
  title={Contrastive learning rivals masked image modeling in fine-tuning via feature distillation},
  author={Wei, Yixuan and Hu, Han and Xie, Zhenda and Zhang, Zheng and Cao, Yue and Bao, Jianmin and Chen, Dong and Guo, Baining},
  journal={arXiv preprint arXiv:2205.14141},
  year={2022}
}

@article{kusupati2022matryoshka,
  title={Matryoshka representation learning},
  author={Kusupati, Aditya and Bhatt, Gantavya and Rege, Aniket and Wallingford, Matthew and Sinha, Aditya and Ramanujan, Vivek and Howard-Snyder, William and Chen, Kaifeng and Kakade, Sham and Jain, Prateek and others},
  journal={Advances in Neural Information Processing Systems},
  volume={35},
  pages={30233--30249},
  year={2022}
}

@article{chen2025cautious,
  title={Cautious Weight Decay},
  author={Chen, Lizhang and Li, Jonathan and Liang, Kaizhao and Su, Baiyu and Xie, Cong and Pierse, Nuo Wang and Liang, Chen and Lao, Ni and Liu, Qiang},
  journal={arXiv preprint arXiv:2510.12402},
  year={2025}
}

@article{liu2025muon,
  title={Muon is scalable for llm training},
  author={Liu, Jingyuan and Su, Jianlin and Yao, Xingcheng and Jiang, Zhejun and Lai, Guokun and Du, Yulun and Qin, Yidao and Xu, Weixin and Lu, Enzhe and Yan, Junjie and others},
  journal={arXiv preprint arXiv:2502.16982},
  year={2025}
}

@article{amsel2025polar,
  title={The polar express: Optimal matrix sign methods and their application to the muon algorithm},
  author={Amsel, Noah and Persson, David and Musco, Christopher and Gower, Robert M},
  journal={arXiv preprint arXiv:2505.16932},
  year={2025}
}

@article{loshchilov2017decoupled,
  title={Decoupled weight decay regularization},
  author={Loshchilov, Ilya and Hutter, Frank},
  journal={arXiv preprint arXiv:1711.05101},
  year={2017}
}

@article{kingma2014adam,
  title={Adam: A method for stochastic optimization},
  author={Kingma, Diederik P and Ba, Jimmy},
  journal={arXiv preprint arXiv:1412.6980},
  year={2014}
}

@article{dosovitskiy2020image,
  title={An image is worth 16x16 words: Transformers for image recognition at scale},
  author={Dosovitskiy, Alexey and Beyer, Lucas and Kolesnikov, Alexander and Weissenborn, Dirk and Zhai, Xiaohua and Unterthiner, Thomas and Dehghani, Mostafa and Minderer, Matthias and Heigold, Georg and Gelly, Sylvain and others},
  journal={arXiv preprint arXiv:2010.11929},
  year={2020}
}

@article{raghu2021vision,
  title={Do vision transformers see like convolutional neural networks?},
  author={Raghu, Maithra and Unterthiner, Thomas and Kornblith, Simon and Zhang, Chiyuan and Dosovitskiy, Alexey},
  journal={Advances in neural information processing systems},
  volume={34},
  pages={12116--12128},
  year={2021}
}

@inproceedings{shao2019objects365,
  title={Objects365: A large-scale, high-quality dataset for object detection},
  author={Shao, Shuai and Li, Zeming and Zhang, Tianyuan and Peng, Chao and Yu, Gang and Zhang, Xiangyu and Li, Jing and Sun, Jian},
  booktitle={Proceedings of the IEEE/CVF International Conference on Computer Vision},
  pages={8430--8439},
  year={2019}
}

@article{schenck2025learning,
  title={Learning the ropes: Better 2d and 3d position encodings with string},
  author={Schenck, Connor and Reid, Isaac and Jacob, Mithun George and Bewley, Alex and Ainslie, Joshua and Rendleman, David and Jain, Deepali and Sharma, Mohit and Dubey, Avinava and Wahid, Ayzaan and others},
  journal={arXiv preprint arXiv:2502.02562},
  year={2025}
}

@inproceedings{yu2025comrope,
  title={Comrope: {S}calable and robust rotary position embedding parameterized by trainable commuting angle matrices},
  author={Yu, Hao and Jiang, Tangyu and Jia, Shuning and Yan, Shannan and Liu, Shunning and Qian, Haolong and Li, Guanghao and Dong, Shuting and Yuan, Chun},
  booktitle={Proceedings of the Computer Vision and Pattern Recognition Conference},
  pages={4508--4517},
  year={2025}
}

@article{ostmeier2024liere,
  title={Liere: Lie rotational positional encodings},
  author={Ostmeier, Sophie and Axelrod, Brian and Varma, Maya and Moseley, Michael E and Chaudhari, Akshay and Langlotz, Curtis},
  journal={arXiv preprint arXiv:2406.10322},
  year={2024}
}

@article{van2025circular,
  title={A Circular Argument: Does RoPE need to be Equivariant for Vision?},
  author={van de Geijn, Chase and L{\"u}ddecke, Timo and Turishcheva, Polina and Ecker, Alexander S},
  journal={arXiv preprint arXiv:2511.08368},
  year={2025}
}

@inproceedings{ranzinger2024radio,
  title={AM-RADIO: Agglomerative Vision Foundation Model Reduce All Domains Into One},
  author={Ranzinger, Mike and Heinrich, Greg and Kautz, Jan and Molchanov, Pavlo},
  booktitle={Proceedings of the IEEE/CVF Conference on Computer Vision and Pattern Recognition},
  pages={12490--12500},
  year={2024}
}

@inproceedings{jaegle2021perceiver,
  title={Perceiver: General perception with iterative attention},
  author={Jaegle, Andrew and Gimeno, Felix and Brock, Andy and Vinyals, Oriol and Zisserman, Andrew and Carreira, Joao},
  booktitle={International Conference on Machine Learning},
  pages={4651--4664},
  year={2021},
  organization={PMLR}
}

@article{shah2018deep,
  title={Deep continuous clustering},
  author={Shah, Sohil Atul and Koltun, Vladlen},
  journal={arXiv preprint arXiv:1803.01449},
  year={2018}
}

@article{ke2022learning,
  title={Learning hierarchical image segmentation for recognition and by recognition},
  author={Ke, Tsung-Wei and Mo, Sangwoo and Yu, Stella X},
  journal={arXiv preprint arXiv:2210.00314},
  year={2022}
}

@inproceedings{aasan2024spitting,
  title={A spitting image: {M}odular superpixel tokenization in vision transformers},
  author={Aasan, Marius and Kolbj{\o}rnsen, Odd and Solberg, Anne Schistad and Rivera, Ad{\'\i}n Ramirez},
  booktitle={European Conference on Computer Vision, MELEX Workshop},
  pages={124--142},
  year={2024},
  organization={Springer}
}

@article{zeng2024tcformer,
  title={TCFormer: Visual recognition via token clustering transformer},
  author={Zeng, Wang and Jin, Sheng and Xu, Lumin and Liu, Wentao and Qian, Chen and Ouyang, Wanli and Luo, Ping and Wang, Xiaogang},
  journal={IEEE Transactions on Pattern Analysis and Machine Intelligence},
  volume={46},
  number={12},
  pages={9521--9535},
  year={2024},
  publisher={IEEE}
}

@article{liang2023clusterfomer,
  title={Clusterfomer: clustering as a universal visual learner},
  author={Liang, James and Cui, Yiming and Wang, Qifan and Geng, Tong and Wang, Wenguan and Liu, Dongfang},
  journal={Advances in neural information processing systems},
  volume={36},
  pages={64029--64042},
  year={2023}
}

@article{braso2025native,
  title={Native segmentation vision transformers},
  author={Bras{\'o}, Guillem and O{\v{s}}ep, Aljo{\v{s}}a and Leal-Taix{\'e}, Laura},
  journal={arXiv preprint arXiv:2505.16993},
  year={2025}
}

@article{islam2020much,
  title={How much position information do convolutional neural networks encode?},
  author={Islam, Md Amirul and Jia, Sen and Bruce, Neil DB},
  journal={arXiv preprint arXiv:2001.08248},
  year={2020}
}

@article{luo2016understanding,
  title={Understanding the effective receptive field in deep convolutional neural networks},
  author={Luo, Wenjie and Li, Yujia and Urtasun, Raquel and Zemel, Richard},
  journal={Advances in neural information processing systems},
  volume={29},
  year={2016}
}

@article{rao2023gfnet,
  title={GFNet: Global filter networks for visual recognition},
  author={Rao, Yongming and Zhao, Wenliang and Zhu, Zheng and Zhou, Jie and Lu, Jiwen},
  journal={IEEE Transactions on Pattern Analysis and Machine Intelligence},
  volume={45},
  number={9},
  pages={10960--10973},
  year={2023},
  publisher={IEEE}
}

@article{park2022vision,
  title={How do vision transformers work?},
  author={Park, Namuk and Kim, Songkuk},
  journal={arXiv preprint arXiv:2202.06709},
  year={2022}
}

@article{luce1949method,
  title={A method of matrix analysis of group structure},
  author={Luce, R Duncan and Perry, Albert D},
  journal={Psychometrika},
  volume={14},
  number={2},
  pages={95--116},
  year={1949},
  publisher={Springer}
}

@article{zhang2015identification,
  title={Identification of core-periphery structure in networks},
  author={Zhang, Xiao and Martin, Travis and Newman, Mark EJ},
  journal={Physical Review E},
  volume={91},
  number={3},
  pages={032803},
  year={2015},
  publisher={APS}
}

@article{rombach2014core,
  title={Core-periphery structure in networks},
  author={Rombach, M Puck and Porter, Mason A and Fowler, James H and Mucha, Peter J},
  journal={SIAM Journal on Applied mathematics},
  volume={74},
  number={1},
  pages={167--190},
  year={2014},
  publisher={SIAM}
}

@inproceedings{nauen2025transformer,
  title={Which transformer to favor: a comparative analysis of efficiency in vision transformers},
  author={Nauen, Tobias Christian and Palacio, Sebastian and Raue, Federico and Dengel, Andreas},
  booktitle={2025 IEEE/CVF Winter Conference on Applications of Computer Vision (WACV)},
  pages={6955--6966},
  year={2025},
  organization={IEEE}
}

@inproceedings{li2023rethinking,
  title={Rethinking vision transformers for mobilenet size and speed},
  author={Li, Yanyu and Hu, Ju and Wen, Yang and Evangelidis, Georgios and Salahi, Kamyar and Wang, Yanzhi and Tulyakov, Sergey and Ren, Jian},
  booktitle={Proceedings of the IEEE/CVF International Conference on Computer Vision},
  pages={16889--16900},
  year={2023}
}

@article{liu2021pay,
  title={Pay attention to mlps},
  author={Liu, Hanxiao and Dai, Zihang and So, David and Le, Quoc V},
  journal={Advances in neural information processing systems},
  volume={34},
  pages={9204--9215},
  year={2021}
}

@article{touvron2022resmlp,
  title={Resmlp: Feedforward networks for image classification with data-efficient training},
  author={Touvron, Hugo and Bojanowski, Piotr and Caron, Mathilde and Cord, Matthieu and El-Nouby, Alaaeldin and Grave, Edouard and Izacard, Gautier and Joulin, Armand and Synnaeve, Gabriel and Verbeek, Jakob and others},
  journal={IEEE transactions on pattern analysis and machine intelligence},
  volume={45},
  number={4},
  pages={5314--5321},
  year={2022},
  publisher={IEEE}
}

@article{tolstikhin2021mlp,
  title={Mlp-mixer: An all-mlp architecture for vision},
  author={Tolstikhin, Ilya O and Houlsby, Neil and Kolesnikov, Alexander and Beyer, Lucas and Zhai, Xiaohua and Unterthiner, Thomas and Yung, Jessica and Steiner, Andreas and Keysers, Daniel and Uszkoreit, Jakob and others},
  journal={Advances in neural information processing systems},
  volume={34},
  pages={24261--24272},
  year={2021}
}

@article{pu2025linear,
  title={Linear Differential Vision Transformer: Learning Visual Contrasts via Pairwise Differentials},
  author={Pu, Yifan and Ying, Jixuan and Li, Qixiu and Ye, Tianzhu and Han, Dongchen and Wang, Xiaochen and Wang, Ziyi and Shao, Xinyu and Huang, Gao and Li, Xiu},
  journal={arXiv preprint arXiv:2511.00833},
  year={2025}
}

@article{zhang2024slicing,
  title={Slicing vision transformer for flexible inference},
  author={Zhang, Yitian and Coskun, Huseyin and Ma, Xu and Wang, Huan and Ma, Ke and Chen, Xi and Hu, Derek H and Fu, Yun},
  journal={Advances in Neural Information Processing Systems},
  volume={37},
  pages={42649--42671},
  year={2024}
}

@article{khosla2025ren,
  title={REN: Fast and Efficient Region Encodings from Patch-Based Image Encoders},
  author={Khosla, Savya and TV, Sethuraman and Lee, Barnett and Schwing, Alexander and Hoiem, Derek},
  journal={arXiv preprint arXiv:2505.18153},
  year={2025}
}

@article{mei2024spformer,
  title={Spformer: Enhancing vision transformer with superpixel representation},
  author={Mei, Jieru and Chen, Liang-Chieh and Yuille, Alan and Xie, Cihang},
  journal={arXiv preprint arXiv:2401.02931},
  year={2024}
}

@inproceedings{jampani2018superpixel,
  title={Superpixel sampling networks},
  author={Jampani, Varun and Sun, Deqing and Liu, Ming-Yu and Yang, Ming-Hsuan and Kautz, Jan},
  booktitle={Proceedings of the European conference on computer vision (ECCV)},
  pages={352--368},
  year={2018}
}

@inproceedings{xu2022groupvit,
  title={Groupvit: Semantic segmentation emerges from text supervision},
  author={Xu, Jiarui and De Mello, Shalini and Liu, Sifei and Byeon, Wonmin and Breuel, Thomas and Kautz, Jan and Wang, Xiaolong},
  booktitle={Proceedings of the IEEE/CVF conference on computer vision and pattern recognition},
  pages={18134--18144},
  year={2022}
}

@article{aasan2025differentiable,
  title={Differentiable Hierarchical Visual Tokenization},
  author={Aasan, Marius and Hjelkrem-Tan, Martine and Catalano, Nico and Choi, Changkyu and Rivera, Ad{\'\i}n Ram{\'\i}rez},
  journal={arXiv preprint arXiv:2511.02652},
  year={2025}
}

@article{achanta2012slic,
  title={SLIC superpixels compared to state-of-the-art superpixel methods},
  author={Achanta, Radhakrishna and Shaji, Appu and Smith, Kevin and Lucchi, Aurelien and Fua, Pascal and S{\"u}sstrunk, Sabine},
  journal={IEEE transactions on pattern analysis and machine intelligence},
  volume={34},
  number={11},
  pages={2274--2282},
  year={2012},
  publisher={IEEE}
}

@article{locatello2020object,
  title={Object-centric learning with slot attention},
  author={Locatello, Francesco and Weissenborn, Dirk and Unterthiner, Thomas and Mahendran, Aravindh and Heigold, Georg and Uszkoreit, Jakob and Dosovitskiy, Alexey and Kipf, Thomas},
  journal={Advances in neural information processing systems},
  volume={33},
  pages={11525--11538},
  year={2020}
}

@inproceedings{leiber2021dip,
  title={Dip-based deep embedded clustering with k-estimation},
  author={Leiber, Collin and Bauer, Lena GM and Schelling, Benjamin and B{\"o}hm, Christian and Plant, Claudia},
  booktitle={Proceedings of the 27th ACM SIGKDD Conference on Knowledge Discovery \& Data Mining},
  pages={903--913},
  year={2021}
}

@inproceedings{leiber2024dying,
  title={Dying Clusters Is All You Need-Deep Clustering With an Unknown Number of Clusters},
  author={Leiber, Collin and Strau{\ss}, Niklas and Schubert, Matthias and Seidl, Thomas},
  booktitle={2024 IEEE International Conference on Data Mining Workshops (ICDMW)},
  pages={726--733},
  year={2024},
  organization={IEEE}
}

@article{ren2020deep,
  title={Deep density-based image clustering},
  author={Ren, Yazhou and Wang, Ni and Li, Mingxia and Xu, Zenglin},
  journal={Knowledge-Based Systems},
  volume={197},
  pages={105841},
  year={2020},
  publisher={Elsevier}
}

@inproceedings{duan2019improving,
  title={Improving spectral clustering with deep embedding and cluster estimation},
  author={Duan, Liang and Aggarwal, Charu and Ma, Shuai and Sathe, Saket},
  booktitle={2019 IEEE International Conference on Data Mining (ICDM)},
  pages={170--179},
  year={2019},
  organization={IEEE}
}

@inproceedings{kong2018recurrent,
  title={Recurrent pixel embedding for instance grouping},
  author={Kong, Shu and Fowlkes, Charless C},
  booktitle={Proceedings of the IEEE conference on computer vision and pattern recognition},
  pages={9018--9028},
  year={2018}
}

@article{comaniciu2002mean,
  title={Mean shift: A robust approach toward feature space analysis},
  author={Comaniciu, Dorin and Meer, Peter},
  journal={IEEE Transactions on pattern analysis and machine intelligence},
  volume={24},
  number={5},
  pages={603--619},
  year={2002},
  publisher={IEEE}
}

@inproceedings{ronen2022deepdpm,
  title={Deepdpm: Deep clustering with an unknown number of clusters},
  author={Ronen, Meitar and Finder, Shahaf E and Freifeld, Oren},
  booktitle={Proceedings of the IEEE/CVF Conference on Computer Vision and Pattern Recognition},
  pages={9861--9870},
  year={2022}
}

@inproceedings{bianchi2020spectral,
  title={Spectral clustering with graph neural networks for graph pooling},
  author={Bianchi, Filippo Maria and Grattarola, Daniele and Alippi, Cesare},
  booktitle={International Conference on Machine Learning},
  pages={874--883},
  year={2020},
  organization={PMLR}
}

@article{stewart2023differentiable,
  title={Differentiable clustering with perturbed spanning forests},
  author={Stewart, Lawrence and Bach, Francis and Llinares-L{\'o}pez, Felipe and Berthet, Quentin},
  journal={Advances in Neural Information Processing Systems},
  volume={36},
  pages={31158--31176},
  year={2023}
}

@inproceedings{saha2023end,
  title={End-to-end differentiable clustering with associative memories},
  author={Saha, Bishwajit and Krotov, Dmitry and Zaki, Mohammed J and Ram, Parikshit},
  booktitle={International Conference on Machine Learning},
  pages={29649--29670},
  year={2023},
  organization={PMLR}
}

@inproceedings{yang2017towards,
  title={Towards k-means-friendly spaces: Simultaneous deep learning and clustering},
  author={Yang, Bo and Fu, Xiao and Sidiropoulos, Nicholas D and Hong, Mingyi},
  booktitle={International Conference on Machine Learning},
  pages={3861--3870},
  year={2017},
  organization={PMLR}
}

@inproceedings{xie2016unsupervised,
  title={Unsupervised deep embedding for clustering analysis},
  author={Xie, Junyuan and Girshick, Ross and Farhadi, Ali},
  booktitle={International Conference on Machine Learning},
  pages={478--487},
  year={2016},
  organization={PMLR}
}

@inproceedings{hu2025improving,
  title={Improving Bilinear {RNN} with Closed-loop Control},
  author={Hu, Jiaxi and Pan, Yongqi and Du, Jusen and Lan, Disen and Tang, Xiaqiang and Wen, Qingsong and Liang, Yuxuan and Sun, Weigao},
  booktitle={The Thirty-ninth Annual Conference on Neural Information Processing Systems},
  year={2025}
}

@article{meng2025polaformer,
  title={Polaformer: Polarity-aware linear attention for vision transformers},
  author={Meng, Weikang and Luo, Yadan and Li, Xin and Jiang, Dongmei and Zhang, Zheng},
  journal={arXiv preprint arXiv:2501.15061},
  year={2025}
}

@inproceedings{han2023flatten,
  title={Flatten transformer: {V}ision transformer using focused linear attention},
  author={Han, Dongchen and Pan, Xuran and Han, Yizeng and Song, Shiji and Huang, Gao},
  booktitle={Proceedings of the IEEE/CVF International Conference on Computer Vision},
  pages={5961--5971},
  year={2023}
}

@inproceedings{you2023castling,
  title={Castling-{ViT}: {C}ompressing self-attention via switching towards linear-angular attention at vision transformer inference},
  author={You, Haoran and Xiong, Yunyang and Dai, Xiaoliang and Wu, Bichen and Zhang, Peizhao and Fan, Haoqi and Vajda, Peter and Lin, Yingyan Celine},
  booktitle={Proceedings of the IEEE/CVF conference on computer vision and pattern recognition},
  pages={14431--14442},
  year={2023}
}

@misc{cai2022efficientvit,
      title={Efficient{ViT}: {M}ulti-Scale Linear Attention for High-Resolution Dense Prediction}, 
      author={Han Cai and Junyan Li and Muyan Hu and Chuang Gan and Song Han},
      year={2024},
      archivePrefix={arXiv},
      primaryClass={cs.CV},
      url={https://arxiv.org/abs/2205.14756}, 
}

@inproceedings{bolya2022hydra,
  title={Hydra attention: {E}fficient attention with many heads},
  author={Bolya, Daniel and Fu, Cheng-Yang and Dai, Xiaoliang and Zhang, Peizhao and Hoffman, Judy},
  booktitle={European conference on computer vision, CADL Workshop},
  pages={35--49},
  year={2022},
  organization={Springer}
}

@inproceedings{han2024agent,
  title={Agent attention: On the integration of softmax and linear attention},
  author={Han, Dongchen and Ye, Tianzhu and Han, Yizeng and Xia, Zhuofan and Pan, Siyuan and Wan, Pengfei and Song, Shiji and Huang, Gao},
  booktitle={European conference on computer vision},
  pages={124--140},
  year={2024},
  organization={Springer}
}

@article{beck2024xlstm,
  title={x{LSTM}: Extended long short-term memory},
  author={Beck, Maximilian and P{\"o}ppel, Korbinian and Spanring, Markus and Auer, Andreas and Prudnikova, Oleksandra and Kopp, Michael and Klambauer, G{\"u}nter and Brandstetter, Johannes and Hochreiter, Sepp},
  journal={Advances in Neural Information Processing Systems},
  volume={37},
  pages={107547--107603},
  year={2024}
}

@article{irie2021going,
  title={Going beyond linear transformers with recurrent fast weight programmers},
  author={Irie, Kazuki and Schlag, Imanol and Csord{\'a}s, R{\'o}bert and Schmidhuber, J{\"u}rgen},
  journal={Advances in neural information processing systems},
  volume={34},
  pages={7703--7717},
  year={2021}
}

@article{zhang2024gated,
  title={Gated slot attention for efficient linear-time sequence modeling},
  author={Zhang, Yu and Yang, Songlin and Zhu, Ruijie and Zhang, Yue and Cui, Leyang and Wang, Yiqiao and Wang, Bolun and Shi, Freda and Wang, Bailin and Bi, Wei and others},
  journal={Advances in Neural Information Processing Systems},
  volume={37},
  pages={116870--116898},
  year={2024}
}

@article{gu2023mamba,
  title={Mamba: Linear-time sequence modeling with selective state spaces},
  author={Gu, Albert and Dao, Tri},
  journal={arXiv preprint arXiv:2312.00752},
  year={2023}
}

@article{qin2024hgrn2,
  title={Hgrn2: Gated linear rnns with state expansion},
  author={Qin, Zhen and Yang, Songlin and Sun, Weixuan and Shen, Xuyang and Li, Dong and Sun, Weigao and Zhong, Yiran},
  journal={arXiv preprint arXiv:2404.07904},
  year={2024}
}

@article{peng2025rwkv,
  title={{RWKV}-7 ``{G}oose'' with expressive dynamic state evolution},
  author={Peng, Bo and Zhang, Ruichong and Goldstein, Daniel and Alcaide, Eric and Du, Xingjian and Hou, Haowen and Lin, Jiaju and Liu, Jiaxing and Lu, Janna and Merrill, William and others},
  journal={arXiv preprint arXiv:2503.14456},
  year={2025}
}

@article{sun2023retentive,
  title={Retentive network: A successor to transformer for large language models},
  author={Sun, Yutao and Dong, Li and Huang, Shaohan and Ma, Shuming and Xia, Yuqing and Xue, Jilong and Wang, Jianyong and Wei, Furu},
  journal={arXiv preprint arXiv:2307.08621},
  year={2023}
}

@inproceedings{mao2022fine,
  title={Fine-tuning pre-trained transformers into decaying fast weights},
  author={Mao, Huanru Henry},
  booktitle={Proceedings of the 2022 conference on empirical methods in natural language processing},
  pages={10236--10242},
  year={2022}
}

@inproceedings{peng2022abc,
  title={ABC: Attention with bounded-memory control},
  author={Peng, Hao and Kasai, Jungo and Pappas, Nikolaos and Yogatama, Dani and Wu, Zhaofeng and Kong, Lingpeng and Schwartz, Roy and Smith, Noah A},
  booktitle={Proceedings of the 60th Annual Meeting of the Association for Computational Linguistics (Volume 1: Long Papers)},
  pages={7469--7483},
  year={2022}
}

@article{peng2021random,
  title={Random feature attention},
  author={Peng, Hao and Pappas, Nikolaos and Yogatama, Dani and Schwartz, Roy and Smith, Noah A and Kong, Lingpeng},
  journal={arXiv preprint arXiv:2103.02143},
  year={2021}
}

@article{yang2023gated,
  title={Gated linear attention transformers with hardware-efficient training},
  author={Yang, Songlin and Wang, Bailin and Shen, Yikang and Panda, Rameswar and Kim, Yoon},
  journal={arXiv preprint arXiv:2312.06635},
  year={2023}
}

@article{gu2022parameterization,
  title={On the parameterization and initialization of diagonal state space models},
  author={Gu, Albert and Gupta, Ankit and Goel, Karan and R{\'e}, Christopher},
  journal={Advances in neural information processing systems},
  volume={35},
  pages={35971--35983},
  year={2022}
}

@misc{dolga2024latte,
      title={Unifying Linear-Time Attention via Latent Probabilistic Modelling}, 
      author={Rares Dolga and Lucas Maystre and Marius Cobzarenco and David Barber},
      year={2025},
      archivePrefix={arXiv},
      primaryClass={cs.CL},
      url={https://arxiv.org/abs/2402.17512}, 
}

@article{ma2021luna,
  title={Luna: Linear unified nested attention},
  author={Ma, Xuezhe and Kong, Xiang and Wang, Sinong and Zhou, Chunting and May, Jonathan and Ma, Hao and Zettlemoyer, Luke},
  journal={Advances in Neural Information Processing Systems},
  volume={34},
  pages={2441--2453},
  year={2021}
}

@article{jaegle2021perceiverio,
  title={Perceiver IO: A general architecture for structured inputs \& outputs},
  author={Jaegle, Andrew and Borgeaud, Sebastian and Alayrac, Jean-Baptiste and Doersch, Carl and Ionescu, Catalin and Ding, David and Koppula, Skanda and Zoran, Daniel and Brock, Andrew and Shelhamer, Evan and others},
  journal={ICLR},
  year={2022}
}

@article{joshi2025replacing,
  title={RACE attention: A strictly linear-time attention layer for training on outrageously large contexts},
  author={Joshi, Sahil and Chowdhury, Agniva and Kanakamedala, Amar and Singh, Ekam and Tu, Evan and Shrivastava, Anshumali},
  journal={arXiv preprint arXiv:2510.04008},
  year={2025}
}

@inproceedings{lee2019set,
  title={Set transformer: A framework for attention-based permutation-invariant neural networks},
  author={Lee, Juho and Lee, Yoonho and Kim, Jungtaek and Kosiorek, Adam and Choi, Seungjin and Teh, Yee Whye},
  booktitle={International Conference on Machine Learning},
  pages={3744--3753},
  year={2019},
  organization={PMLR}
}

@inproceedings{tay2021synthesizer,
  title={Synthesizer: Rethinking self-attention for transformer models},
  author={Tay, Yi and Bahri, Dara and Metzler, Donald and Juan, Da-Cheng and Zhao, Zhe and Zheng, Che},
  booktitle={International Conference on Machine Learning},
  pages={10183--10192},
  year={2021},
  organization={PMLR}
}

@inproceedings{lee2022fnet,
  title={Fnet: Mixing tokens with fourier transforms},
  author={Lee-Thorp, James and Ainslie, Joshua and Eckstein, Ilya and Ontanon, Santiago},
  booktitle={Proceedings of the 2022 Conference of the north American chapter of the Association for Computational Linguistics: human language technologies},
  pages={4296--4313},
  year={2022}
}

@article{acharya2024star,
  title={Star attention: Efficient llm inference over long sequences},
  author={Acharya, Shantanu and Jia, Fei and Ginsburg, Boris},
  journal={arXiv preprint arXiv:2411.17116},
  year={2024}
}

@article{beltagy2020longformer,
  title={Longformer: The long-document transformer},
  author={Beltagy, Iz and Peters, Matthew E and Cohan, Arman},
  journal={arXiv preprint arXiv:2004.05150},
  year={2020}
}

@article{zaheer2020big,
  title={Big bird: Transformers for longer sequences},
  author={Zaheer, Manzil and Guruganesh, Guru and Dubey, Kumar Avinava and Ainslie, Joshua and Alberti, Chris and Ontanon, Santiago and Pham, Philip and Ravula, Anirudh and Wang, Qifan and Yang, Li and others},
  journal={Advances in neural information processing systems},
  volume={33},
  pages={17283--17297},
  year={2020}
}

@article{roy2021efficient,
  title={Efficient content-based sparse attention with routing transformers},
  author={Roy, Aurko and Saffar, Mohammad and Vaswani, Ashish and Grangier, David},
  journal={Transactions of the Association for Computational Linguistics},
  volume={9},
  pages={53--68},
  year={2021}
}

@article{kitaev2020reformer,
  title={Reformer: The efficient transformer},
  author={Kitaev, Nikita and Kaiser, {\L}ukasz and Levskaya, Anselm},
  journal={arXiv preprint arXiv:2001.04451},
  year={2020}
}

@article{zhang2024hedgehog,
  title={The hedgehog \& the porcupine: Expressive linear attentions with softmax mimicry},
  author={Zhang, Michael and Bhatia, Kush and Kumbong, Hermann and R{\'e}, Christopher},
  journal={arXiv preprint arXiv:2402.04347},
  year={2024}
}

@inproceedings{xiong2021nystromformer,
  title={Nystr{\"o}mformer: A nystr{\"o}m-based algorithm for approximating self-attention},
  author={Xiong, Yunyang and Zeng, Zhanpeng and Chakraborty, Rudrasis and Tan, Mingxing and Fung, Glenn and Li, Yin and Singh, Vikas},
  booktitle={Proceedings of the AAAI conference on artificial intelligence},
  volume={35},
  number={16},
  pages={14138--14148},
  year={2021}
}

@article{wang2020linformer,
  title={Linformer: Self-attention with linear complexity},
  author={Wang, Sinong and Li, Belinda Z and Khabsa, Madian and Fang, Han and Ma, Hao},
  journal={arXiv preprint arXiv:2006.04768},
  year={2020}
}

@article{lu2026zeros,
  title={{ZeroS}: {Z}ero-Sum Linear Attention for Efficient Transformers},
  author={Lu, Jiecheng and Han, Xu and Sun, Yan and Pati, Viresh and Kim, Yubin and Somani, Siddhartha and Yang, Shihao},
  journal={arXiv preprint arXiv:2602.05230},
  year={2026}
}

@inproceedings{katharopoulos2020transformers,
  title={Transformers are rnns: {F}ast autoregressive transformers with linear attention},
  author={Katharopoulos, Angelos and Vyas, Apoorv and Pappas, Nikolaos and Fleuret, Fran{\c{c}}ois},
  booktitle={International Conference on Machine Learning},
  pages={5156--5165},
  year={2020},
  organization={PMLR}
}

@inproceedings{kim2023region,
  title={Region-aware pretraining for open-vocabulary object detection with vision transformers},
  author={Kim, Dahun and Angelova, Anelia and Kuo, Weicheng},
  booktitle={Proceedings of the IEEE/CVF conference on computer vision and pattern recognition},
  pages={11144--11154},
  year={2023}
}

@inproceedings{heo2024rotary,
  title={Rotary position embedding for vision transformer},
  author={Heo, Byeongho and Park, Song and Han, Dongyoon and Yun, Sangdoo},
  booktitle={European Conference on Computer Vision},
  pages={289--305},
  year={2024},
  organization={Springer}
}

@article{everingham2015pascal,
  title={The pascal visual object classes challenge: A retrospective},
  author={Everingham, Mark and Eslami, SM Ali and Van Gool, Luc and Williams, Christopher KI and Winn, John and Zisserman, Andrew},
  journal={International Journal of Computer Vision},
  volume={111},
  pages={98--136},
  year={2015},
  publisher={Springer}
}

@article{ade20k,
  title={Semantic understanding of scenes through the ade20k dataset},
  author={Zhou, Bolei and Zhao, Hang and Puig, Xavier and Xiao, Tete and Fidler, Sanja and Barriuso, Adela and Torralba, Antonio},
  journal={International Journal of Computer Vision},
  volume={127},
  pages={302--321},
  year={2019},
  publisher={Springer}
}

@inproceedings{jajal2026adaperceiver,
  title={Adaperceiver: Transformers with adaptive width, depth, and tokens},
  author={Jajal, Purvish and Eliopoulos, Nicholas John and Chou, Benjamin Shiue-Hal and Thiruvathukal, George K and Lu, Yung-Hsiang and Davis, James C},
  booktitle={Proceedings of the IEEE/CVF Conference on Computer Vision and Pattern Recognition},
  pages={2618--2628},
  year={2026}
}

@inproceedings{context,
  title={The role of context for object detection and semantic segmentation in the wild},
  author={Mottaghi, Roozbeh and Chen, Xianjie and Liu, Xiaobai and Cho, Nam-Gyu and Lee, Seong-Whan and Fidler, Sanja and Urtasun, Raquel and Yuille, Alan},
  booktitle={Proceedings of the IEEE conference on computer vision and pattern recognition},
  pages={891--898},
  year={2014}
}

@inproceedings{caesar2018coco,
  title={Coco-stuff: Thing and stuff classes in context},
  author={Caesar, Holger and Uijlings, Jasper and Ferrari, Vittorio},
  booktitle={Proceedings of the IEEE conference on computer vision and pattern recognition},
  pages={1209--1218},
  year={2018}
}

@inproceedings{cocoobjects,
  title={Microsoft coco: Common objects in context},
  author={Lin, Tsung-Yi and Maire, Michael and Belongie, Serge and Hays, James and Perona, Pietro and Ramanan, Deva and Doll{\'a}r, Piotr and Zitnick, C Lawrence},
  booktitle={Computer vision--ECCV 2014: 13th European conference, zurich, Switzerland, September 6-12, 2014, proceedings, part v 13},
  pages={740--755},
  year={2014},
  organization={Springer}
}

@inproceedings{silberman2012nyu,
  title={Indoor segmentation and support inference from rgbd images},
  author={Silberman, Nathan and Hoiem, Derek and Kohli, Pushmeet and Fergus, Rob},
  booktitle={Computer Vision--ECCV 2012: 12th European Conference on Computer Vision, Florence, Italy, October 7-13, 2012, Proceedings, Part V 12},
  pages={746--760},
  year={2012},
  organization={Springer}
}

@article{kitti_geiger2013vision,
  title={Vision meets robotics: The kitti dataset},
  author={Geiger, Andreas and Lenz, Philip and Stiller, Christoph and Urtasun, Raquel},
  journal={The International Journal of Robotics Research},
  volume={32},
  number={11},
  pages={1231--1237},
  year={2013},
  publisher={Sage Publications Sage UK: London, England}
}

@inproceedings{deng2009imagenet,
  title={Imagenet: A large-scale hierarchical image database},
  author={Deng, Jia and Dong, Wei and Socher, Richard and Li, Li-Jia and Li, Kai and Fei-Fei, Li},
  booktitle={2009 IEEE conference on computer vision and pattern recognition},
  pages={248--255},
  year={2009},
  organization={Ieee}
}

@inproceedings{recht2019imagenetv2,
  title={Do imagenet classifiers generalize to imagenet?},
  author={Recht, Benjamin and Roelofs, Rebecca and Schmidt, Ludwig and Shankar, Vaishaal},
  booktitle={International Conference on Machine Learning},
  pages={5389--5400},
  year={2019},
  organization={PMLR}
}

@article{beyer2020weINreal,
  title={Are we done with imagenet?},
  author={Beyer, Lucas and H{\'e}naff, Olivier J and Kolesnikov, Alexander and Zhai, Xiaohua and Oord, A{\"a}ron van den},
  journal={arXiv preprint arXiv:2006.07159},
  year={2020}
}

@inproceedings{oxford,
  title={Cats and dogs},
  author={Parkhi, Omkar M and Vedaldi, Andrea and Zisserman, Andrew and Jawahar, CV},
  booktitle={2012 IEEE conference on computer vision and pattern recognition},
  pages={3498--3505},
  year={2012},
  organization={IEEE}
}

@inproceedings{Radford2021LearningTV_CLIP,
  title={Learning Transferable Visual Models From Natural Language Supervision},
  author={Alec Radford and Jong Wook Kim and Chris Hallacy and A. Ramesh and Gabriel Goh and Sandhini Agarwal and Girish Sastry and Amanda Askell and Pamela Mishkin and Jack Clark and Gretchen Krueger and Ilya Sutskever},
  booktitle={ICML},
  year={2021}
}

@article{fang2023data_dfn,
  title={Data filtering networks},
  author={Fang, Alex and Jose, Albin Madappally and Jain, Amit and Schmidt, Ludwig and Toshev, Alexander and Shankar, Vaishaal},
  journal={arXiv preprint arXiv:2309.17425},
  year={2023}
}

@misc{OpenCLIP_ilharco_gabriel_2021_5143773,
  author       = {Ilharco, Gabriel and
                  Wortsman, Mitchell and
                  Wightman, Ross and
                  Gordon, Cade and
                  Carlini, Nicholas and
                  Taori, Rohan and
                  Dave, Achal and
                  Shankar, Vaishaal and
                  Namkoong, Hongseok and
                  Miller, John and
                  Hajishirzi, Hannaneh and
                  Farhadi, Ali and
                  Schmidt, Ludwig},
  title        = {OpenCLIP},
  month        = jul,
  year         = 2021,
  publisher    = {Zenodo},
  version      = {0.1},
  doi          = {10.5281/zenodo.5143773}
}

@article{tschannen2025siglip,
  title={{SigLIP}~2: {M}ultilingual vision-language encoders with improved semantic understanding, localization, and dense features},
  author={Tschannen, Michael and Gritsenko, Alexey and Wang, Xiao and Naeem, Muhammad Ferjad and Alabdulmohsin, Ibrahim and Parthasarathy, Nikhil and Evans, Talfan and Beyer, Lucas and Xia, Ye and Mustafa, Basil and others},
  journal={arXiv preprint arXiv:2502.14786},
  year={2025}
}

@inproceedings{heinrich2025radiov2,
  title={{RADIOv2.5}: {I}mproved baselines for agglomerative vision foundation models},
  author={Heinrich, Greg and Ranzinger, Mike and Yin, Hongxu and Lu, Yao and Kautz, Jan and Tao, Andrew and Catanzaro, Bryan and Molchanov, Pavlo},
  booktitle={Proceedings of the Computer Vision and Pattern Recognition Conference},
  pages={22487--22497},
  year={2025}
}

@misc{mmseg2020,
    title={{MMSegmentation}: OpenMMLab Semantic Segmentation Toolbox and Benchmark},
    author={MMSegmentation Contributors},
    howpublished = {\url{https://github.com/open-mmlab/mmsegmentation}},
    year={2020}
}

@inproceedings{bhat2021adabins,
  title={Adabins: Depth estimation using adaptive bins},
  author={Bhat, Shariq Farooq and Alhashim, Ibraheem and Wonka, Peter},
  booktitle={Proceedings of the IEEE/CVF conference on computer vision and pattern recognition},
  pages={4009--4018},
  year={2021}
}

@inproceedings{kobayashi2020attention,
  title={Attention is not only a weight: Analyzing transformers with vector norms},
  author={Kobayashi, Goro and Kuribayashi, Tatsuki and Yokoi, Sho and Inui, Kentaro},
  booktitle={Proceedings of the 2020 Conference on Empirical Methods in Natural Language Processing (EMNLP)},
  pages={7057--7075},
  year={2020}
}

@inproceedings{wu2022flowformer,
  title     = {Flowformer: Linearizing Transformers with Conservation Flows},
  author    = {Wu, Haixu and Wu, Jialong and Xu, Jiehui and
               Wang, Jianmin and Long, Mingsheng},
  booktitle = {International Conference on Machine Learning},
  pages     = {24226--24242},
  year      = {2022}
}
